\documentclass[runningheads]{llncs}
\usepackage{graphicx}
\usepackage{comment}
\usepackage{amsmath,amssymb} % define this before the line numbering.
\usepackage{color}
\usepackage[font=small,labelfont=bf]{caption,subcaption} % font=small should give 9pt fonts given 10pt normal text.
\usepackage{amssymb}
\usepackage{pifont}
\usepackage{multirow}
\usepackage{paralist}
\usepackage{url}
\usepackage{subfiles}

\usepackage[normalem]{ulem} % For strike-through fonts. Example: \sout{Hello World}

\usepackage[pagebackref=true,breaklinks=true,colorlinks,bookmarks=false]{hyperref}

\usepackage{xspace}
\newcommand*{\eg}{e.g.\@\xspace}
\newcommand*{\ie}{i.e.\@\xspace}

\usepackage{soul}
\usepackage[table]{xcolor}

\definecolor{darkgreen}{RGB}{0, 170, 0}
\definecolor{darkred}{RGB}{200, 0, 0}
\newcommand{\cmark}{{\color{darkgreen} \ding{51}}}%
\newcommand{\xmark}{{\color{darkred} \ding{55}}}%

\newcommand{\tablefontsize}{\fontsize{8}{10}\selectfont}
\renewcommand{\paragraph}[1]{\vspace{3pt}\noindent\textbf{#1 ---}}

\newcommand{\sectionSpaceBefore}{\vspace{-1pt}}
\newcommand{\sectionSpaceAfter}{\vspace{-5pt}}
\newcommand{\subSectionSpaceBefore}{\vspace{-1pt}}
\newcommand{\subSectionSpaceAfter}{\vspace{-5pt}}

\begin{document}
% \renewcommand\thelinenumber{\color[rgb]{0.2,0.5,0.8}\normalfont\sffamily\scriptsize\arabic{linenumber}\color[rgb]{0,0,0}}
% \renewcommand\makeLineNumber {\hss\thelinenumber\ \hspace{6mm} \rlap{\hskip\textwidth\ \hspace{6.5mm}\thelinenumber}}
% \linenumbers
\pagestyle{headings}
\mainmatter
\def\ECCVSubNumber{2683}  % Insert your submission number here

\title{Learning Stereo from Single Images} % Replace with your title

% INITIAL SUBMISSION
%\begin{comment}
% \titlerunning{ECCV-20 submission ID \ECCVSubNumber}
% \authorrunning{ECCV-20 submission ID \ECCVSubNumber}
% \author{Anonymous ECCV submission}
% \institute{Paper ID \ECCVSubNumber}
%\end{comment}
%******************

% CAMERA READY SUBMISSION
\titlerunning{Learning Stereo from Single Images}
% If the paper title is too long for the running head, you can set
% an abbreviated paper title here

\author{Jamie Watson$^1$\hspace{10pt}
 Oisin Mac Aodha$^2$\hspace{10pt}
 Daniyar Turmukhambetov$^1$\\
 Gabriel~J.~Brostow$^{1,3}$\hspace{32pt}
 Michael Firman$^1$}
 %
% \index{Mac Aodha, Oisin}
%
\institute{$^1$Niantic \hspace{12pt} $^2$University of Edinburgh \hspace{12pt} $^3$UCL}

\authorrunning{J. Watson et al.}
% First names are abbreviated in the running head.
% If there are more than two authors, 'et al.' is used.
%
% \institute{Niantic \and The University of Edinburgh \\
% \url{https://github.com/nianticlabs/stereo-from-mono/}
% }

%******************
\maketitle

\begin{abstract}
Supervised deep networks are among the best methods for finding correspondences in stereo image pairs. Like all supervised approaches, these networks require ground truth data during training. However, collecting large quantities of accurate dense correspondence data is very challenging. We propose that it is unnecessary to have such a high reliance on ground truth depths or even corresponding stereo pairs.
Inspired by recent progress in monocular depth estimation, we generate plausible disparity maps from single images. In turn, we use those flawed disparity maps in a carefully designed pipeline to generate stereo training pairs. Training in this manner makes it possible to convert any collection of single RGB images into stereo training data. This results in a significant reduction in human effort, with no need to collect real depths or to hand-design synthetic data. We can consequently train a stereo matching network from scratch on datasets like COCO, which were previously hard to exploit for stereo.
Through extensive experiments we show that our approach outperforms stereo networks trained with standard synthetic datasets, when evaluated on KITTI, ETH3D, and Middlebury.
Code to reproduce our results is available at \url{https://github.com/nianticlabs/stereo-from-mono/}.

\end{abstract}

%%%%%%%%%%%%%%%%%%%%%%%%%%%%%%%%%%%%%%%%
% \sectionSpaceBefore
\vspace{-4pt}
\section{Introduction}
\sectionSpaceAfter
%%%%%%%%%%%%%%%%%%%%%%%%%%%%%%%%%%%%%%%%
Given a pair of scanline rectified images, the goal of stereo matching is to estimate the per-pixel horizontal displacement (\ie disparity) between the corresponding location of every pixel from the first view to the second, or vice versa.
This is typically framed as a matching problem, where the current best performance is achieved by deep stereo networks \eg \cite{yin2019hierarchical,guo2019group,chang2018pyramid,Zhang2019GANet}.
These networks rely on expensive to obtain ground truth correspondence training data, for example captured with LiDAR scanners \cite{geiger2013vision,geiger2012we,schops2017multi}.
Synthetic data is a common alternative; given a collection of 3D scenes, large quantities of artificially rendered stereo pairs can be created with ground truth disparities~\cite{mayer2016sceneflow,Dosovitskiy17,gaidon2016virtual,cabon2020virtual}.
Deployment in novel \emph{real} scenes currently hinges on finetuning with additional correspondence data from that target domain.
Assuming access to stereo pairs, a promising alternative could be to do self-supervised finetuning using image reconstruction losses and no ground truth correspondence.
Unfortunately, self-supervised performance is still short of supervised methods for real domains~\cite{zhou2017unsupervised,tonioni2017unsupervised,tonioni2019real}.

\begin{figure}[t]
\centering
\includegraphics[width=1.0\textwidth,page=1]{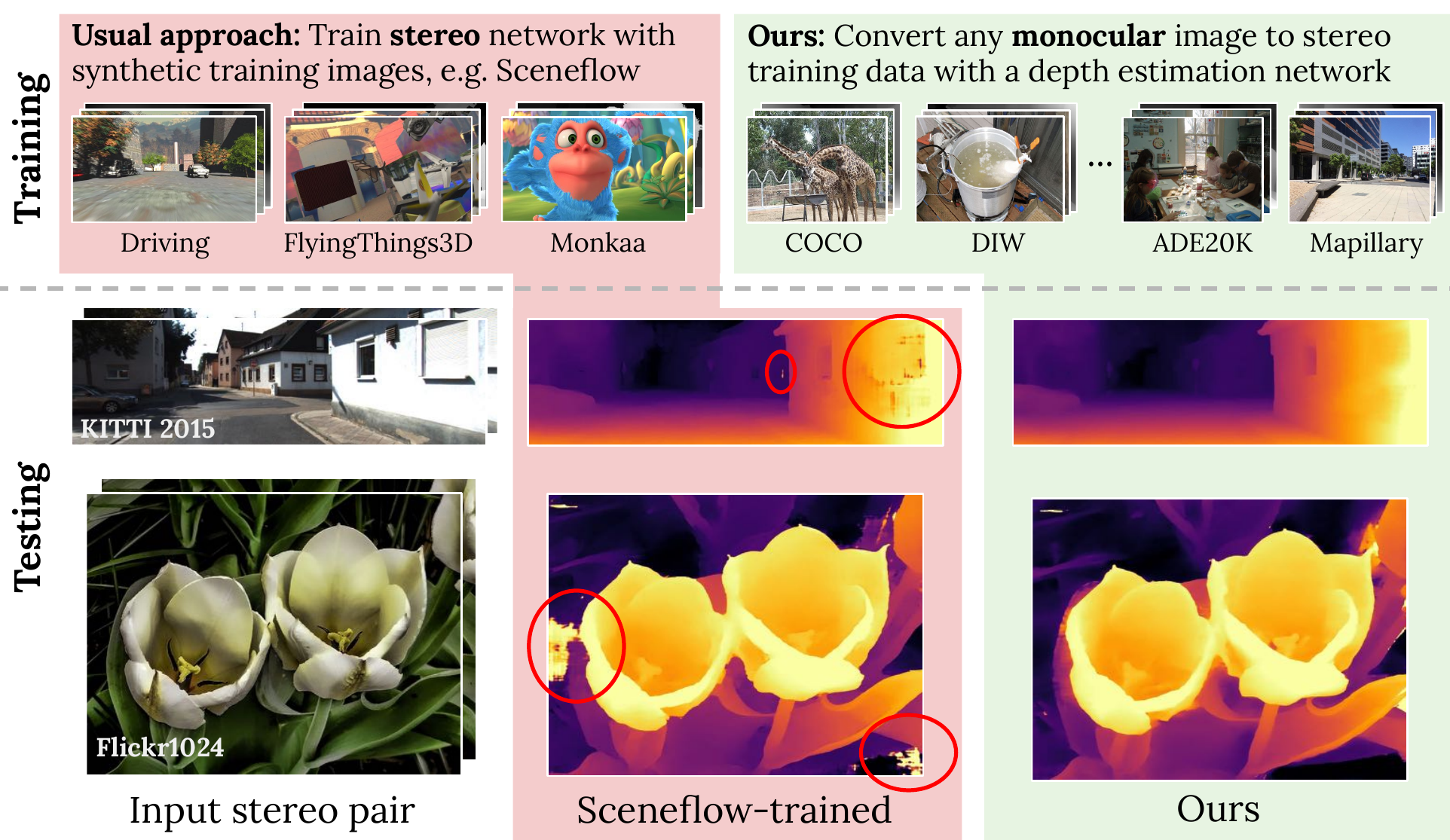}
\vspace{-12pt}
\caption{{\bf Training stereo correspondence from monocular data.}
Deep stereo networks are typically trained on synthetic data, with per-dataset finetuning when ground truth disparity is available.
We instead train on real images, which we convert to tuples of stereo training data using off-the-shelf monocular depth estimation.
Our approach offers better generalization to new datasets, fewer visual artifacts, and improved scores.
}
\label{fig:teaser}
\vspace{-15pt}
\end{figure}

We tackle the problem of generating training data for deep stereo matching networks.
Recent works have shown that pretraining on large amounts of noisily labeled data improves performance on image classification~\cite{moredata1,moredata2,moredata3,moredata4}.
A similar approach has not, to the best of our knowledge, been applied to training stereo matching networks.
We introduce a fully automatic stereo data synthesis approach which only requires a collection of single images as input.
In doing so, we open up the data available for deep stereo matching to any 2D image.
Our approach uses a monocular depth estimation network to predict a disparity map for each training image.
We refine this initial prediction before using it to synthesize a corresponding stereo pair from a new viewpoint, taking care to address issues associated with occlusions and collisions.
Using natural images, as opposed to synthetic ones, ensures that the stereo network sees \emph{realistic} textures and layouts, with \emph{plausible} disparity maps during training.
We show that networks trained from monocular data, or even low quality depth, provide an informative signal when generating training data.
This is because the stereo network still has to learn how to perform dense matching between the real input image and our synthesized new view.
In our experiments, we show that progressively better monocular depth yields gains in stereo matching accuracy. This suggests that our method will benefit `for free' as future monocular depth networks, including self-supervised ones, improve further.
We also show that increasing the amount of monocular derived training data also increases stereo performance.
We present an example of a stereo network trained with our data compared to using conventional synthetic data in Fig.~\ref{fig:teaser}.

We make the following contributions:
\vspace{-5pt}
\begin{enumerate}
    \item A fully automatic pipeline for generating stereo training data from unstructured collections of single images, given a depth-from-color model.
    \item A comparison to different approaches for stereo pair synthesis, showing that even simple approaches can produce useful stereo training data.
    \item A detailed experimental evaluation across multiple domains with different stereo algorithms, showing that our generated training data enables better generalization to unseen domains than current synthetic datasets.
\end{enumerate}

%%%%%%%%%%%%%%%%%%%%%%%%%%%%%%%%%%%%%%%%
\sectionSpaceBefore
\section{Related Work}
\sectionSpaceAfter
%%%%%%%%%%%%%%%%%%%%%%%%%%%%%%%%%%%%%%%%
%The most related works can be grouped into methods for learning-based stereo matching, ground truth training data generation, and stereo domain adaptation.

\paragraph{Stereo Matching}
Traditionally, matching between stereo image pairs was posed as an energy minimization problem consisting of separate, hand-designed, appearance matching and data smoothness terms \eg \cite{hirschmuller2007stereo,woodford2009global,bleyer2011patchmatch}.
\cite{ladicky2015learning,vzbontar2016stereo} instead learned the appearance matching term while still retaining conventional stereo methods for the additional processing.
Subsequently, fully end-to-end approaches replaced the entire conventional stereo matching pipeline~\cite{mayer2016sceneflow,kendall2017end}.
Recent advances have focused on improving stereo network architectures and training losses.
This has included innovations like spatial pyramids \cite{chang2018pyramid} and
more sophisticated approaches for regularizing the generated cost volume~\cite{Zhang2019GANet,zhang2019domaininvariant,cheng2019learning}.
In this work, we propose an approach for generating training data for deep stereo matching algorithms that is agnostic to the choice of stereo network.
Our experiments across multiple different stereo networks show consistent improvements, independent of the underlying architecture used.

Just as recent works have adapted stereo models to help improve monocular depth estimation~\cite{guo2018learning,luo2018single,tosi2019learning,watson-2019-depth-hints}, in our work we leverage  monocular models to improve stereo.
This is related to concurrent work~\cite{Aleotti2020}, where a monocular completion network is utilized to help self-supervised stereo.
However, in contrast to these approaches, we do not require stereo pairs or ground truth disparity for data synthesis, but instead use monocular depth networks to generate training data for stereo networks.

\paragraph{Stereo Training Data}
While deep stereo algorithms have proven very successful, the best performing methods are still fully supervised and thus require  large quantities of stereo pairs with \emph{ground truth} correspondence data at training time \cite{geiger2012we,schops2017multi}.
Acquiring this data for real world scenes typically involves specially designed depth sensing hardware, coupled with conventional intensity cameras for capturing scene appearance \eg \cite{geiger2013vision,geiger2012we,schops2017multi}.
In addition to currently being laborious and expensive, these commonly used laser-based depth sensors are limited in the types of scenes they can operate in \eg no direct sunlight.% and only with non-reflective surface materials.

Other potential sources of stereo data include static stereo cameras~\cite{xian2018monocular}, internet videos~\cite{wang2019web}, 3D movies~\cite{lasinger2019towards}, and multi-view images~\cite{li2018megadepth}.
Crucially however, these images do not come with ground truth correspondence.
Estimating disparity for these images is especially challenging due to issues such as negative disparity values in the case of 3D movies and extreme appearance changes over time in multi-view image collections.
These issues make it difficult to apply conventional stereo algorithms to process the data.
It also poses a `chicken and egg' problem, as we first need accurate stereo to extract reliable correspondences.

Synthetically rendered correspondence data is an attractive alternative as it leverages $3$D graphics advances in place of specialized hardware.
One of the first large-scale synthetic datasets for stereo matching consisted of simple 3D primitives floating in space \cite{mayer2016sceneflow}.
Other synthetic datasets have been proposed to better match the statistics of real world domains, such as driving scenes \cite{Dosovitskiy17,gaidon2016virtual,cabon2020virtual} or indoor man-made environments \cite{li2018interiornet,wang2019irs}.
With synthetic data, it is also possible to vary properties of the cameras viewing the scene (\eg the focal length) so the trained models can be more robust to these variations at test time~\cite{zhang2019domaininvariant}.

While dense correspondence networks can be trained from unrealistic synthetic data \cite{dosovitskiy2015flownet,mayer2018makes}, the best performance is still achieved when synthetic training data is similar in appearance and depth to the target domain \cite{zhang2019domaininvariant}, and most networks still finetune on stereo pairs from the target domain for state-of-the-art performance \eg \cite{chang2018pyramid,Zhang2019GANet,zhang2019domaininvariant}.
This suggests that the goal of general stereo matching by training with just synthetic data has not yet been reached.
It is possible to simulate stereo training data that matches the target domain, \eg \cite{tosi2019learning,Dosovitskiy17,zhang2019domaininvariant}.
However, constructing varied scenes with realistic shape and appearance requires significant manual effort, and can be expensive to render.
We perform experiments that show that our approach to generating training data generalizes better to new scenes, while requiring no manual effort to create.

\paragraph{Image Augmentation Based Training Data}
An alternative to using computer generated synthetic data is to synthesize images by augmenting natural training images with pre-defined and known pixel-level transformations.
The advantage of this approach is that it mostly retains the realistic appearance information of the input images.
One simple method is to apply a random homography to an input image to create its corresponding matching pair~\cite{detone18superpoint}.
Alternative methods include more sophisticated transformations such as pasting one image, or segmented foreground object, on top of another background image~\cite{dosovitskiy2015flownet,mayer2018makes,janai2018unsupervised}, random non-rigid deformations of foreground objects \cite{kanazawa2016warpnet,thewlis2017unsupervised,le2018unsupervised}, and using supervised depth and normals to augment individual objects~\cite{alhaija2018geometric}.

Augmentation-based approaches are largely heuristic and do not produce realistic correspondence data due to their inability to model plausible occlusions and depth.
In contrast, we use off-the-shelf networks trained for monocular depth estimation to guide our image synthesis, so depths and occlusions are plausible.
We perform experimental comparisons with existing image augmentation approaches and show superior performance with our method.

\paragraph{Domain Adaptation for Stereo}
There is a large body of work that explores the problem of domain adaptation in stereo by addressing the domain gap between synthetic training data and real images.
Existing approaches use additional supervision in the form of highly confident correspondences \cite{tonioni2017unsupervised}, apply iterative refinement \cite{pang2018zoom}, meta-learning \cite{tonioni2019learning}, consistency checks \cite{zhou2017unsupervised,zhong2017self}, occlusion reasoning \cite{li2018occlusion}, or temporal information \cite{tonioni2019real,wang2019unos}.
In the context of fully supervised adaptation, \cite{zhang2019domaininvariant} proposed a novel domain normalization layer to account for statistical differences between different data domains.

To address the domain adaptation problem we propose a method for automatically synthesizing large quantities of diverse stereo training data from single image collections.
We show that our approach results in a large performance increase when testing on domains that have not been seen at all during training.

%%%%%%%%%%%%%%%%%%%%%%%%%%%%%%%%%%%%%%%%%%%%
\sectionSpaceBefore
\section{Method}
\sectionSpaceAfter
%%%%%%%%%%%%%%%%%%%%%%%%%%%%%%%%%%%%%%%%%%%%

\begin{figure}[t]
\centering
% \includegraphics[width=1.0\textwidth,page=2]{figures/diagrams.pdf}
% Bear is coco image: ('train2017', '000000000502')
\includegraphics[width=1.0\textwidth]{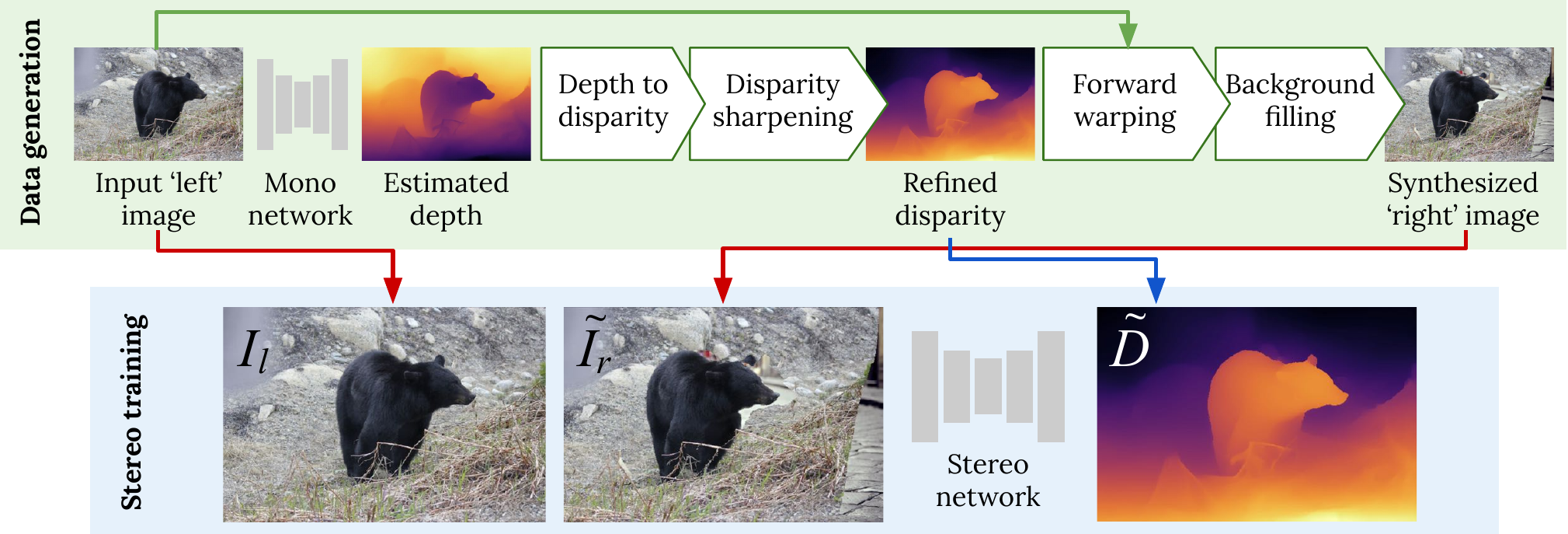}
\vspace{-15pt}
\caption{{\bf Overview of our data generation approach.}  We use off-the-shelf monocular depth estimation to synthesize stereo training pairs from single images. We convert estimated depth to a sharpened disparity map, and then forward warp pixels from the input left image $I_l$ to create a plausible and geometrically accurate right image $\tilde{I}_r$.}
\label{fig:overview}
\vspace{-15pt}
\end{figure}

Supervised training of deep stereo networks requires rectified pairs of color images $I_l$ and $I_r$, together with a ground truth disparity map $D$.\footnote{For simplicity of notation we assume that disparity and depth maps are all aligned to the left input image $I_l$.}
We propose a method to create plausible tuples of training data $(I_l, \tilde{I}_r, \tilde{D})$ from collections of individual and unrelated input images.
We use a pretrained monocular depth network to estimate a plausible depth map $Z$ for an image $I_l$, which we then use to synthesize $\tilde{I}_r$ and $\tilde{D}$.
This section describes in detail the important design decisions required at each step of this process.
Our goal here is similar to works that synthesize novel views of a single image \eg \cite{choi2019extreme,zhou2016view}.
However, our aim is to synthesize a new view together with a geometrically correct depth map aligned with the input image, which is non-trivial to obtain from existing methods.
We choose a simple and efficient algorithm for this based on pixel manipulation, which we found to be surprisingly effective.

%%%%%%%%%%%%%%%%%%%%%%%%%%%%%%%%%%%%%%%%%%%%%%%%%%%%%%
\subSectionSpaceBefore
\subsection{Stereo Training Data from Monocular Depth}
\subSectionSpaceAfter

At test time, a monocular depth estimation network $g$ takes as input a single color image, and for each pixel estimates the depth $Z$ to the camera,
\begin{align}
    Z = g(I).
\end{align}
These monocular networks can be parameterized as deep neural networks that are trained with ground truth depth \cite{eigen2014depth} or via self-supervision~\cite{garg2016unsupervised,godard2017unsupervised,zhou2017unsupervisedsfm}.
To train our stereo network we need to covert the estimated depth $Z$ to a disparity map $\tilde{D}$. To achieve our aim of generalizable stereo matching, we want to simulate stereo pairs with a wide range of baselines and focal lengths. We achieve this by converting depth to disparity with $\tilde{D}=\frac{s \, Z_{\max}}{Z}$, where $s$ is a randomly sampled (uniformly from $[d_{\min}, d_{\max} ]$) scaling factor which ensures the generated disparities are in a plausible range.

Our goal is to synthesize a right image $\tilde{I}_r$ from the input `left' image $I_l$ and the predicted disparity $\tilde{D}$.
The disparity map and the color image are both aligned to the left image $I_l$, so backward warping \cite{schwarz2007non,garg2016unsupervised} is not appropriate.
Instead, we use $\tilde{D}$ to synthesize a stereo pair $\tilde{I}_r$ via \emph{forward} warping \cite{schwarz2007non}. For each pixel $i$ in $I_l$, we translate it $\tilde{D}^i$ pixels to the left, and perform interpolation of the warped pixel values to obtain $\tilde{I}_r$.
% Each pixel location in $\tilde{I}_r$ is thus assigned the color value of the warped pixel which lands most closely to it.
Due to the nature of forward warping, $\tilde{I}_r$ will contain missing pixel artifacts due to occluded regions, and in some places collisions will occur when multiple pixels will land at the same location.
Additionally, monocular networks often incorrectly predict depths around depth discontinuities, resulting in unrealistic depth maps.
Handling these issues is key to synthesising realistic stereo pairs, as we demonstrate in our ablation experiments.

%%%%%%%%%%%%%%%%%%%%%%%%%%%%%%%%%%%%
\subSectionSpaceBefore
\subsection{Handling Occlusion and Collisions}
\subSectionSpaceAfter

Naively synthesizing images based on the estimated disparity will result in two main sources of artifacts: occlusion holes and collisions.
Occluded regions are pixels in one image in a stereo pair which are not visible in the corresponding image \cite{wang2019local}.
In the forward warping process, pixels in $\tilde{I}_r$ with no match in $I_l$ manifest themselves as \emph{holes} in the reconstructed image $\tilde{I}_r$.
While the synthesized image gives an accurate stereo pair for $I_l$, the blank regions result in unnatural artifacts.
We can increase the realism of the synthesized image by filling the missing regions with a texture from a randomly selected image $I_b$ from our training set, following \cite{dwibedi2017cut}.
We perform color transfer between $I_l$ and $I_b$ using \cite{reinhard2001color} to obtain $\hat{I}_b$.
We subsequently set all missing pixels in $\tilde{I}_r$ to the values of $\hat{I}_b$ at the corresponding positions.
This results in textures from natural images being copied into the missing regions of $\tilde{I}_r$.
Separately, \emph{collisions} are pixels that appear in $I_l$ but are not in $\tilde{I}_r$, and thus multiple pixels from $I_l$ can map to the same point in $\tilde{I}_r$.
For collisions, we select the pixel from $I_l$ which has the greater disparity value since these pixels are closer and should be visible in both images.

%%%%%%%%%%%%%%%%%%%%%%%%%%%%%%%%%%%%
\subSectionSpaceBefore
\subsection{Depth Sharpening}
\subSectionSpaceAfter

\begin{figure*}[t]
\centering
\includegraphics[width=1.0\textwidth]{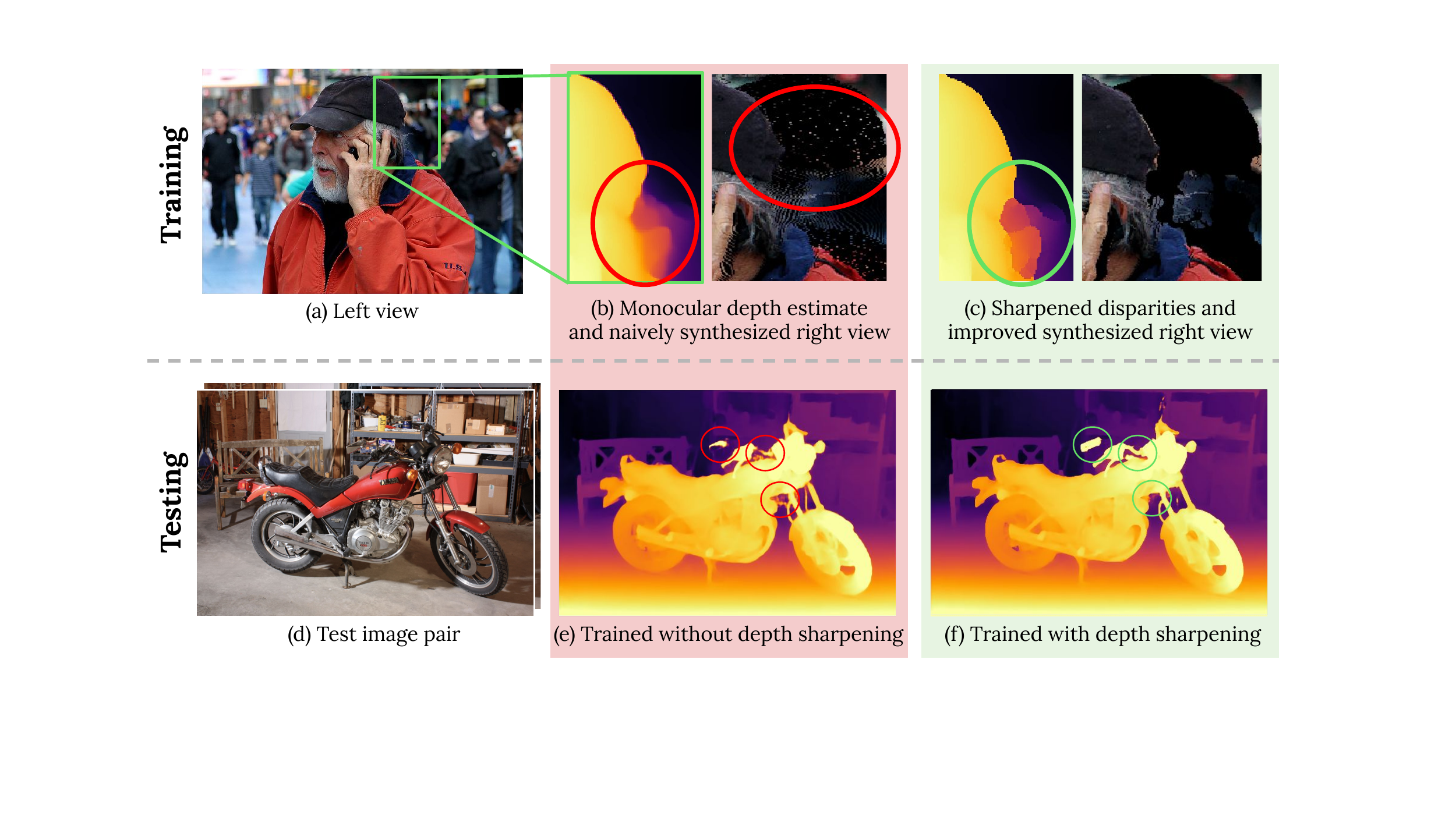}
\vspace{-15pt}
\caption{{\bf Depth map sharpening.}
Depth maps from monocular depth networks typically have `blurry' edges, which manifest themselves as flying pixels in $\tilde{I}_r$ (b).
Our depth sharpening method reduces the effect of flying pixels (c).
Models trained with depth sharpening have better scores (Table~\ref{tab:ablation}) and fewer visual artifacts; (e) versus (f).}
\label{fig:depth_sharpening}
\vspace{-5pt}
\end{figure*}

As opposed to realistic sharp discontinuities, monocular depth estimation networks typically produce blurry edges at depth discontinuities~\cite{ramamonjisoa2019sharpnet}.
When forward warping during image synthesis this leads to `flying pixels'~\cite{reynolds2011capturing}, where individual pixels are warped to empty regions between two depth surfaces.
In Fig.~\ref{fig:depth_sharpening} we observe this phenomenon for the monocular depth network of \cite{lasinger2019towards}.
Not correcting this problem would result in a collection of synthesized training images where the stereo matching algorithm is expected to match pixels in $I_l$ to individual, isolated pixels in $\tilde{I}_r$; a highly unrealistic test-time scenario.
To address flying pixels in our synthesized $\tilde{I}_r$ images we perform \emph{depth sharpening}.
We identify flying pixels as those for which the disparity map has a Sobel edge filter response of greater than 3~\cite{sobel19683x3,Hu2018RevisitingSI}.
These pixels are then assigned the disparity of the nearest `non-flying' pixel in image space.
We use this corrected disparity map for both image sampling and when training our stereo networks.

%%%%%%%%%%%%%%%%%%%%%%%%%%%%%%%%%%%%
\subSectionSpaceBefore
\subsection{Implementation Details}
\subSectionSpaceAfter
\label{sec:implementation}
\paragraph{Training Details}
Unless otherwise stated, we use the widely-used PSMNet hourglass network  variant~\cite{chang2018pyramid} for stereo matching, using the official architecture and training code.
We train all models with a batch size of 2 for 250,000 steps, equivalent to 7 epochs of the Sceneflow dataset~\cite{mayer2016sceneflow}.
With the exception of some baselines, we do not use any synthetic data when training our models.
We follow \cite{chang2018pyramid} in using a learning rate of 0.001 for all experiments.
Unless otherwise stated, we perform monocular depth estimation using MiDaS~\cite{lasinger2019towards}, a network with a ResNeXt-101~\cite{Xie2016AggregatedRT} backbone, trained on a mixture of monocular and stereo datasets including 3D movies.
Their training depths were generated by running a supervised optical flow network~\cite{ilg2017flownet} on their stereo training pairs.

Our networks are set to predict a maximum disparity of 192. We set $d_{\min}=50$ and $d_{\max}=225$, which produces training images with a diverse range of baselines and focal lengths, including stereo pairs with disparity ranges outside the network's modeling range for further robustness to test time scenarios.
During training, we take crops of size $608 \times 320$. If images are smaller than these dimensions (or significantly larger, e.g. Mapillary~\cite{MVD2017}), we isotropically resize to match the constraining dimension.
Additionally, we follow PSMNet~\cite{chang2018pyramid} by performing ImageNet~\cite{imagenet_cvpr09} normalization on all images.

\paragraph{Training Images}
Our method allows us to train using any color images.
To maximise the ability of our algorithm to learn to adapt to different test domains, we train models on a combination of varied single image datasets which we call the `Mono for Stereo' dataset, or \textbf{MfS}.
MfS comprises all the training images (without depth or semantic labels) from COCO 2017~\cite{lin2014microsoft},  Mapillary Vistas \cite{MVD2017}, ADE20K \cite{zhou2017scene}, Depth in the Wild \cite{nips2016single}, and DIODE~\cite{diode_dataset} (see Fig.~\ref{fig:teaser} for examples).
This results in 597,727 training images, an order of magnitude more than Sceneflow's 35,454 training pairs, from which we use a random subset of 500,000, unless stated otherwise.

% Individual counts are as below, which sum up to 597,727:
% COCO 2017: 118287
% Mapillary: 18000
% ADE20K: 20210
% DIW: 415772
% DIODE: 25458

\paragraph{Color Augmentation}
Cameras in a stereo pair are typically nominally identical.
However, each image in a pair may have slightly different white balancing, lens geometry, glare, and motion blur.
To account for these differences when training stereo networks we augment $\tilde{I}_r$ with pixel-level noise with standard deviation 0.05, and we randomly adjust contrast, brightness, saturation, and hue with ranges of $\pm0.2$, $\pm0.2$, $\pm0.2$, and $\pm0.01$.
Finally, with $50\%$ probability, we add Gaussian blur to $\tilde{I}_r$ with kernel size $\sigma\sim\text{Unif}\,[0, 1]$.
Interestingly, we observe that these simple augmentations also significantly  improve the performance of the baseline stereo models trained on Sceneflow~\cite{mayer2016sceneflow} alone.

%%%%%%%%%%%%%%%%%%%%%%%%%%%%%%%%%%%
\sectionSpaceBefore
\section{Experiments}
\sectionSpaceAfter
%%%%%%%%%%%%%%%%%%%%%%%%%%%%%%%%%%%

In this section we present experimental results to validate our approach to stereo data generation.
Our experiments show that:
\vspace{-5pt}
\begin{enumerate}
    \item Our approach produces better and more general stereo matching networks compared to training on alternative sources of data, such as synthetic scenes
    \item We are robust to errors in monocular depth predictions, and we perform well regardless of the source of monocular training data
    \item Our design decisions are sensible, validated through ablation of our method
    \item Our performance gains are agnostic to the stereo architecture used
    \item As the amount of generated data increases, so does stereo performance
    \item Finetuning models trained with our approach compared to synthetic data with ground truth disparity results in better performance.
\end{enumerate}

\subSectionSpaceBefore
\subsection{Evaluation Datasets and Metrics}
\subSectionSpaceAfter
We evaluate multiple stereo models on a variety of datasets to demonstrate generality of our approach.
Our test datasets are:
(i) \textbf{KITTI 2012} and \textbf{2015} \cite{geiger2012we,geiger2013vision}: real-world driving scenes with LiDAR ground truth,
(ii) \textbf{ETH3D} \cite{schops2017multi}: grayscale stereo pairs from indoor and outdoor environments, again with LiDAR ground truth, and
(iii) \textbf{Middlebury} \cite{scharstein2014high}: high-resolution stereo pairs with structured light ground truth.
Middlebury and ETH3D are particularly compelling for demonstrating general-purpose stereo matching as their training data is very limited, with just 15 and 27 pairs respectively.
Note that we never train on any of these datasets, with the exception of KITTI finetuning experiments in Section~\ref{sec:kitti_finetuning}.
For Middlebury, we evaluate on the 15 training images provided for the official stereo benchmark.
We report results on the full training set for the two view benchmark for ETH3D. For both KITTI 2012 and 2015 we evaluate on the validation set from \cite{chang2018pyramid}, and provide the image indices in the supplementary material.
For all datasets we report end-point-error (EPE) and the thresholded error rate: the $\%$ of pixels with predicted disparity more than $\tau$ pixels from the ground truth, \eg$>$3px.
Unless otherwise stated, we evaluate every pixel for which there is a valid ground truth disparity value (i.e.~we include occluded pixels).
We only report scores with the most commonly reported value of $\tau$ for each dataset.
Finally, we show qualitative results on \textbf{Flickr1024} \cite{flickr1024}, a diverse dataset of stereo pairs originally captured by internet users for stereoscopic viewing.

%%%%%%%%%%%%%%%%%%%%%%%%%%%%%%%%%%%%%%%%
\subSectionSpaceBefore
\subsection{Comparison to Alternative Data Generation Methods}
\subSectionSpaceAfter

%%%%%%%% KITTI VAL SET FOR EVAL %%%%%%%%%%%%%
\begin{table}[t]
\vspace{-10pt}
\tablefontsize
\centering
\caption{{\bf Our approach outperforms alternative stereo training data generation methods.}
Each row represents a single PSMNet trained with a different data synthesis approach, without any dataset-specific finetuning.
We can see that simple synthesis methods still produce valuable training data.
However, our approach in the bottom row, which incorporates a monocular depth network~\cite{lasinger2019towards}, outperforms even synthetic Sceneflow data.
Numbers here are directly comparable to  Table~\ref{tab:monodepth_model_compare}, showing we beat the baselines no matter what depth network or training data we use.}

\label{tab:we_beat_alternative_data_sources}
\resizebox{1.0\textwidth}{!}{
\begin{tabular}{|l|l||cc|cc|cc|cc|}
\hline
Synthesis approach & Training data &
\multicolumn{2}{c|}{KITTI '12}  &
\multicolumn{2}{c|}{KITTI '15} &
\multicolumn{2}{c|}{Middlebury} &
\multicolumn{2}{c|}{ETH3D} \\
& & EPE & $>$3px & EPE & $>$3px & EPE & $>$2px & EPE & $>$1px \\
\hline
 Affine warps (\eg \cite{detone18superpoint}) & MfS & 3.74 & 14.78 & 2.33 & 14.94 & 21.50 & 61.19 & 1.28 & 24.21 \\
 Random pasted shapes (\eg \cite{mayer2018makes}) & MfS & 2.45 & 11.89 & 1.49 & 7.57 & 11.38 & 40.37 & 0.77 & 14.19 \\
 Random superpixels & MfS & 1.28 & 6.08 & 1.23 & 5.90 & 12.02 & 32.92 & 2.01 & 13.95 \\
 Synthetic & Sceneflow \cite{mayer2016sceneflow}  & 1.04 & 5.76 & 1.19 & 5.80 & 9.42 & 34.99 & 1.25 & 13.04 \\
 SVSM selection module~\cite{luo2018single} & MfS &  1.04 & 5.66 & 1.10 &5.47 &8.45 &31.34 &0.57 &11.61 \\
 \hline
 Ours &  MfS  & \textbf{0.95} & \textbf{4.43} & \textbf{1.04} & \textbf{4.75} & \textbf{6.33} & \textbf{26.42} & \textbf{0.53} & \textbf{8.17} \\
\hline
\end{tabular}}
\vspace{-10pt}
\end{table}

In our first experiment, we compare our approach to alternative baseline methods for stereo data generation.
To do this, we train a PSMNet stereo network using data generated by each of the baselines and compare the network's stereo estimation performance on three held out datasets.
We compare to:
(i)~\textbf{Affine warps}: warps of single images similar to~\cite{detone18superpoint},
(ii)~\textbf{Random pasted shapes}: our implementation of the optical flow data generation method of~\cite{mayer2018makes}, adapted for stereo training,
(iii)~\textbf{Random superpixels}: where we initialise a disparity map as a plane, and assign disparity values drawn uniformly from $[d_{\min}, d_{\max} ]$ to randomly selected image superpixels~\cite{Felzenszwalb2004},
(iv)~\textbf{Synthetic}: 35K synthetic image pairs from Sceneflow~\cite{mayer2016sceneflow}, and
(v)~\textbf{SVSM selection module}: our implementation of the selection module from \cite{luo2018single}, where $\tilde{I}_r$ is the weighted sum of shifted left images, with weightings derived from the depth output of~\cite{lasinger2019towards}.
With the exception of the synthetic baseline, all results use the MfS dataset to synthesize stereo pairs.
Detailed descriptions are in the supplementary material.
Table~\ref{tab:we_beat_alternative_data_sources} shows that our fully automatic approach outperforms all of these methods.
Qualitative results are in Figs.~\ref{fig:teaser}, \ref{fig:qualitative_comparisons} and~\ref{fig:more_baseline_qualtitative}.

\begin{figure}
\captionsetup[subfigure]{font=footnotesize,labelfont=footnotesize}
\newcommand{\itemwidth}{1.0\textwidth}
\centering
  \begin{subfigure}[b]{0.32\linewidth}
    \centering\includegraphics[width=\itemwidth]{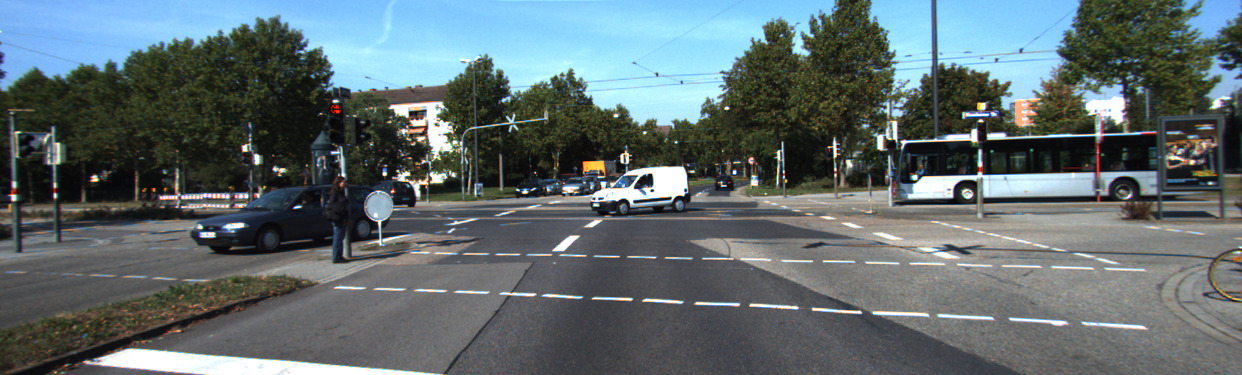}\vspace{-5pt}
    \caption{Input (left) image}
  \end{subfigure}
  \begin{subfigure}[b]{0.32\linewidth}
    \centering\includegraphics[width=\itemwidth]{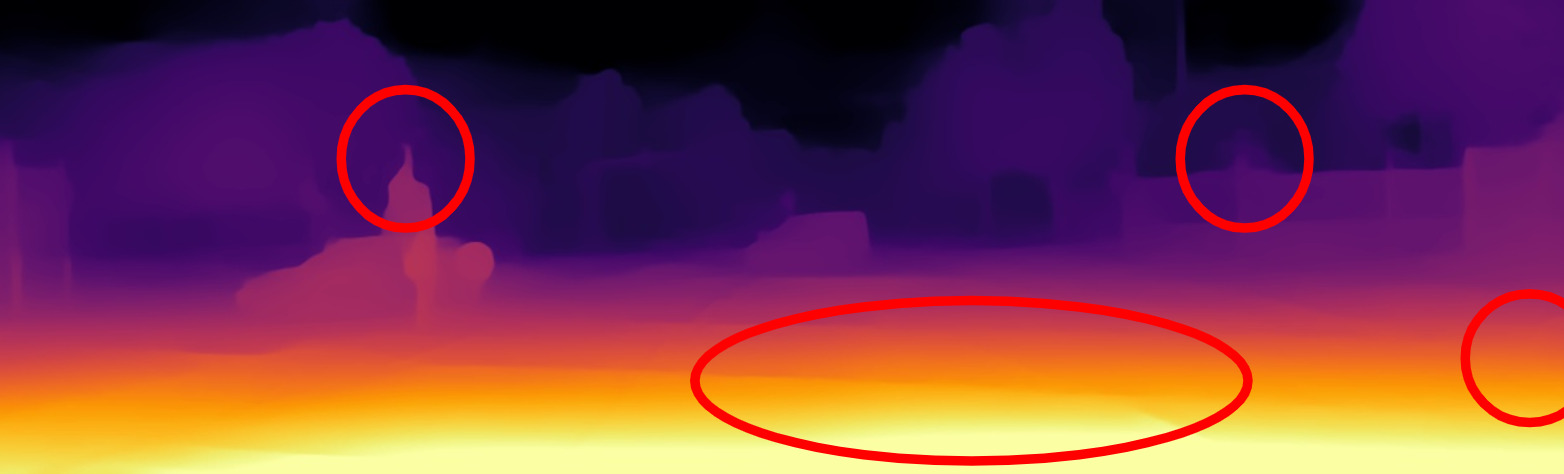}\vspace{-5pt}
    \caption{Mono~estimation \cite{lasinger2019towards}}
  \end{subfigure}
  \begin{subfigure}[b]{0.32\linewidth}
    \centering\includegraphics[width=\itemwidth]{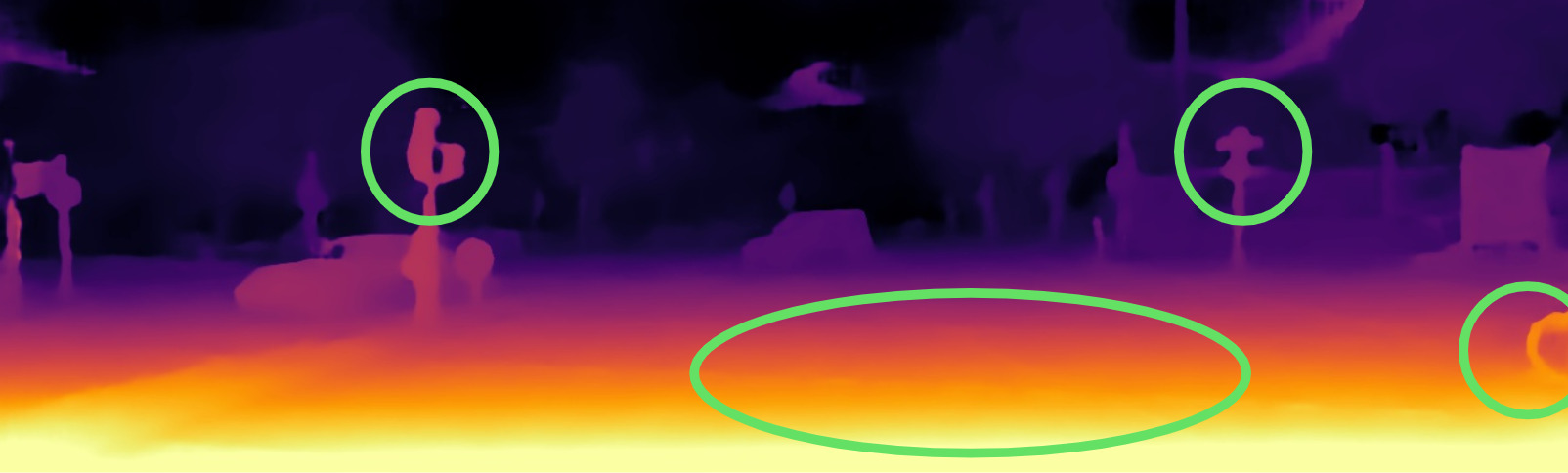}\vspace{-5pt}
    \caption{Our stereo estimation}
  \end{subfigure}
  \vspace{-5pt}
\caption{{\bf We can recover from errors in monocular depth estimation.}
Problems present in monocular depths (b), such as missing objects and uneven ground do not transfer to our eventual stereo predictions (c).}
\label{fig:we_fix_mono}
\vspace{-2pt}
\end{figure}

\begin{figure}
\vspace{-2pt}
\captionsetup[subfigure]{font=footnotesize,labelfont=footnotesize}
\newcommand{\itemwidth}{1.0\textwidth}
\centering
  \begin{subfigure}[b]{0.32\linewidth}
    \centering\includegraphics[width=\itemwidth]{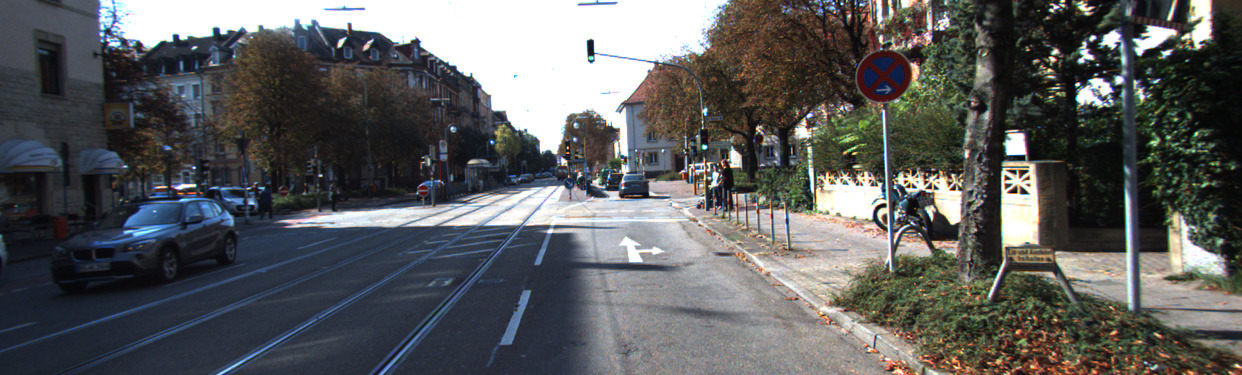}\vspace{-5pt}
    \caption{Input (left) image}
  \end{subfigure}
  \begin{subfigure}[b]{0.32\linewidth}
    \centering\includegraphics[width=\itemwidth]{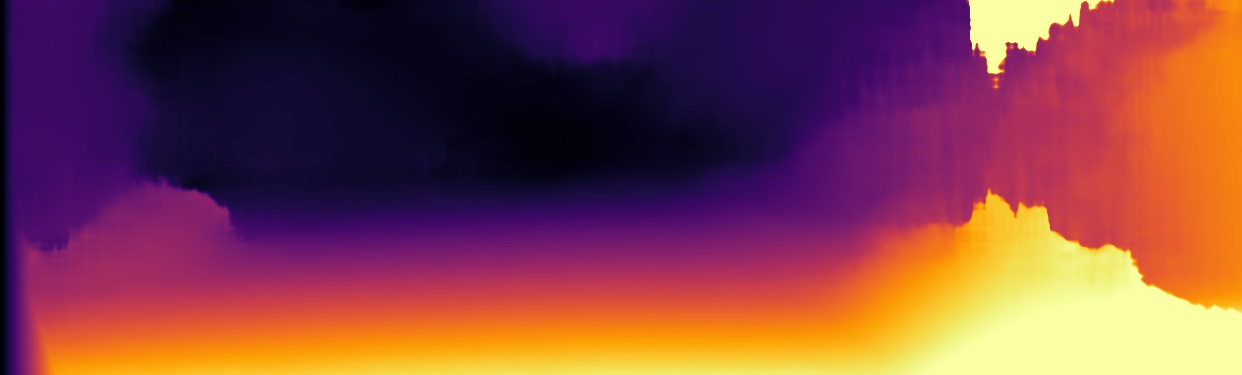}\vspace{-5pt}
    \caption{Affine warps}
  \end{subfigure}
  \begin{subfigure}[b]{0.32\linewidth}
    \centering\includegraphics[width=\itemwidth]{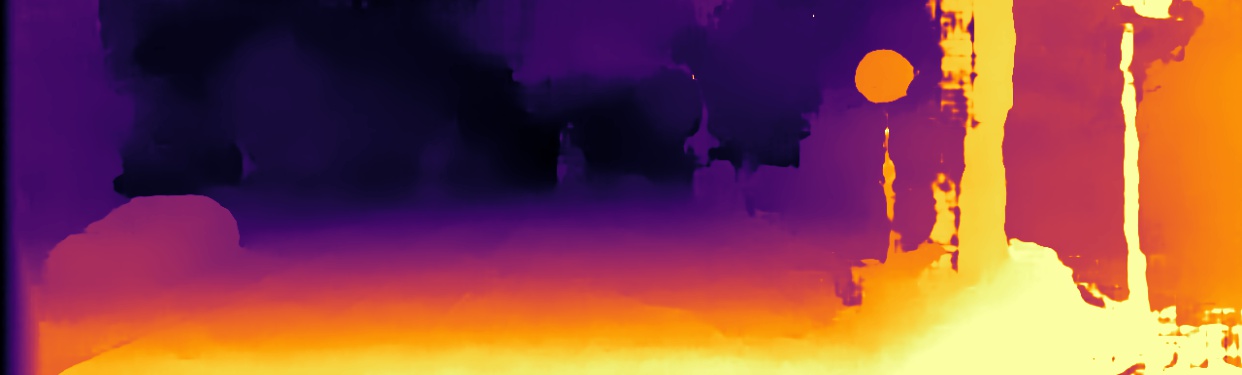}\vspace{-5pt}
    \caption{Random pasted shapes}
  \end{subfigure} \\\vspace{3pt}
  \begin{subfigure}[b]{0.32\linewidth}
    \centering\includegraphics[width=\itemwidth]{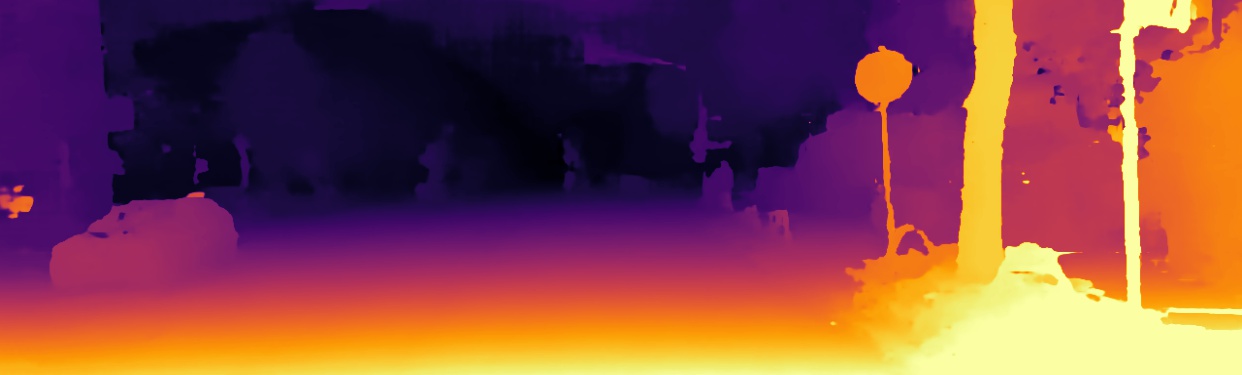}\vspace{-5pt}
    \caption{Random superpixels}
  \end{subfigure}
  \begin{subfigure}[b]{0.32\linewidth}
    \centering\includegraphics[width=\itemwidth]{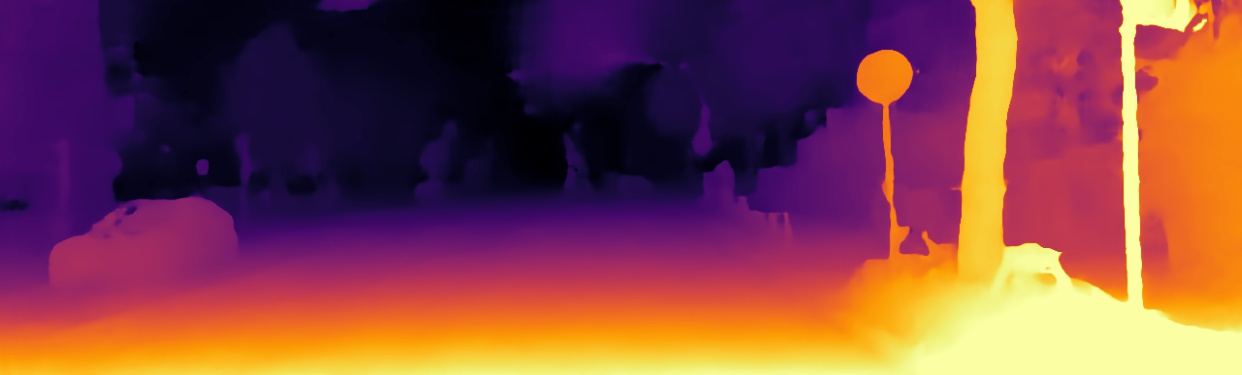}\vspace{-5pt}
    \caption{Ours (PSMNet)}
  \end{subfigure}
  \begin{subfigure}[b]{0.32\linewidth}
    \centering\includegraphics[width=\itemwidth]{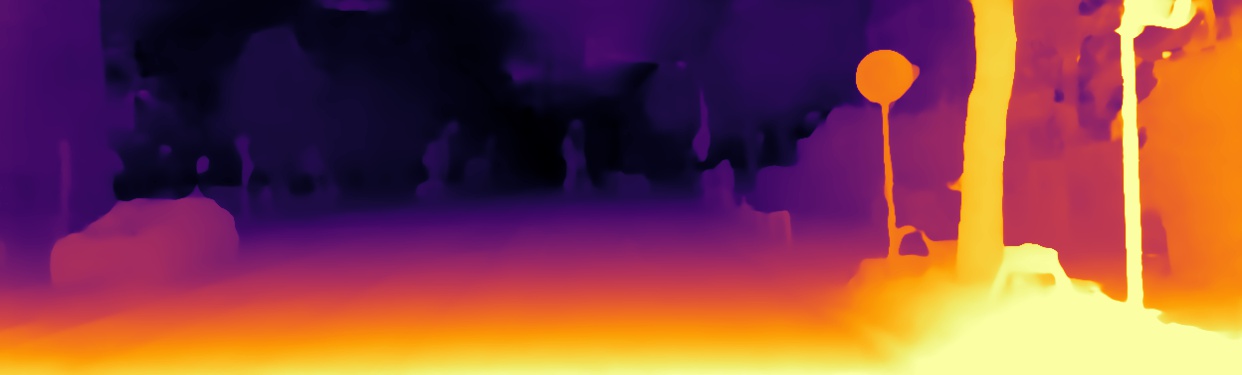}\vspace{-5pt}
    \caption{Ours (GANet)}
  \end{subfigure}
  \vspace{-5pt}
\caption{{\bf Comparison to baseline data generation methods.}
We show that simple image synthesis can be a surprisingly effective method for generating stereo training data \eg (c) and (d).
However, our approach produces sharper and cleaner disparities, whether using PSMNet (e) or the larger GANet (f).
\vspace{-5pt}
}
\label{fig:more_baseline_qualtitative}
\end{figure}

%%%%%%%%%%%%%%%%%%%%%%%%%%%%%%%%%%%%%%%%
\subSectionSpaceBefore
\subsection{Model Architecture Ablation}
\subSectionSpaceAfter

In Table~\ref{tab:architecture_ablation} we use our data synthesis approach to train three different stereo matching networks.
With the exception of iResNet~\cite{liang2018learning}, we used the same architectures and training code provided by the original authors, and trained each model for 250,000 iterations from scratch.
% , without Sceneflow pretraining.
For fair comparison, we also trained the same models without our data but instead with Sceneflow using the same color augmentations described in Section~\ref{sec:implementation}.
Interestingly, we observe that this results in an improvement over the publicly released model snapshots.
We see that our approach consistently outperforms Sceneflow training irrespective of the stereo network used.

\begin{table}[h]
\tablefontsize
\centering
\vspace{-10pt}
\caption{{\bf Our approach is agnostic to stereo model architecture.}
We compare three diverse stereo architectures, from the lightweight iResnet~\cite{liang2018learning}, through to the computationally expensive, but state-of-the-art, GANet~\cite{Zhang2019GANet}.
Rows marked with~$\dagger$ use the publicly released model weights.
We see that our synthesized data (MfS) consistently performs better than training with the synthetic Sceneflow dataset.}
\begin{tabular}{|l|l||cc|cc|cc|cc|}
\hline
Architecture & Training data &
\multicolumn{2}{c|}{KITTI '12}  &
\multicolumn{2}{c|}{KITTI '15} &
\multicolumn{2}{c|}{Middlebury} &
\multicolumn{2}{c|}{ETH3D} \\
& & EPE & $>$3px & EPE & $>$3px & EPE & $>$2px & EPE & $>$1px \\
\hline

iResnet \cite{liang2018learning} & Sceneflow & 1.07 & 6.40 & \textbf{1.15} & 5.98 & 12.13 & 41.65 & 1.57 & 17.90 \\
iResnet \cite{liang2018learning} & MfS & \textbf{0.93} & \textbf{5.38} & \textbf{1.15} & \textbf{5.75} & \textbf{8.52} & \textbf{33.36} & \textbf{0.61} & \textbf{11.65} \\
\hline

%  Self-supervised finetune & Sceneflow + KITTI S & & & & \\
 PSMNet \cite{chang2018pyramid}$\dagger$ & Sceneflow & 6.06 & 25.38 & 6.04 & 25.07 & 17.54 & 54.60 & 1.21 & 19.52 \\
 PSMNet  \cite{chang2018pyramid}& Sceneflow & 1.04 & 5.76 & 1.19 & 5.80 & 9.42 & 34.99 & 1.25 & 13.04 \\
 PSMNet  \cite{chang2018pyramid}& MfS & \textbf{0.95} & \textbf{4.43} & \textbf{1.04} & \textbf{4.75} & \textbf{6.33} & \textbf{26.42} & \textbf{0.53} & \textbf{8.17} \\
\hline

 GANet  \cite{Zhang2019GANet}$\dagger$& Sceneflow & 1.30  & 7.97 & 1.51 & 9.52 & 11.88 & 37.34 & 0.48 & 8.19 \\
 GANet  \cite{Zhang2019GANet} & Sceneflow & 0.96 & 5.24 & 1.14 & 5.43 & 9.81 & 32.20 & 0.48 & 9.45 \\
 GANet  \cite{Zhang2019GANet}& MfS & \textbf{0.81} & \textbf{4.31} & \textbf{1.02} & \textbf{4.56} & \textbf{5.66} & \textbf{24.41} & \textbf{0.43} & \textbf{6.62} \\
\hline
\end{tabular}
\label{tab:architecture_ablation}
\vspace{-10pt}
\end{table}

\begin{table}[h]
%\vspace{-10pt}
\tablefontsize
\centering
\caption{{\bf Stereo performance improves with better monocular depths.}
Here we train PSMNet with MfS and our data synthesis approach without any dataset-specific finetuning.
Each row uses a different monocular depth network, each with a different architecture and source of supervision.
As expected, better monocular depth gives improved stereo matching, but all are competitive compared to synthetic training data.
Below each row we show the \% difference relative to the Sceneflow baseline.
}
\resizebox{1.0\textwidth}{!}{
% {\small
\begin{tabular}{|l|l||cc|cc|cc|cc|cc|}
\hline
Monocular & Monocular & \multicolumn{2}{c|}{KITTI '12}  &
\multicolumn{2}{c|}{KITTI '15} &
\multicolumn{2}{c|}{Middlebury} &
\multicolumn{2}{c|}{ETH3D} \\
model & training data & EPE & $>$3px & EPE & $>$3px & EPE & $>$2px & EPE & $>$1px \\
%D1 noc & D1 occ & EPE noc & EPE occ \\
\hline

    {DiW} \cite{nips2016single} &   Human labelling &  0.93  & 5.08  & 1.13 & 6.05 & 9.96 & 32.96 & 0.63 & 12.23 \\
  &&	\color{darkgreen}\emph{	-10.6	}&\color{darkgreen}\emph{	-11.8	}&\color{darkgreen}\emph{	-5.0	}&\color{darkred}\emph{	4.3	}&\color{darkred}\emph{	5.7	}&\color{darkgreen}\emph{	-5.8	}&\color{darkgreen}\emph{	-49.6	}&\color{darkgreen}\emph{	-6.2}
  \\
  \hline
  Monodepth2 (M) \cite{godard2019digging} & KITTI Monocular & 0.92  & 5.01  & 1.10 & 5.11 & 9.41 & 30.76 & 0.60 & 12.56 \\
  &&	\color{darkgreen}\emph{	-11.5	}&\color{darkgreen}\emph{	-13.0	}&\color{darkgreen}\emph{	-7.6	}&\color{darkgreen}\emph{	-11.9	}&\color{darkgreen}\emph{	-0.1	}&\color{darkgreen}\emph{	-12.1	}&\color{darkgreen}\emph{	-52.0	}&\color{darkgreen}\emph{	-3.7}	\\
 \hline
  Megadepth \cite{li2018megadepth} & SfM reconstructions & \textbf{0.89} & \textbf{4.29} & 1.07 & 4.96 & 7.66 & 28.86 & 0.54 & 10.41 \\
  &&	\color{darkgreen}\emph{	-14.4	}&\color{darkgreen}\emph{	-25.5	}&\color{darkgreen}\emph{-10.1	}&\color{darkgreen}\emph{-14.5	}&\color{darkgreen}\emph{	-18.7	}&\color{darkgreen}\emph{	-17.5	}&\color{darkgreen}\emph{	-56.8	}&\color{darkgreen}\emph{	-20.2	}	\\
  \hline

 {MiDaS} \cite{lasinger2019towards} &  3D Movies + others&   0.95 & 4.43 & \textbf{1.04} & \textbf{4.75} & \textbf{6.33} & \textbf{26.42} & \textbf{0.53} & \textbf{8.17} \\
 &&	\color{darkgreen}\emph{	-8.7	}&\color{darkgreen}\emph{	-23.1	}&\color{darkgreen}\emph{	-12.6	}&\color{darkgreen}\emph{	-18.1	}&\color{darkgreen}\emph{	-32.8	}&\color{darkgreen}\emph{	-24.5	}&\color{darkgreen}\emph{	-57.6	}&\color{darkgreen}\emph{	-37.3	}	\\
\hline
\hline
\multicolumn{2}{|c||}{Sceneflow baseline} & 1.04 & 5.76 & 1.19 & 5.80 & 9.42 & 34.99 & 1.25 & 13.04 \\
\hline

\end{tabular}}
\label{tab:monodepth_model_compare}
\vspace{-10pt}
\end{table}

%%%%%%%%%%%%%%%%%%%%%%%%%%%%%%%%%%%%%%%%
\subSectionSpaceBefore
\subsection{Comparing Different Monocular Depth Networks}
\subSectionSpaceAfter
The majority of our experiments are performed using monocular depth from {MiDaS}~\cite{lasinger2019towards}, which itself is trained with a variety of different datasets \eg flow derived pseudo ground truth and structured light depth.
In Table~\ref{tab:monodepth_model_compare} we show that our approach still produces competitive stereo performance when we use alternative monocular networks that have been trained using much \emph{weaker} supervision.
This includes using monocular video only self-supervision on the KITTI dataset (Monodepth2 (M)~\cite{godard2019digging}), pairwise ordinal human annotations (DiW~\cite{nips2016single}) and multi-view image collections (Megadepth~\cite{li2018megadepth}).
It is worth noting that even though Megadepth~\cite{li2018megadepth} training  depth is computed from unstructured image collections, without using any trained correspondence matching, using it to synthesize training data still outperforms the synthetic Sceneflow baseline.
This is likely due to the fact that stereo matching is fundamentally different from monocular depth estimation, and even potentially unrealistic disparity information can help the stereo network learn how to better match pixels.
While the recent {MiDaS}~\cite{lasinger2019towards} performs best overall, we observe a consistent performance increase in stereo estimation with better monocular depth.
This is very encouraging, as it indicates that our performance may only get better in future as monocular depth networks improve.
We also show qualitatively in Fig.~\ref{fig:we_fix_mono} how our stereo predictions can recover from errors made by monocular models.

%%%%%%%%%%%%%%%%%%%%%%%%%%%%%%%%%%%%%%%%
\subSectionSpaceBefore
\subsection{Ablating Components of Our Method}
\subSectionSpaceAfter

In Table~\ref{tab:ablation} we train stereo networks with data synthesized with and without depth sharpening and background texture filling (\ie$I_b$).
For the case where we do not use background texture filling we simply leave black pixels.
We observe that disabling these components results in worse stereo performance compared to the full pipeline, with Fig.~\ref{fig:depth_sharpening} demonstrating the qualitative impact of depth sharpening.
This highlights the importance of realistic image synthesis.
A more sophisticated warping approach \cite{niklaus2020softmax} or inpainting with a neural network \cite{yu2019free} should increase the realism and is left as future work.
We instead opt for a simple warping and background filling approach for computational efficiency, which allows us to generate new views online during training.

\begin{table}[t]
\tablefontsize
\centering
\vspace{-10pt}
\caption{{\bf Depth sharpening and background filling helps.} Disabling depth sharpening and background filling hurts PSMNet stereo performance when trained on MfS.}
\begin{tabular}{|c|c||cc|cc|cc|cc|}
\hline
  & & \multicolumn{2}{c|}{KITTI '12}  &
\multicolumn{2}{c|}{KITTI '15} &
\multicolumn{2}{c|}{Middlebury} &
\multicolumn{2}{c|}{ETH3D} \\
 Sharpening & Background filling & EPE & $>$3px & EPE & $>$3px & EPE & $>$2px & EPE & $>$1px \\
\hline
  \xmark & \cmark & 1.02 & 4.75 & 1.06 & 4.76 & 7.57 & 28.16 & \textbf{0.47} & 8.21 \\
  \cmark  & \xmark & 0.96 & 4.81 & 1.10 & 5.27 & 7.11 & 27.22 & 0.68 & 8.70 \\
 \cmark & \cmark  & \textbf{0.95} & \textbf{4.43} & \textbf{1.04} & \textbf{4.75} & \textbf{6.33} & \textbf{26.42} & 0.53 & \textbf{8.17} \\
 \hline
\end{tabular}
 \vspace{-10pt}
\label{tab:ablation}
\end{table}

%%%%%%%%%%%%%%%%%%%%%%%%%%%%%%%%%%%%%%%%
\subSectionSpaceBefore
\subsection{Adapting to the Target Domain}
\subSectionSpaceAfter
\label{sec:kitti_finetuning}
Our method allows us to train a generalizable stereo network which outperforms baselines.
However, in cases where we have data from the target domain, \eg in the form of single unpaired images, stereo pairs, or stereo pairs with ground truth LiDAR.
We compare adaptation with each of these data sources in Table~\ref{tab:kitti_finetuning}.
First, we train on stereo pairs generated from the \textbf{KITTI left images} with our method, using the 29K KITTI images from the 33 scenes not included in the KITTI 2015 stereo  training set and monocular predictions from \cite{lasinger2019towards}.
For \textbf{KITTI self-supervised} finetuning, we follow the method proposed in \cite{guo2018learning} using the stereo image pairs from the same 33 scenes.
Finally, for \textbf{KITTI LiDAR} we finetune on 160 training images and point clouds (following \cite{chang2018pyramid}) for 200 epochs.
We give results for unoccluded pixels (\emph{Noc}) and all pixels (\emph{All}); see \cite{chang2018pyramid} for more details of the metrics.
We outperform Sceneflow pretraining in all cases.

\begin{table}[!b]
\tablefontsize
\centering
\vspace{-10pt}
\caption{{\bf KITTI 2015 finetuning.} All trained by us on PSMNet.
Our data generation approach, when finetuned on KITTI LiDAR data (last  row), results in superior performance when compared to Sceneflow pretraining and KITTI finetuning ($4^{th}$ row).
$\dagger$Sceneflow models cannot be finetuned on single KITTI images without our method. }
{\begin{tabular}{|p{2.4cm}|p{3.2cm}||c|c|c|c|}
\hline
\small
%   & & KITTI '12 & KITTI '15  & Midd.  & ETH3D  \\
Pretraining & Finetuning method & EPE Noc & $>$3px Noc & EPE All & $>$3px All  \\%$>$3px & $>$3px & $>$2px  & $>$1px \\
\hline
Sceneflow & None & 1.18 & 5.62 & 1.19 & 5.80 \\
Sceneflow & KITTI left images & $\dagger$ & $\dagger$ & $\dagger$ & $\dagger$ \\
Sceneflow & KITTI self-supervised & 1.03 & 4.90 & 1.05 & 5.07 \\
Sceneflow & KITTI LiDAR  & 0.73 & 2.21 & 0.74 & 2.36 \\
\hline
Ours with MfS & None & 1.02 & 4.57 & 1.04 & 4.75 \\
Ours with MfS & KITTI left images & 1.00 & 4.44 & 1.01 & 4.58 \\
Ours with MfS & KITTI self-supervised & 1.00 & 4.57 & 1.01 & 4.71 \\
Ours with MfS  & KITTI LiDAR  & 0.68 & 2.05 & 0.69 & 2.15 \\
\hline
\end{tabular}}
\vspace{-10pt}
\label{tab:kitti_finetuning}
\end{table}

%%%%%%%%%%%%%%%%%%%%%%%%%%%%%%%%%%%%%%%%
\subSectionSpaceBefore
\subsection{Varying the Amount of Training Data}
\subSectionSpaceAfter
In Table~\ref{tab:dataset_size_compare} we train a PSMNet stereo network by varying the amount of training images from our MfS dataset while keeping the monocular network (MiDaS~\cite{lasinger2019towards}) fixed.
We observe that as the number of training images increases so does stereo performance.
Even in the smallest data regime (35K images), which is comparable in size to Sceneflow~\cite{mayer2016sceneflow}, we still outperform training on the synthetic data alone in all metrics.
This indicates that it is not just the power of the monocular network that drives our performance, but the large and varied image set that we make available to stereo training.

\begin{table}[!t]
\tablefontsize
\centering
\vspace{-10pt}
\caption{{\bf Performance improves with more synthesized images.}
Here we train PSMNet with varying numbers of images from our MfS dataset using our synthesis approach, without any finetuning to the test datasets.
As expected, more training data improves the stereo network, even outperforming synthetic data.
Note that all models were trained for 250K steps.
Below each results row we show the \% difference relative to the Sceneflow baseline.
}
\resizebox{1.0\textwidth}{!}{
\begin{tabular}{|l|l|l||cc|cc|cc|cc|cc|}
\hline
Monocular & Dataset & \# of images & \multicolumn{2}{c|}{KITTI '12}  &
\multicolumn{2}{c|}{KITTI '15} &
\multicolumn{2}{c|}{Middlebury} &
\multicolumn{2}{c|}{ETH3D} \\
 model &  & & EPE & $>$3px & EPE & $>$3px & EPE & $>$2px & EPE & $>$1px \\
%D1 noc & D1 occ & EPE noc & EPE occ \\

\hline
 Megadepth \cite{li2018megadepth} & MfS  & 35,000 & 0.88  & 4.47  & 1.25 & 5.52 & 9.31 & 28.13 & 0.59 & 12.24 \\
 &&&	\color{darkgreen}\emph{	-15.4	}&\color{darkgreen}\emph{	-22.4	}&\color{darkgreen}\emph{	-3.4	}&\color{darkgreen}\emph{	-4.8	}&\color{darkgreen}\emph{	-1.2	}&\color{darkgreen}\emph{	-19.6	}&\color{darkgreen}\emph{	-52.8	}&\color{darkgreen}\emph{	-6.1	}	\\
 \hline
 Megadepth \cite{li2018megadepth} & MfS & 500,000 & 0.89 & \textbf{4.29} & 1.07 & 4.96 & 7.66 & 28.86 & 0.54 & 10.41 \\
  &&&	\color{darkgreen}\emph{	-14.4	}&\color{darkgreen}\emph{	-25.5	}&\color{darkgreen}\emph{-10.1	}&\color{darkgreen}\emph{-14.5	}&\color{darkgreen}\emph{	-18.7	}&\color{darkgreen}\emph{	-17.5	}&\color{darkgreen}\emph{	-56.8	}&\color{darkgreen}\emph{	-20.2	}	\\
 \hline
 {MiDaS} \cite{lasinger2019towards} & MfS  & 35,000 &   \textbf{0.87} & 4.35 & 1.09 & 5.06 & 6.77 & 28.86 & 0.56 & 8.18 \\
  &&&	\color{darkgreen}\emph{	-16.3	}&\color{darkgreen}\emph{	-24.5	}&\color{darkgreen}\emph{	-8.4	}&\color{darkgreen}\emph{	-12.8	}&\color{darkgreen}\emph{	-28.1	}&\color{darkgreen}\emph{	-17.5	}&\color{darkgreen}\emph{	-55.2	}&\color{darkgreen}\emph{	-29.6	}	\\
 \hline
 {MiDaS} \cite{lasinger2019towards} & MfS  & 500,000 &   0.95 & 4.43 & \textbf{1.04} & \textbf{4.75} & \textbf{6.33} & \textbf{26.42} & \textbf{0.53} & \textbf{8.17} \\
  &&&	\color{darkgreen}\emph{	-8.7	}&\color{darkgreen}\emph{	-23.1	}&\color{darkgreen}\emph{	-12.6	}&\color{darkgreen}\emph{	-18.1	}&\color{darkgreen}\emph{	-32.8	}&\color{darkgreen}\emph{	-24.5	}&\color{darkgreen}\emph{	-57.6	}&\color{darkgreen}\emph{	-37.3	}	\\
 \hline
 \hline
\multicolumn{2}{|c||}{Sceneflow baseline} & 35,454 & 1.04 & 5.76 & 1.19 & 5.80 & 9.42 & 34.99 & 1.25 & 13.04 \\
\hline
\end{tabular}}
\label{tab:dataset_size_compare}
\vspace{-10pt}
\end{table}

%%%%%%%%%%%%%%%%%%%%%%%%%%%%%%%%%%%
\sectionSpaceBefore
\section{Discussion}
\sectionSpaceAfter
%%%%%%%%%%%%%%%%%%%%%%%%%%%%%%%%%%%

Many of our results used the monocular depth estimation network from \cite{lasinger2019towards}, which was trained on a mixture of datasets including stereo pairs, which in turn had their depth estimated using an optical flow network trained on synthetic data~\cite{ilg2017flownet}.
However, experiments in Table~\ref{tab:monodepth_model_compare} show we perform well regardless of the source of data used to train the underlying monocular estimator, with good predictions with depth estimation networks trained on SfM reconstructions \cite{li2018megadepth} or monocular video sequences \cite{godard2019digging}.
In fact, even the naive `disparity estimation' methods in Table~\ref{tab:we_beat_alternative_data_sources} generate reasonable stereo training data.
Furthermore, we observe in Table~\ref{tab:dataset_size_compare} that as we increase the amount of training images our stereo performance improves.
This indicates the power of the new additional training data that our approach makes available for stereo training.

%%%%%%%%%%%%%%%%%%%%%%%%%%%%%%%%%%%
\sectionSpaceBefore
\section{Conclusion}
\sectionSpaceAfter
%%%%%%%%%%%%%%%%%%%%%%%%%%%%%%%%%%%
We presented a fully automatic pipeline for synthesizing stereo training data from single monocular images.
We showed that our approach is robust to the source of monocular depth and results in a significant increase in test-time generalization performance for multiple different stereo algorithms.
Importantly, our approach enables the possibility of converting any monocular image into stereo training data, opening up the door to using existing large image collections to improve stereo estimation.
In future we intend to explore different ways to automatically estimate the quality of our synthesized data~\cite{Chen_2019_CVPR} and approaches for performing additional augmentations during training~\cite{cubuk2019autoaugment}.
%Another interesting question is if we can use our improved stereo networks to train better monocular depth networks, \eg by providing sparse depth cues~\cite{klodt2018supervising,watson-2019-depth-hints,tosi2019learning,Aleotti2020}, which in turn could be used to further improve stereo with our method.
Another interesting question is if we can use our improved stereo networks to train better monocular depth networks, \eg by providing sparse depth cues~\cite{klodt2018supervising,watson-2019-depth-hints,tosi2019learning}, which in turn could be used to further improve stereo with our method.

\paragraph{Acknowledgements}
Large thanks to Aron Monszpart for help with baseline code, to Grace Tsai for feedback, and Sara Vicente for suggestions for baseline implementations.
Also thanks to the authors who shared models and images.

\begin{figure}[h]
\centering
\newcommand{\itemwidth}{0.3\columnwidth}
\centering
%images are 347 632 222 and 474
\begin{tabular}{@{\hskip 2mm}c@{\hskip 2mm}c@{\hskip 2mm}c@{\hskip 2mm}c@{}}
& {\scriptsize Input (left) image} & {\scriptsize Trained with Sceneflow} & {\scriptsize Trained with Ours} \\
\raisebox{5pt}{\multirow{2}{*}{\rotatebox{90}{\centering\hspace{4mm}\scriptsize KITTI '15}}} &
\includegraphics[width=\itemwidth]{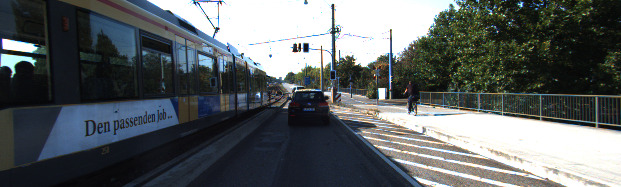} &
\includegraphics[width=\itemwidth]{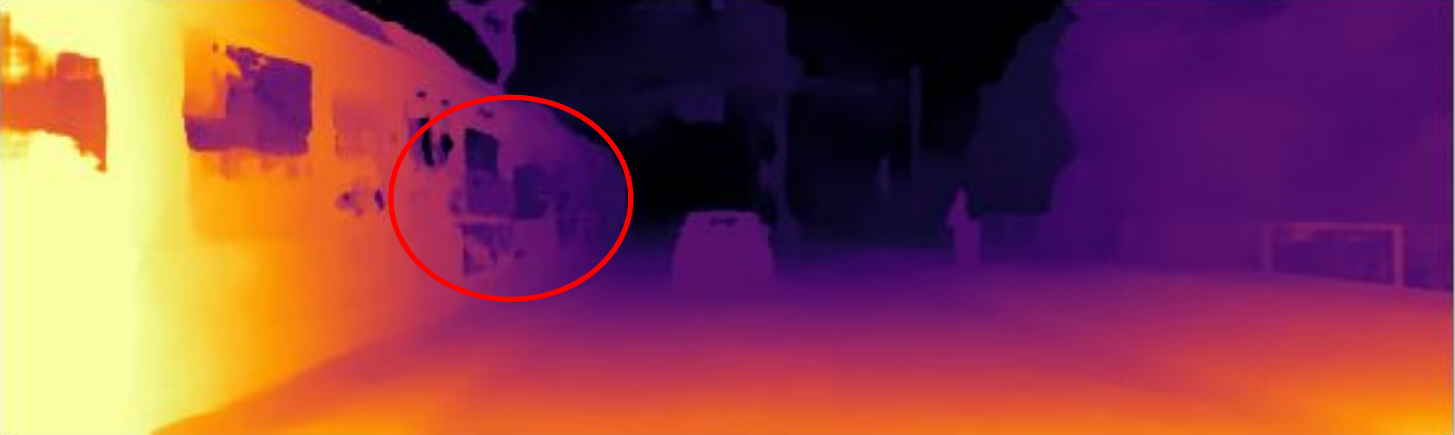} &
\includegraphics[width=\itemwidth]{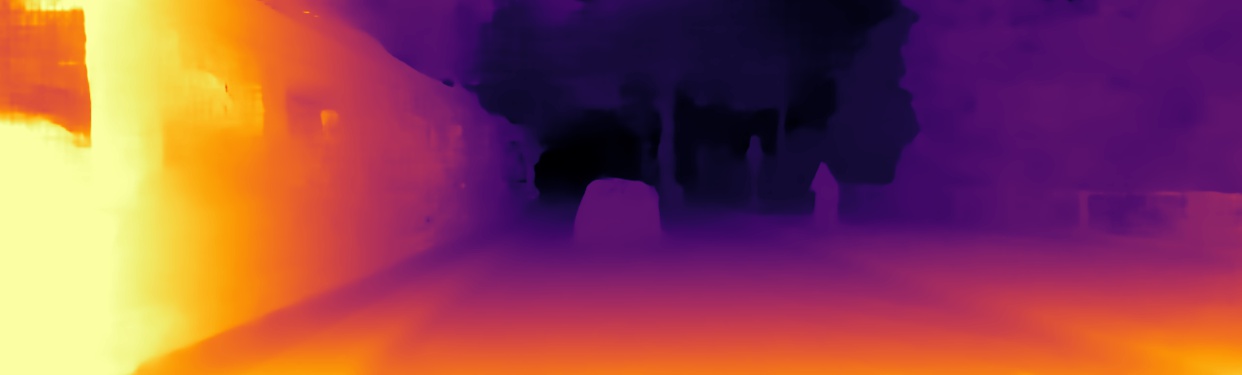} \\
&
\includegraphics[width=\itemwidth]{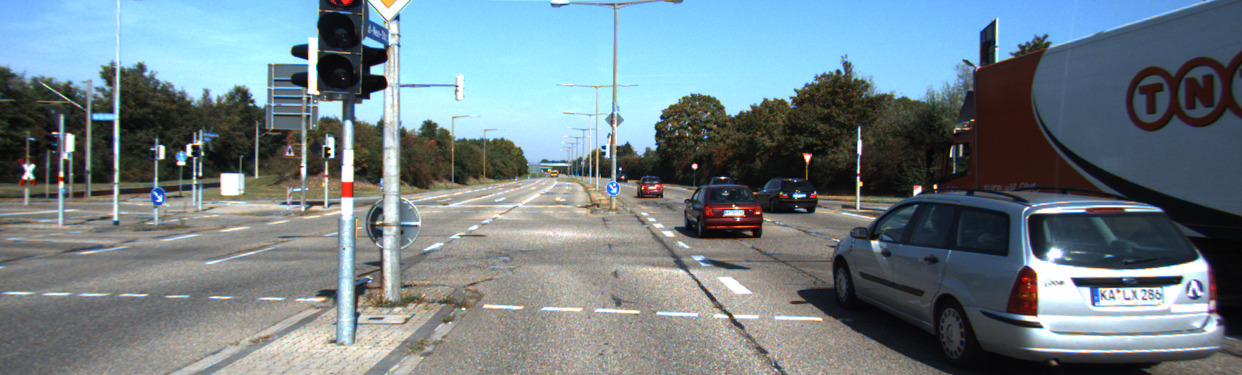} &
\includegraphics[width=\itemwidth]{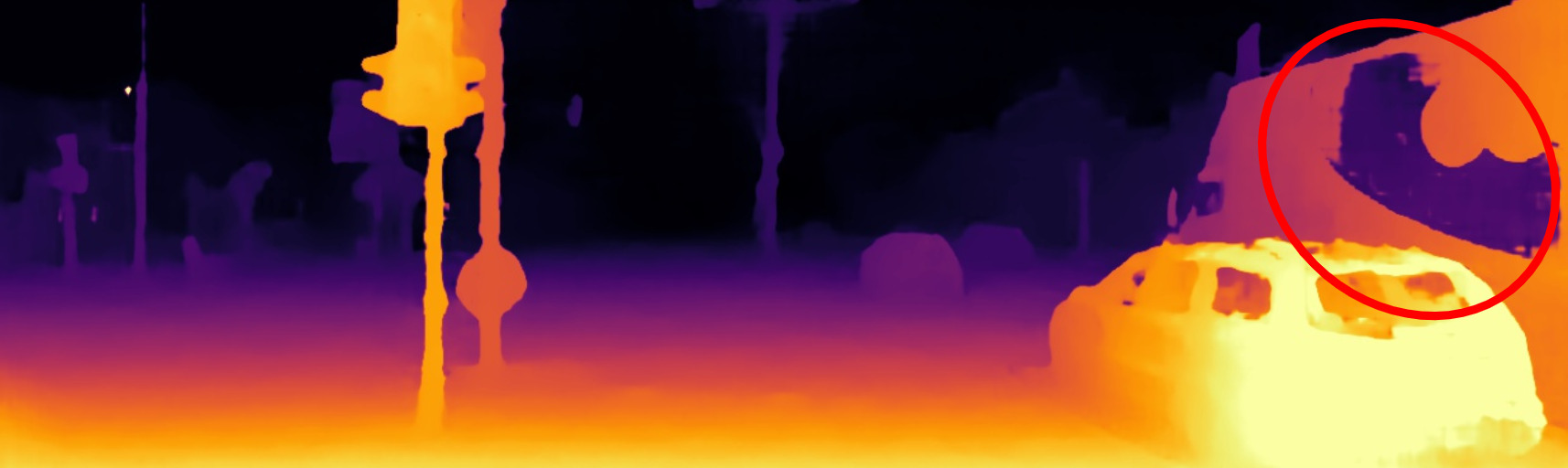} &
\includegraphics[width=\itemwidth]{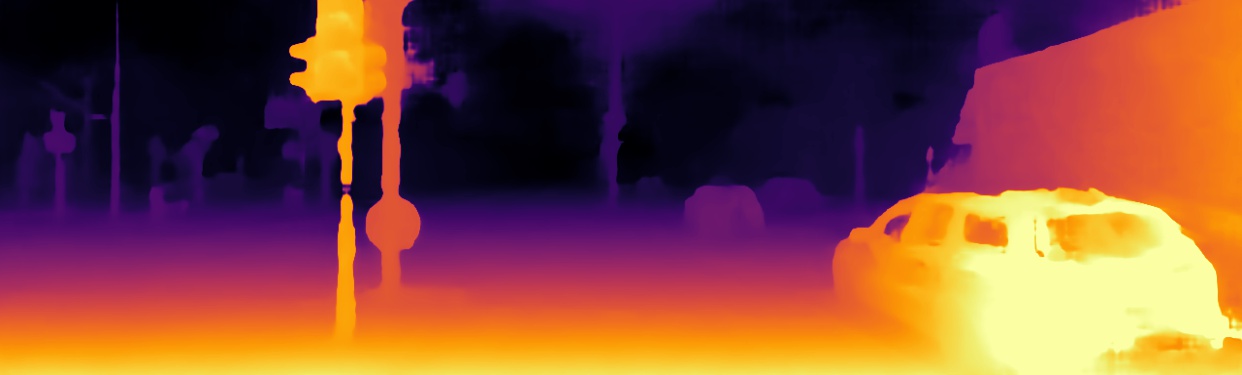} \\

\raisebox{5pt}{\rotatebox{90}{\hspace{4mm}\scriptsize  Middlebury}} &
\includegraphics[width=\itemwidth]{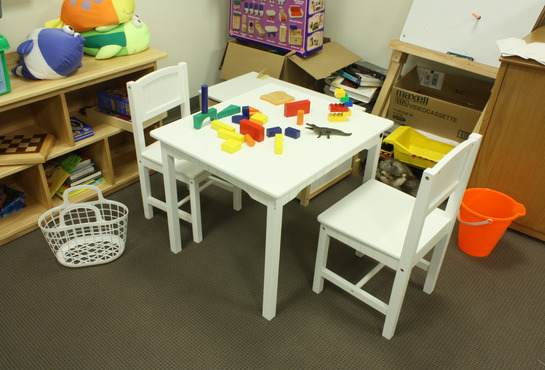} &
\includegraphics[width=\itemwidth]{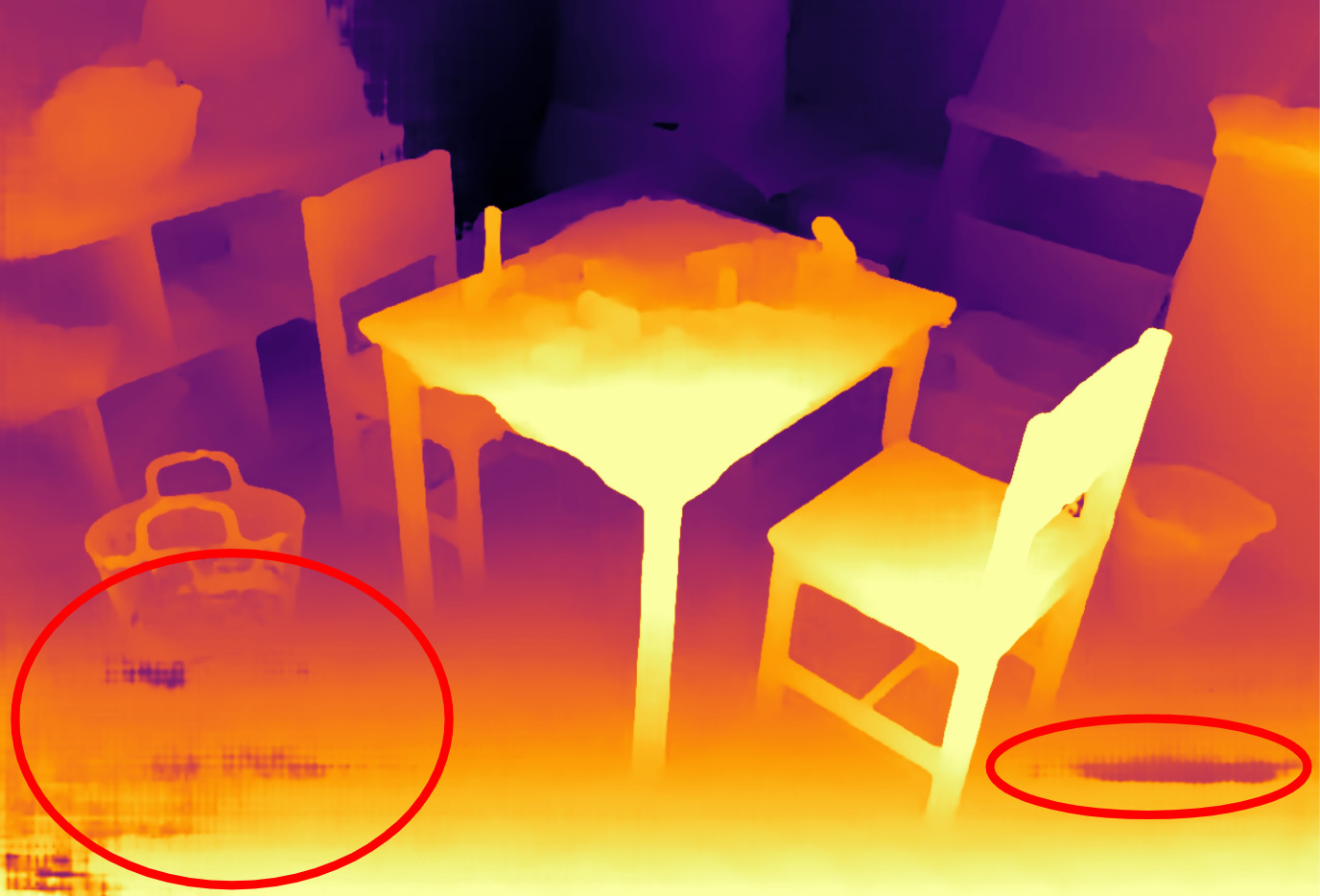} &
\includegraphics[width=\itemwidth]{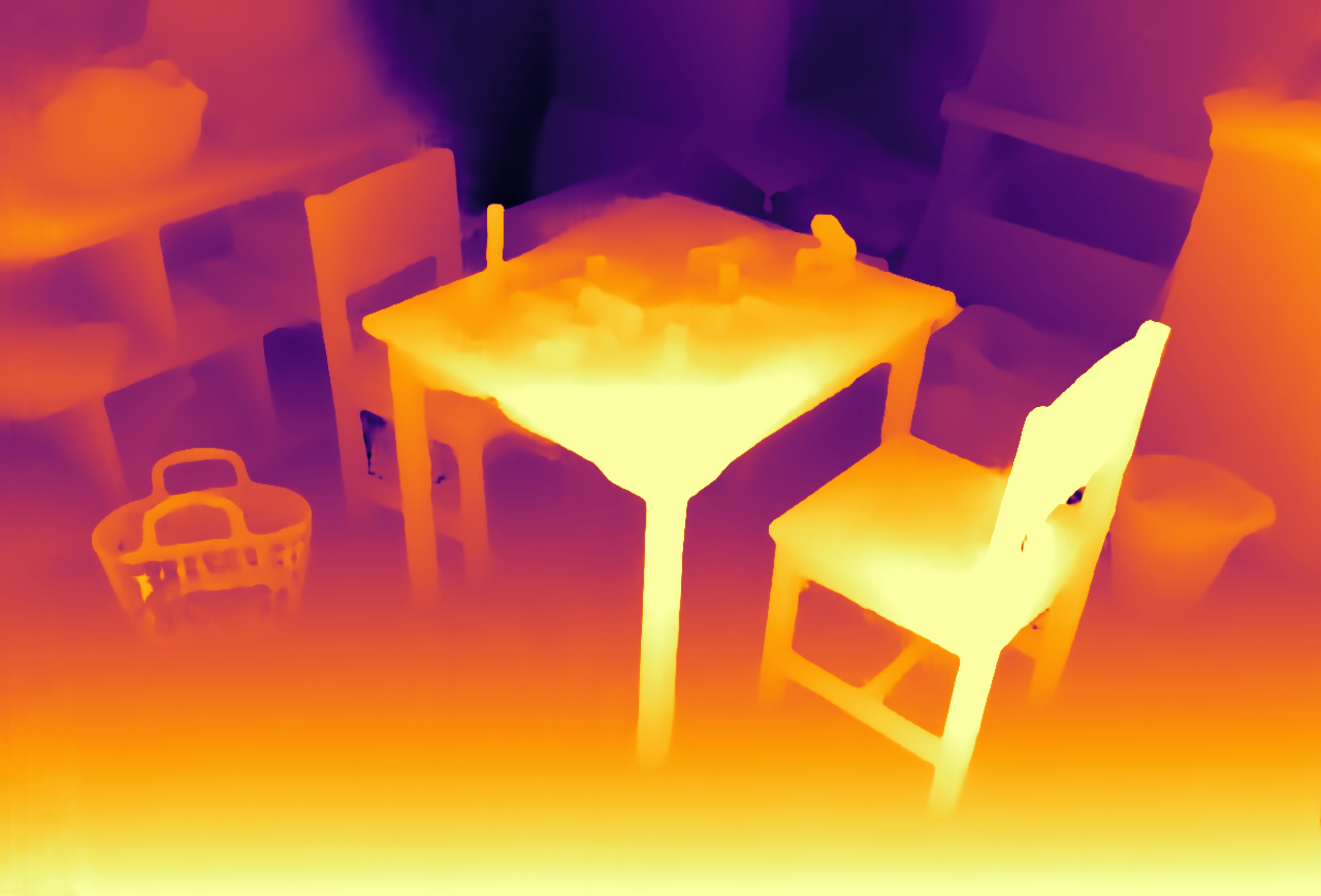} \\

\raisebox{40pt}{\rotatebox{90}{\hspace{4mm}\scriptsize  Flickr1024}} &
\includegraphics[trim=0 30 0 35,clip,width=\itemwidth]{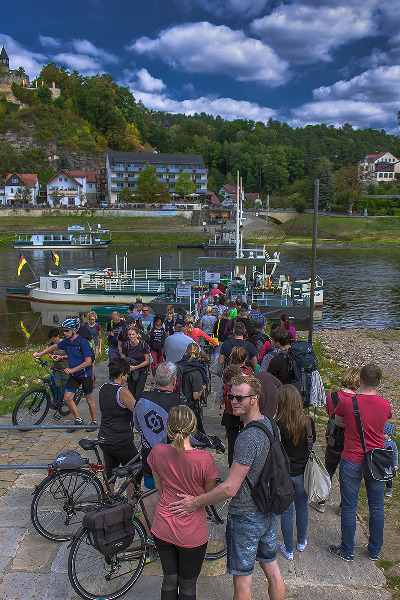} &
\includegraphics[trim=0 10 0 30,clip,width=\itemwidth]{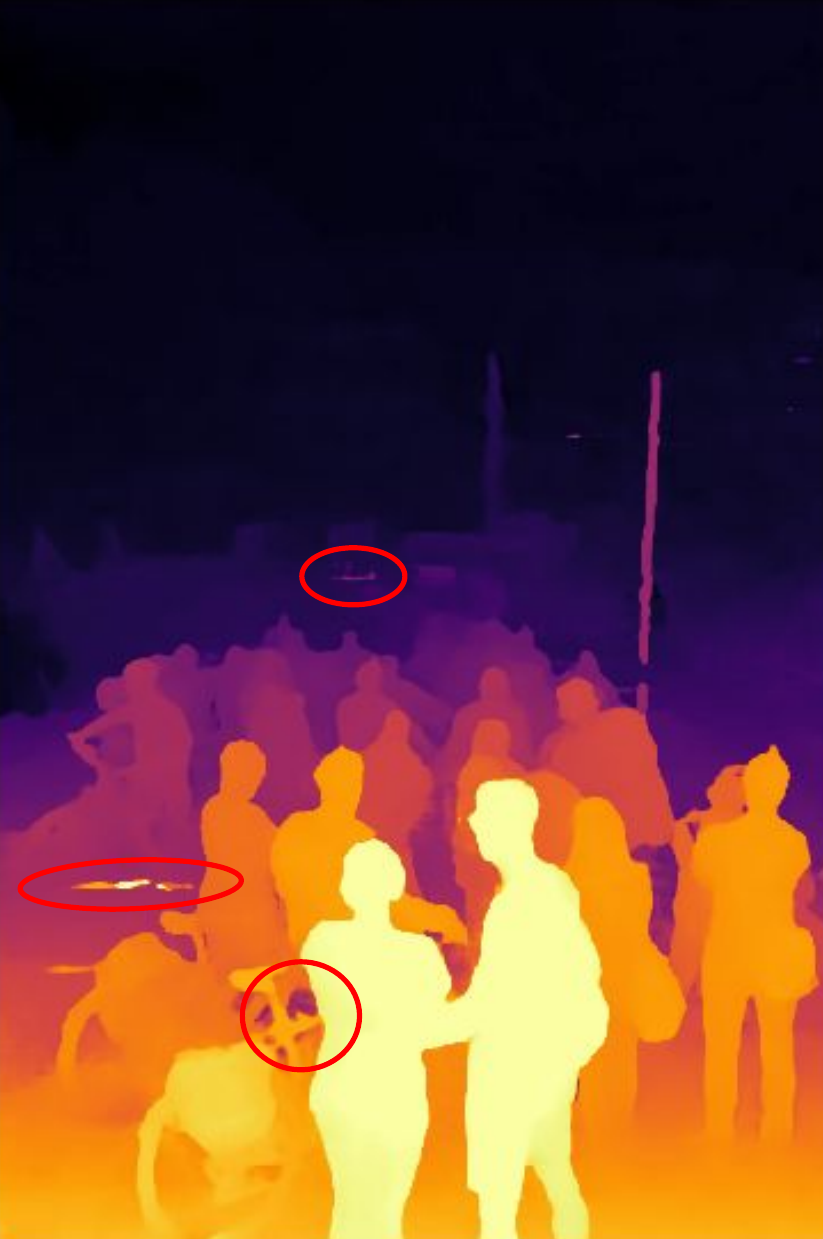} &
\includegraphics[trim=0 30 0 60,clip,width=\itemwidth]{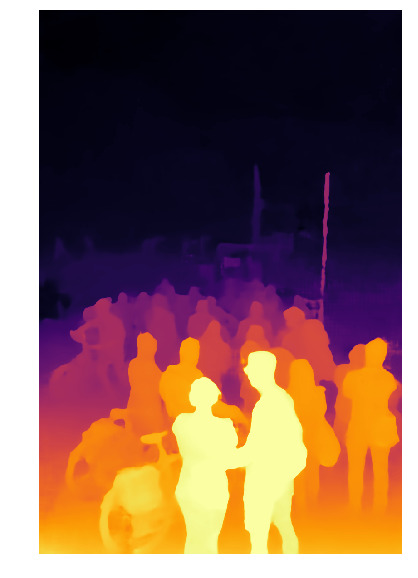}  \\

\raisebox{10pt}{\rotatebox{90}{\hspace{4mm}\scriptsize  ETH3D}} &
\includegraphics[width=\itemwidth]{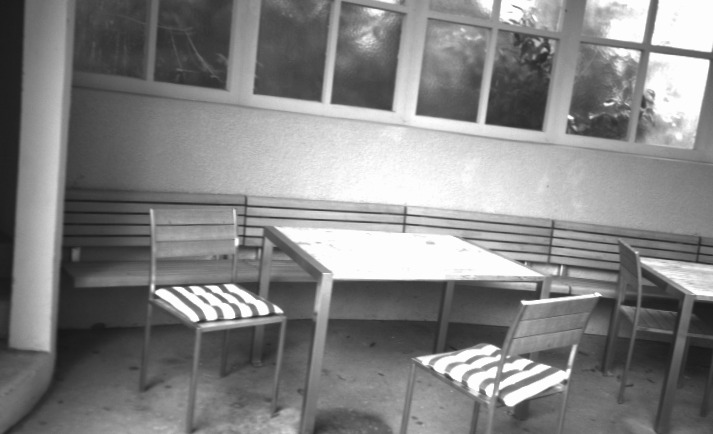} &
\includegraphics[width=\itemwidth]{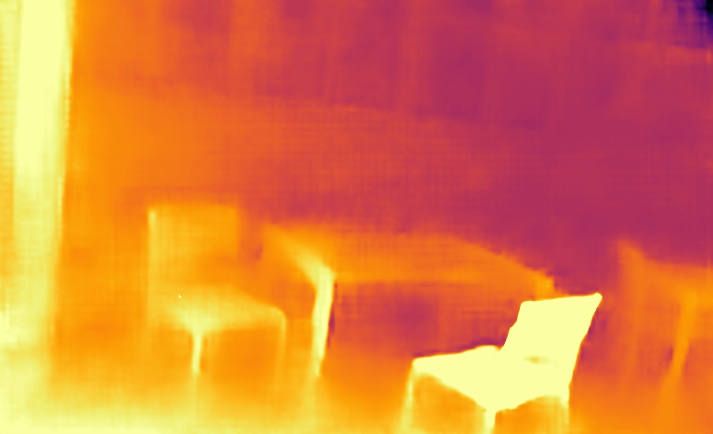} &
\includegraphics[width=\itemwidth]{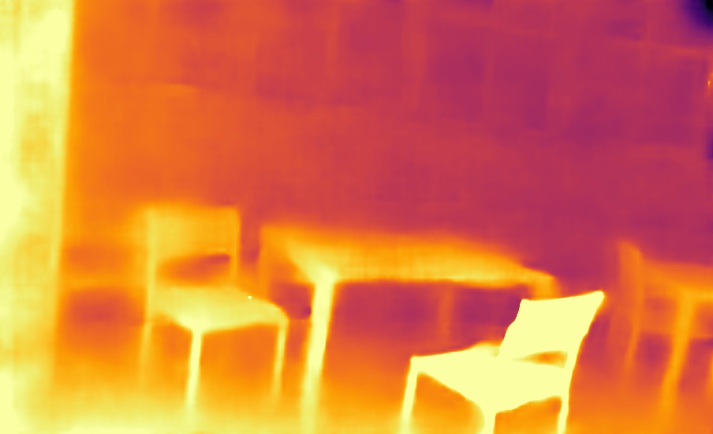} \\

\end{tabular}
% \includegraphics[height=5cm]{example-image-a}
%\vspace{-10pt}
\caption{{\bf Qualitative Results.} Comparing our synthetic data generation approach with Sceneflow, using a PSMNet model trained on both. Our predictions benefit from semantically plausible depths at training time and have fewer artifacts.
}
%\vspace{-10pt}
\label{fig:qualitative_comparisons}
\end{figure}

\clearpage

\bibliographystyle{splncs04}
\bibliography{main}

\clearpage

\renewenvironment{description}[1][0pt]
  {\list{}{\labelwidth=0pt \leftmargin=#1
   \let\makelabel\descriptionlabel}}
  {\endlist}

\section*{\huge Supplementary Material}
%%%%%%%%%%%%%%%%%%%%%%%%%%%%%%%%%%%%
\section{KITTI Test Server Evaluation}
%%%%%%%%%%%%%%%%%%%%%%%%%%%%%%%%%%%%
Figs.~\ref{fig:qual_benchmark_figure} and \ref{fig:qual_benchmark_figure_cont} show the complete set of available qualitative results from the held-out 2015 test set from the KITTI online server.
We show the official benchmark model of (\textbf{GANet official})\footnote{\url{http://www.cvlibs.net/datasets/kitti/eval_scene_flow_detail.php?benchmark=stereo&result=ccb2b24d3e08ec968368f85a4eeab8b668e70b8c}} \cite{Zhang2019GANet},  a GANet retrained by us using only Sceneflow~ \cite{mayer2016sceneflow} (\textbf{Sceneflow GANet}), and GANet trained only with our MfS data (\textbf{MfS GANet (Ours)}).
Note that \textbf{GANet official} uses KITTI data to finetune, as is apparent in the overly smooth predictions on transparent surfaces like car windows (see rows 6 and 7 in Fig.~\ref{fig:qual_benchmark_figure}).
In Fig.~\ref{fig:qual_benchmark_figure_cont} (also rows 6 and 7), can see poor quality predictions in the right of the sky region for the finetuned model, whereas our MfS trained model produces much more sensible predictions.
Our results are generally more faithful to the boundaries of the color input image and have fewer artefacts than the alternative methods despite using no LiDAR or synthetic data during training.
Quantitative results are presented in Table~\ref{tab:kitti_finetuning_sup}.

\newcolumntype{C}[1]{>{\centering\let\newline\\\arraybackslash\hspace{0pt}}m{#1}}

\begin{figure}
    \centering
    \newcommand{\itemwidth}{0.24\columnwidth}
    \centering
    \vspace{10pt}
    \begin{tabular}{C{0.25\columnwidth}C{0.25\columnwidth}C{0.25\columnwidth}C{0.25\columnwidth}}
        \input{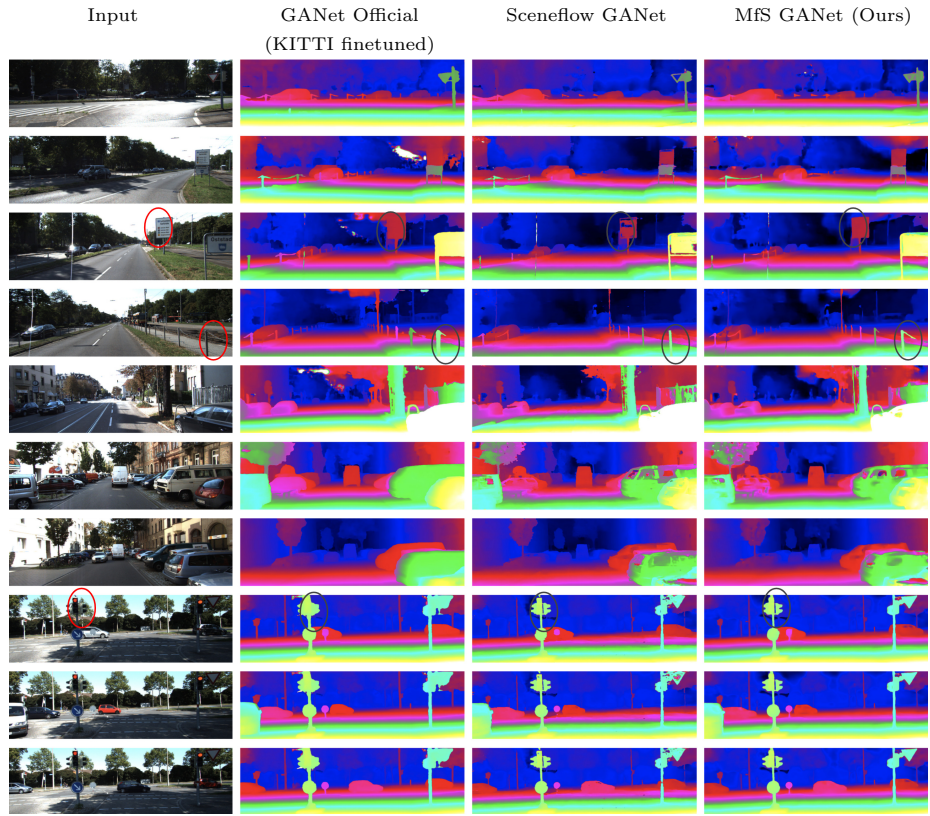}
    \end{tabular}
    \caption{{\bf KITTI 2015 benchmark qualitative comparison. } On the held-out KITTI 2015 benchmark images our predictions, MfS GANet (Ours), generated without any KITTI finetuning, are qualitatively more faithful to the scene geometry than both the Sceneflow-trained model and the official KITTI-finetund GANet~\cite{Zhang2019GANet}. These visualizations are generated by the KITTI upload server.
    \cite{kitti_server}.
    %\url{http://www.cvlibs.net/datasets/kitti/eval_scene_flow.php?benchmark=stereo}.
    }
    \label{fig:qual_benchmark_figure}
\end{figure}

\begin{figure}
    \centering
    \newcommand{\itemwidth}{0.24\columnwidth}
    \centering
    \vspace{10pt}
    \begin{tabular}{C{0.25\columnwidth}C{0.25\columnwidth}C{0.25\columnwidth}C{0.25\columnwidth}}
        \input{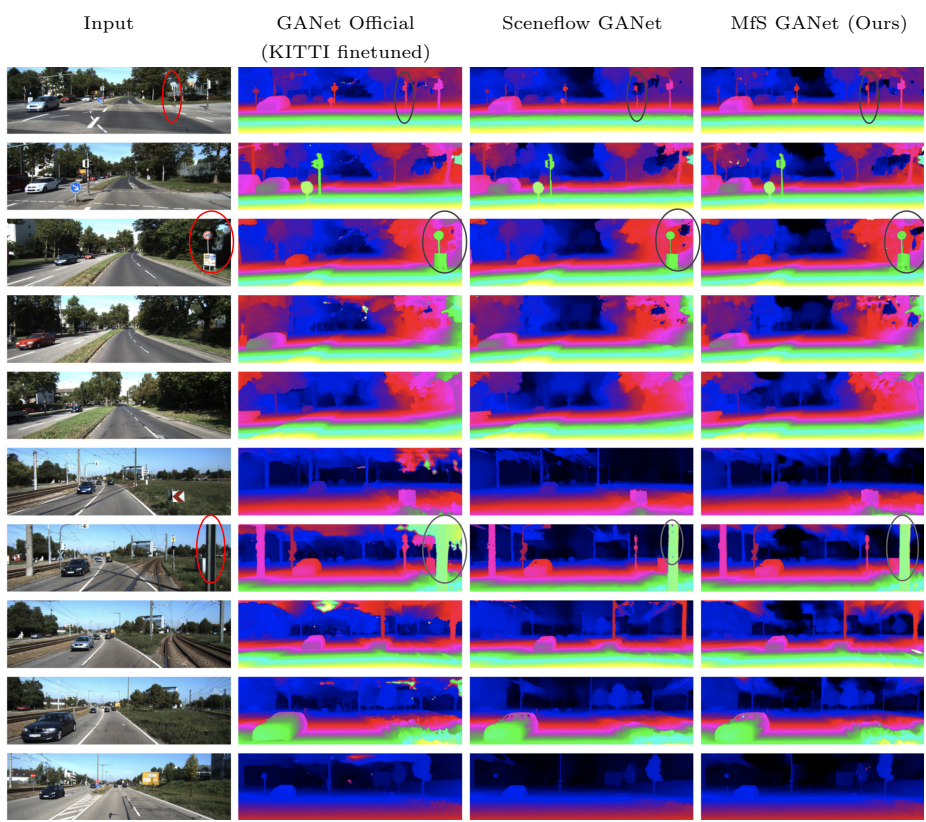}
    \end{tabular}
    \caption{{\bf KITTI 2015 benchmark qualitative comparison -- continued}. See Fig.~\ref{fig:qual_benchmark_figure} caption for details.}
    \label{fig:qual_benchmark_figure_cont}
\end{figure}

\begin{table}
\footnotesize
\centering
\vspace{-5pt}
\caption{{\bf KITTI 2015 Benchmark.} %All trained by us on GANet~\cite{Zhang2019GANet}, without any finetuning with KITTI LiDAR data.
Our model performs better than the Sceneflow trained alternative.
Row three gives the GANet~\cite{Zhang2019GANet} scores (after KITTI LiDAR finetuning) on the same benchmark as reported in their paper \cite{Zhang2019GANet}, and row four gives the current best GANet score on the KITTI benchmark leaderboard.
Our scores are not competitive with these scores from models which have had domain-specific KITTI LiDAR finetuning, but our qualitative results are more faithful (see Figs.~\ref{fig:qual_benchmark_figure} and \ref{fig:qual_benchmark_figure_cont}).}

{\begin{tabular}{|p{4.0cm}||c|c|c|c|c|c|c|}
\hline
\small
Model & KITTI  & D1-bg  & D1-fg  & D1-all  & D1-bg  & D1-fg  & D1-all  \\
& Finetune&  Noc & Noc &  Noc & All & All & All \\
\hline
\textbf{MfS GANet (Ours)}  & & \textbf{2.96}  & \textbf{15.09} & \textbf{4.97} & \textbf{3.13} & \textbf{15.57} & \textbf{5.20} \\
\hline
\textbf{Sceneflow GANet} &  & 3.60 & 16.56 & 5.74 & 3.86 & 17.21  & 6.08 \\
\hline
\hline
\textbf{GANet (in paper)} & \checkmark & & 3.39 &  1.84 & & 3.91 &  2.03 \\
\textbf{GANet Official} & \checkmark & 1.34	& 3.11	 & 1.63 & 1.48	& 3.46	& 1.81 \\

% Noc / All
\hline

\end{tabular}}
\label{tab:kitti_finetuning_sup}
\end{table}

%%%%%%%%%%%%%%%%%%%%%%%%%%%%%%%%%%%%
\section{Image Synthesis Visualisation}
\begin{figure}
\captionsetup[subfigure]{font=footnotesize,labelfont=footnotesize}
\newcommand{\itemwidth}{0.8\textwidth}
\centering
  \begin{subfigure}[b]{0.32\linewidth}
    \centering\includegraphics[width=\itemwidth]{figures/varying_s/varying_s_input.jpg}\vspace{-5pt}
    \caption{Input (left) image}
  \end{subfigure}
  \begin{subfigure}[b]{0.32\linewidth}
    \centering\includegraphics[width=\itemwidth]{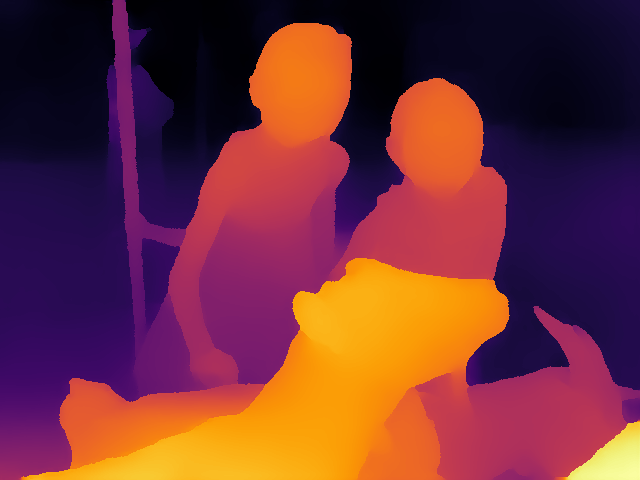}\vspace{-5pt}
    \caption{Mono~estimation~\cite{lasinger2019towards}}
  \end{subfigure} \\
  \begin{subfigure}[b]{0.32\linewidth}
    \centering\includegraphics[width=\itemwidth]{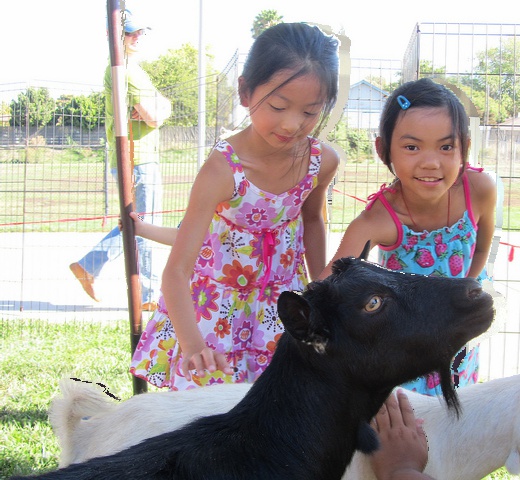}\vspace{-5pt}
    \caption{$\tilde{I}_r (s = 20)$}
  \end{subfigure}
  \begin{subfigure}[b]{0.32\linewidth}
    \centering\includegraphics[width=\itemwidth]{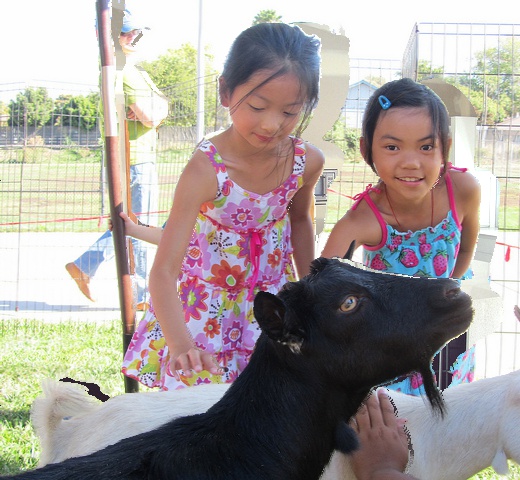}\vspace{-5pt}
    \caption{$\tilde{I}_r (s = 50)$}
  \end{subfigure}
  \begin{subfigure}[b]{0.32\linewidth}
    \centering\includegraphics[width=\itemwidth]{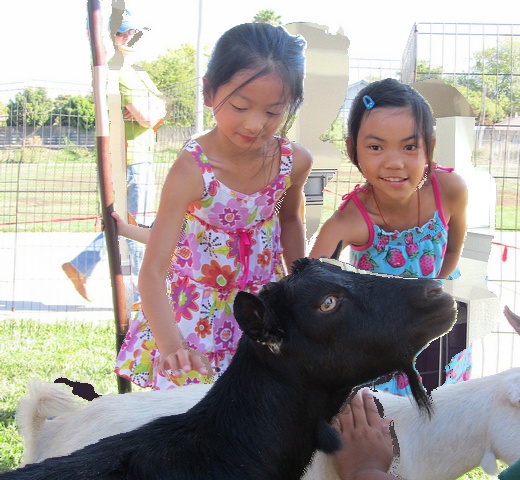}\vspace{-5pt}
    \caption{$\tilde{I}_r (s = 75)$}
  \end{subfigure}
  \begin{subfigure}[b]{0.32\linewidth}
    \centering\includegraphics[width=\itemwidth]{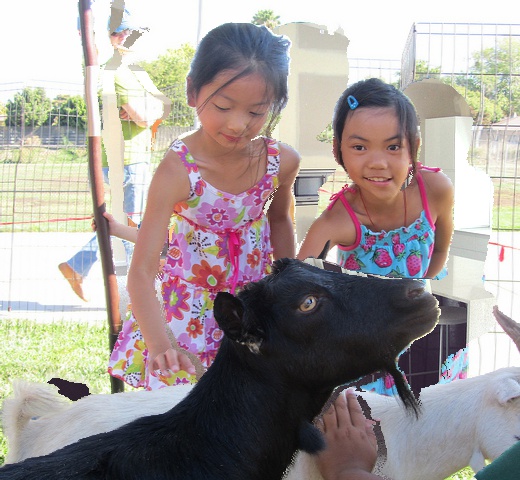}\vspace{-5pt}
    \caption{$\tilde{I}_r (s = 100)$}
  \end{subfigure}
  \begin{subfigure}[b]{0.32\linewidth}
    \centering\includegraphics[width=\itemwidth]{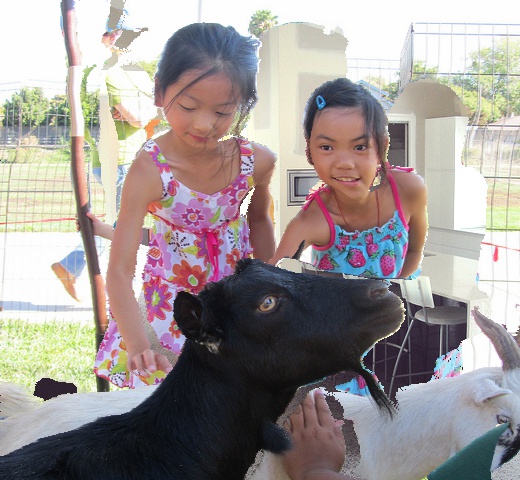}\vspace{-5pt}
    \caption{$\tilde{I}_r (s = 125)$}
  \end{subfigure}
  \begin{subfigure}[b]{0.32\linewidth}
    \centering\includegraphics[width=\itemwidth]{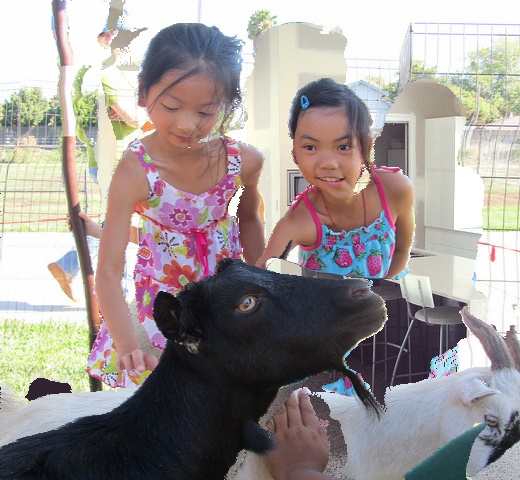}\vspace{-5pt}
    \caption{$\tilde{I}_r (s = 150)$}
  \end{subfigure}

  \vspace{-5pt}
\caption{{\bf Visualization of different scaling factors $s$.} Here we show the effect of varying the scaling factor $s$ (c - h) when constructing the synthesized right image $\tilde{I}_r$ from the input left image (a) and its predicted depth (b).}
\label{fig:image_synthesis_example}
\end{figure}

%%%%%%%%%%%%%%%%%%%%%%%%%%%%%%%%%%%%
When converting predicted depth $Z$ to disparity $D$ we can vary the scaling factor $s$, \ie $\tilde{D}=\frac{s \, Z_{\max}}{Z}$.
In Fig.~\ref{fig:image_synthesis_example} we illustrate the resulting synthesized right image $\tilde{I}_r$ for different values of $s$.

As $s$ becomes larger, we see that it has the effect of changing the relative camera viewpoint of the synthesized right image, \eg we observe the front of the bus occluding the trash can on the sidewalk.
Disoccluded regions are filled in with a texture from a randomly selected image.

%%%%%%%%%%%%%%%%%%%%%%%%%%%%%%%%%%%%
\section{Baseline Algorithm Details}
%%%%%%%%%%%%%%%%%%%%%%%%%%%%%%%%%%%%
Here we describe the baseline image synthesis approaches we compared to in Table 1 in the main paper.
We show qualitative results for these approaches in Fig.~\ref{fig:additional_baseline_results} here and in Fig.~5 in the main paper.

\begin{figure}[t]
\centering
\includegraphics[width=0.6\textwidth]{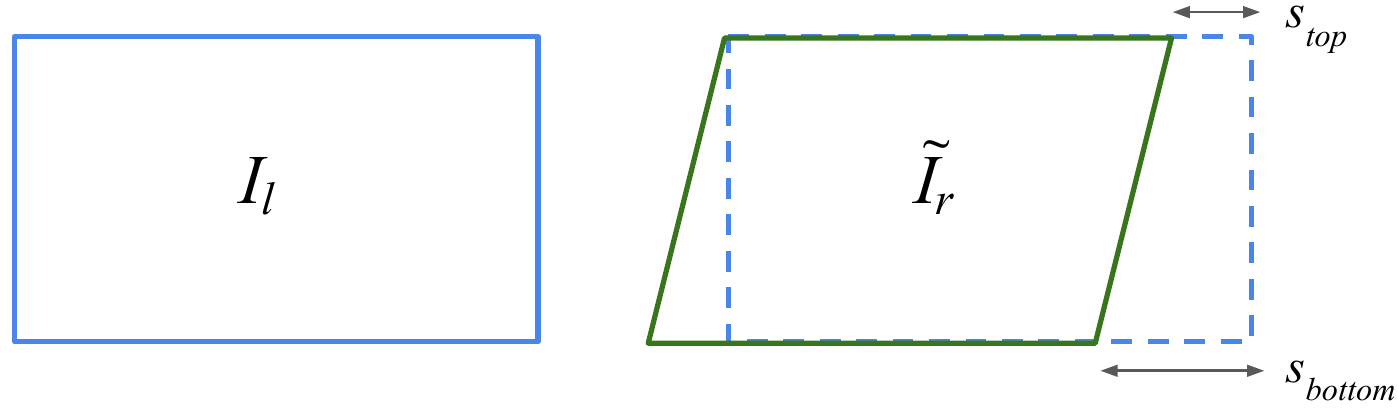} \\
\resizebox{0.75\textwidth}{!}{
\begin{tabular}{ccc}
    \includegraphics[width=0.3\textwidth]{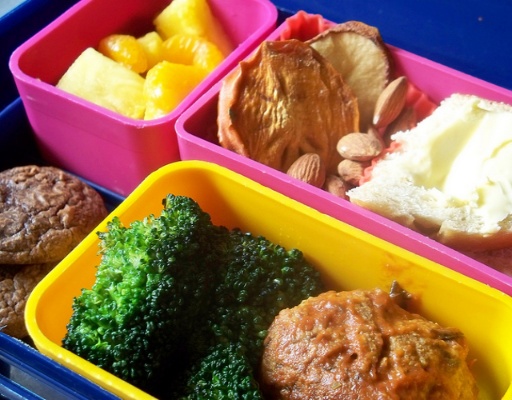} &
    \includegraphics[width=0.3\textwidth]{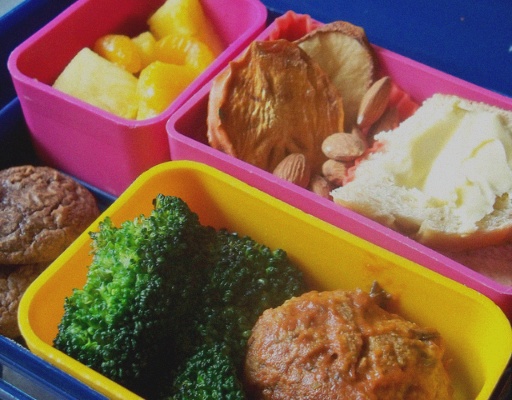} &
    \includegraphics[width=0.3\textwidth]{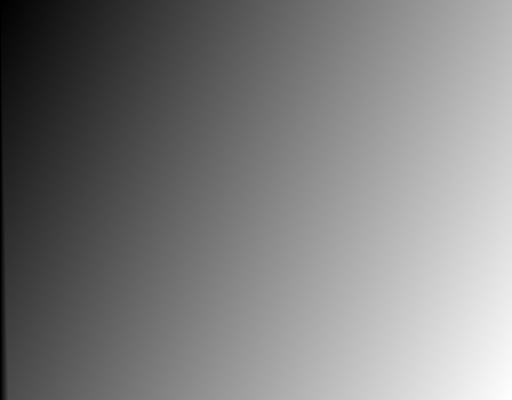} \\
    \includegraphics[width=0.3\textwidth]{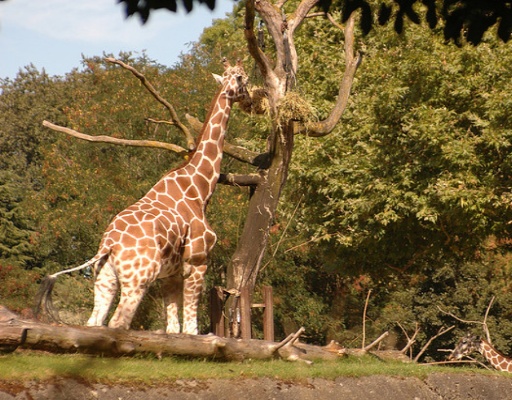} &
    \includegraphics[width=0.3\textwidth]{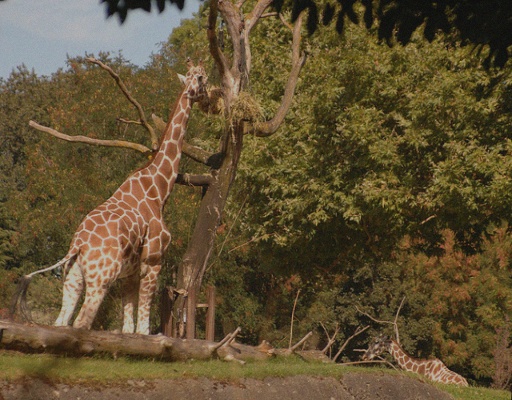} &
    \includegraphics[width=0.3\textwidth]{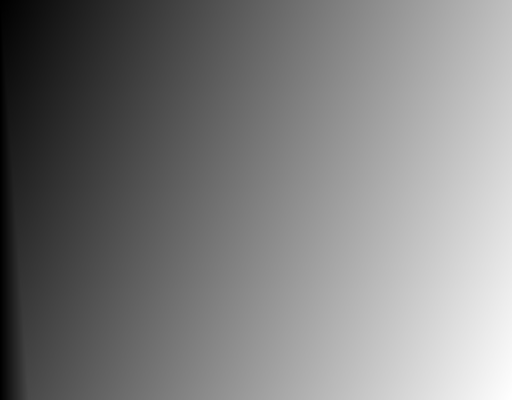} \\
    \includegraphics[width=0.3\textwidth]{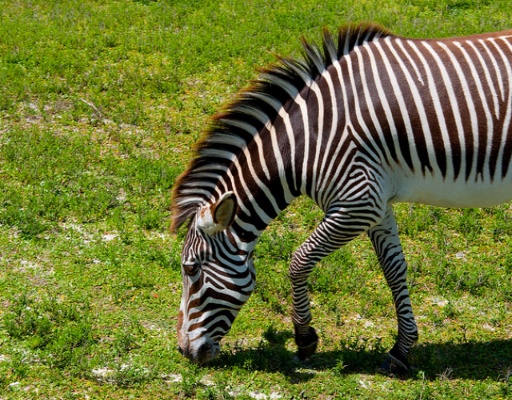} &
    \includegraphics[width=0.3\textwidth]{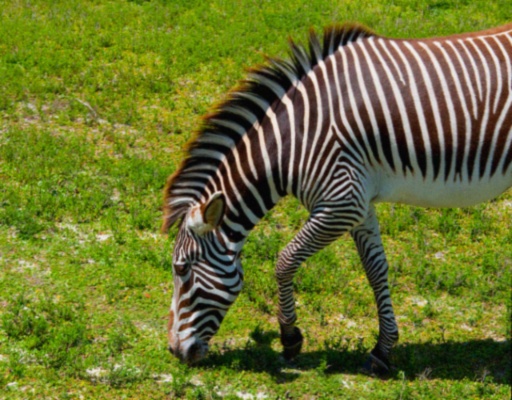} &
    \includegraphics[width=0.3\textwidth]{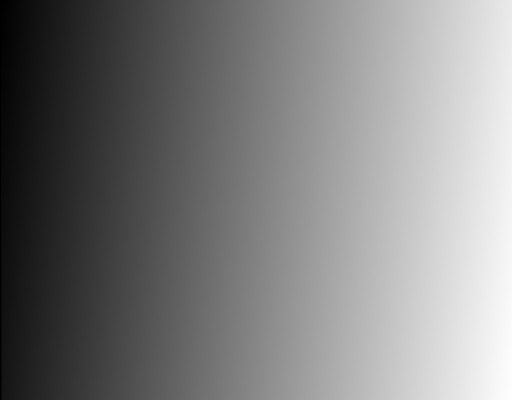} \\
    $I_l$ & $\tilde{I}_r$ & $D$ \\
\end{tabular}
}
\caption{{\bf Affine warp baseline.} The diagram at top shows how $s_{\text{top}}, s_{\text{bottom}}$ relate to the warp applied to $\tilde{I}_r$.
Below, we show examples of the resulting affine warped images.}
\label{fig:affine_warps_examples}
\end{figure}

\renewcommand{\paragraph}[1]{\vspace{6pt}\noindent\textbf{#1 ---}}
\paragraph{Affine Warps}
For the affine warps baseline, similar to \cite{detone18superpoint}, we warp each $I_l$ with a top shift $s_{\text{top}}$ and a bottom shift $s_{\text{bottom}}$, with $50\%$ probability $s_{\text{top}} \sim \text{Unif}\,[0, d_{\max}]$ and $s_{\text{bottom}} \sim \text{Unif}\,[0, s_{\text{top}}]$; and with $50\%$ probability $s_{\text{bottom}} \sim \text{Unif}\,[0, d_{\max}]$ and $s_{\text{top}} \sim \text{Unif}\,[0, s_{\text{bottom}}]$.
Following the warping, $I_l$ and $\tilde{I}_r$ are both cropped to avoid black borders, and the corresponding $D$ is adjusted appropriately to account for this cropping.
Details and example training images are shown in Fig.~\ref{fig:affine_warps_examples}.

\begin{figure}[t]
\centering
\resizebox{0.75\textwidth}{!}{
\begin{tabular}{ccc}
    \includegraphics[width=0.3\textwidth]{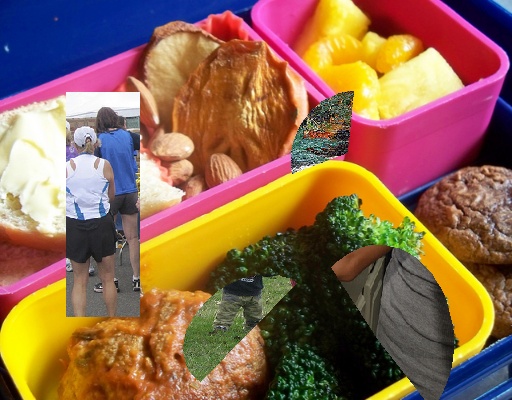} &
    \includegraphics[width=0.3\textwidth]{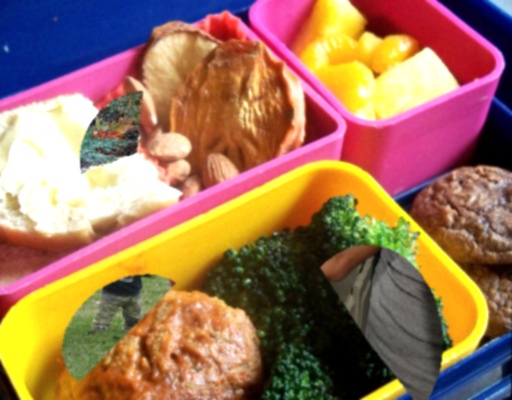} &
    \includegraphics[width=0.3\textwidth]{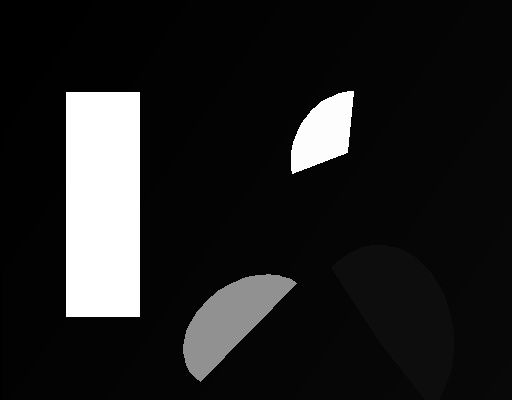} \\
    \includegraphics[width=0.3\textwidth]{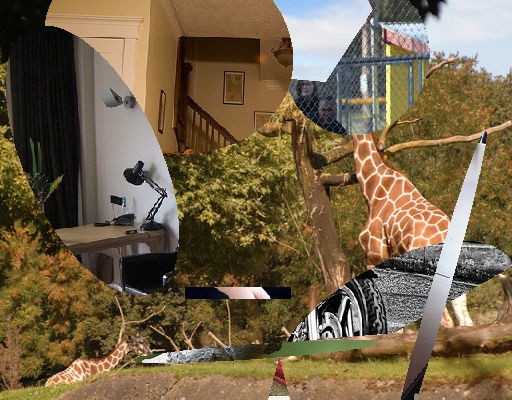} &
    \includegraphics[width=0.3\textwidth]{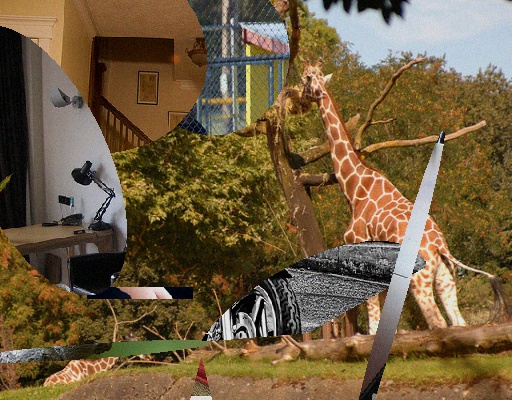} &
    \includegraphics[width=0.3\textwidth]{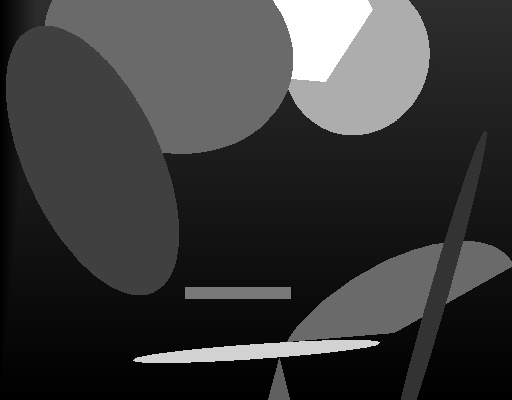} \\
    \includegraphics[width=0.3\textwidth]{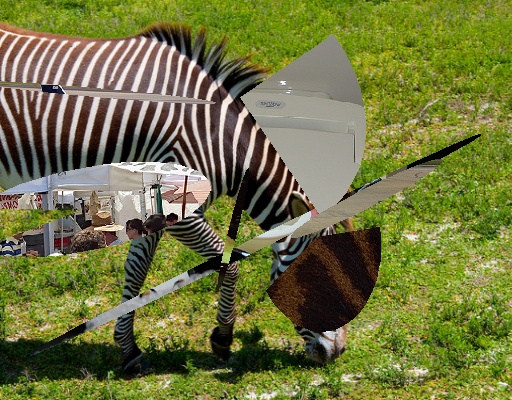} &
    \includegraphics[width=0.3\textwidth]{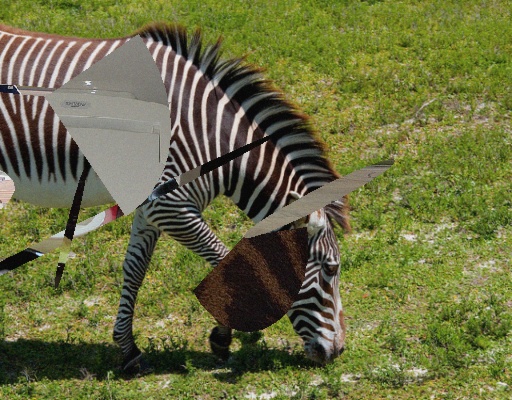} &
    \includegraphics[width=0.3\textwidth]{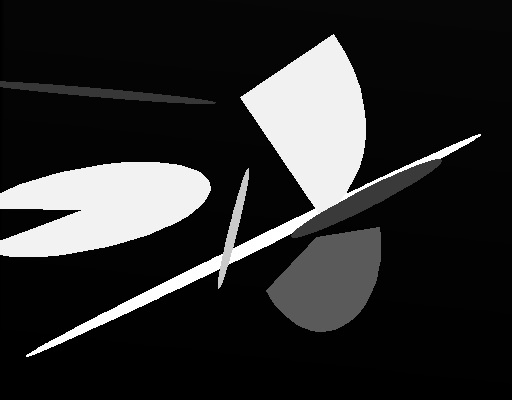} \\
    $I_l$ & $\tilde{I}_r$ & $D$ \\
\end{tabular}
}
\caption{{\bf Random pasted shapes.} Examples images generated using this baseline.}
\label{fig:random_pasted_shapes_examples}
\end{figure}

\paragraph{Random Pasted Shapes}
For the random pasted shapes baseline we paste `foreground' shapes onto a `background' image in a similar manner to~\cite{mayer2018makes}.
We start with affine warped images, but with $d_{\max} = 50$ to help to ensure that foreground `objects' move with greater disparity than the `background' image.
We then sample $\text{num\_patches} \sim \text{Unif}[0, 10]$ and for each patch we load a random image from the training set and mask out a region.
The masked region is pasted on top of $I_l$, and a translated version of the masked region is placed on $\tilde{I}_r$.
Masks are generated with equal probability from rectangles, partial ellipses, polygons and thin objects.
With the exception of rectangles, masks are generated with OpenCV's drawing functions.
Examples of random pasted shapes training tuples are shown in Fig.~\ref{fig:random_pasted_shapes_examples}.
Note that the disparity of the background image is less visible compared with the disparity maps from the Affine Warp baseline, due to the reduced $d_{\max}$ used here.
\vspace{-5pt}
\begin{description}[0.5cm]
\item[Rectangles] are axis-aligned, with $x$-axis and $y$-axis bounds sampled from $\text{Unif}[0.1  W, 0.9 W]$ and $\text{Unif}[0.1 H, 0.9 H]$ respectively.
\item[Partial ellipses] have a center $(x, y)$,  where  $x \sim \text{Unif}[0.1  W, 0.9 W]$ and  $y \sim \text{Unif}[0.1 H, 0.9 H]$.
With $75\%$ probability the ellipse is a full ellipse, with $\text{start\_angle} = 0$ and $\text{end\_angle} = 360$.
Otherwise,  the ellipse is partial, with the angular bounds sampled from  $\text{Unif}[0, 360]$.
The rotation angle of the ellipse is sampled from $\text{Unif}[0, 360]$, and the axes sizes are sampled from $\text{Unif}[0.1 W, 0.9 W]$.
\item[Polygons] are generated using the approach described in \cite{ounsworth2020polygon}, with a number of sides sampled from $\text{Unif}[3, 20]$, $\text{spikyness}=0.8$, $\text{irregularity}=0.5$, and $\text{aveRadius} \sim \text{Unif}[0.01W, 0.3W]$.
\item[Thin objects] are full ellipses with with minor axis $\sim \text{Unif}[0.001H, 0.025H]$ and major axis $\sim \text{Unif}[0.1W, 0.5W]$.
\end{description}
\vspace{-5pt}

\begin{figure}[t]
\centering
\resizebox{0.8\textwidth}{!}{
\begin{tabular}{ccc}
    \includegraphics[width=0.3\textwidth]{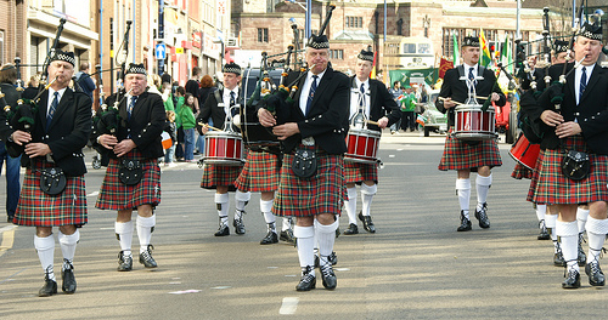} &
    \includegraphics[width=0.3\textwidth]{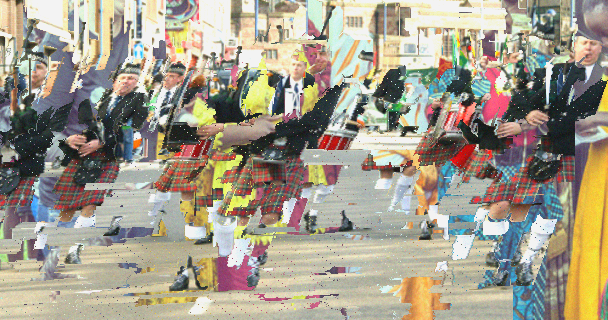} &
    \includegraphics[width=0.3\textwidth]{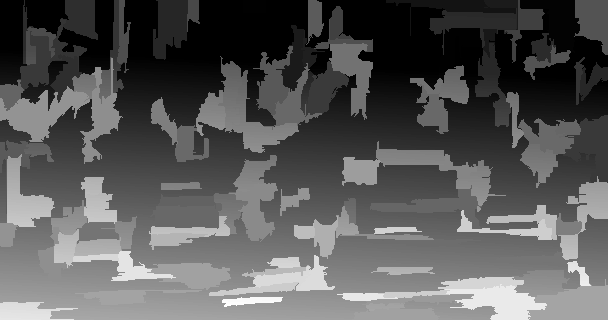} \\
    \includegraphics[width=0.3\textwidth]{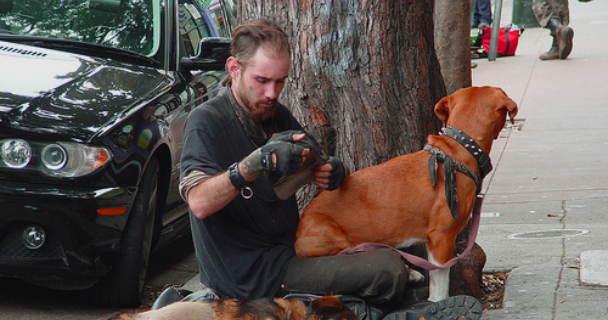} &
    \includegraphics[width=0.3\textwidth]{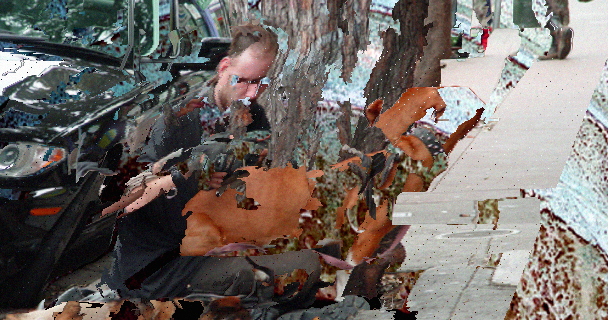} &
    \includegraphics[width=0.3\textwidth]{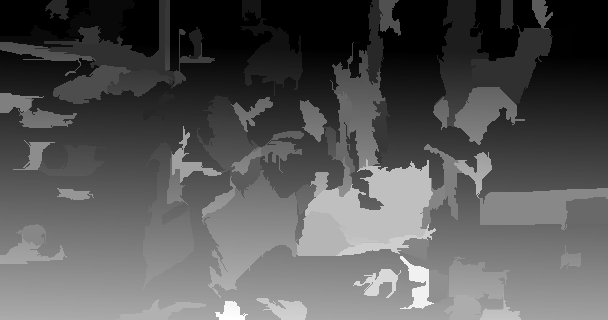} \\
    \includegraphics[width=0.3\textwidth]{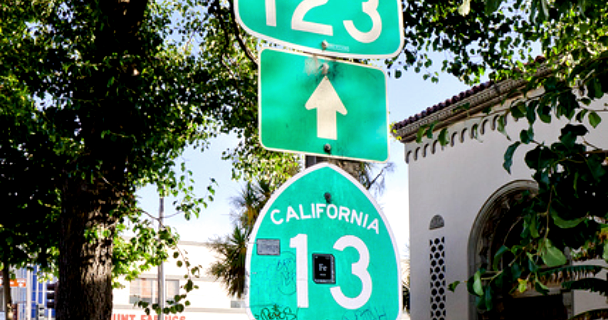} &
    \includegraphics[width=0.3\textwidth]{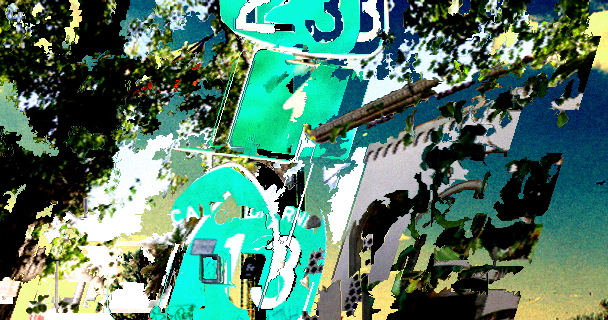} &
    \includegraphics[width=0.3\textwidth]{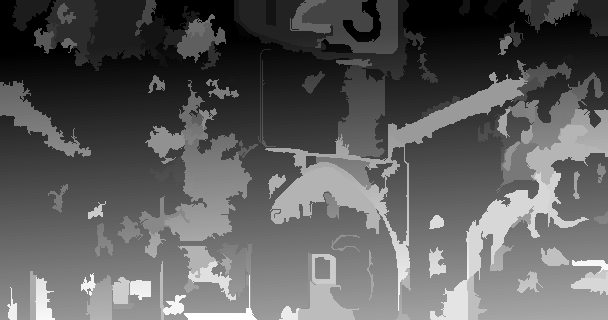} \\
    $I_l$ & $\tilde{I}_r$ & $D$ \\
\end{tabular}
}
\caption{{\bf Random superpixels.} Synthesized stereo pairs using superpixel warping.}
\label{fig:random_super_pixels}
\end{figure}

\begin{figure}
    \centering
    \newcommand{\itemwidth}{0.3\columnwidth}
    \centering
    \vspace{10pt}
    \begin{tabular}{ccc}
        \includegraphics[width=\itemwidth]{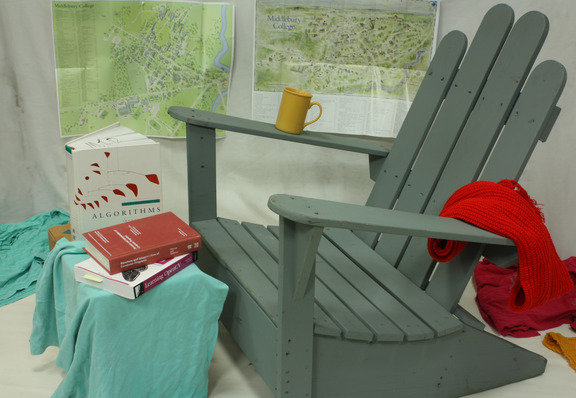} &
        \includegraphics[width=\itemwidth]{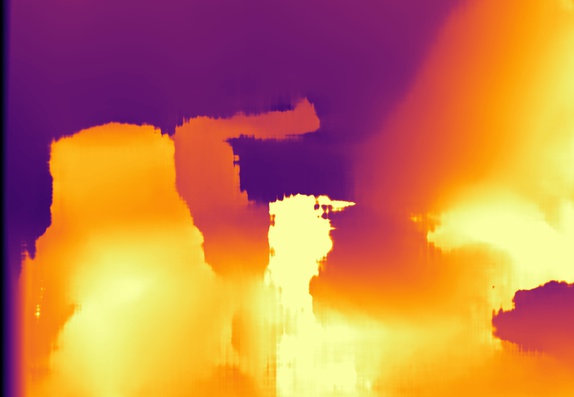} &
        \includegraphics[width=\itemwidth]{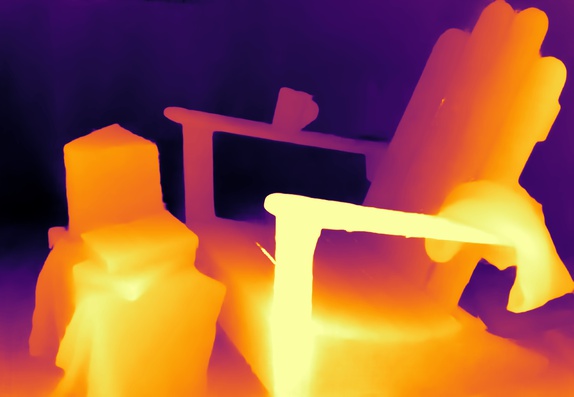} \\
        \scriptsize Input & \scriptsize  Affine warps &  \scriptsize  Superpixels \\
        \includegraphics[width=\itemwidth]{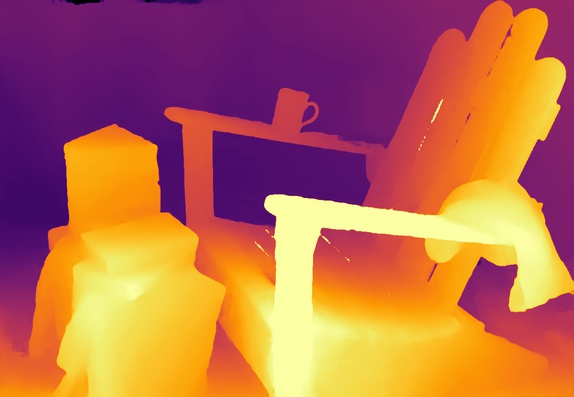} &
        \includegraphics[width=\itemwidth]{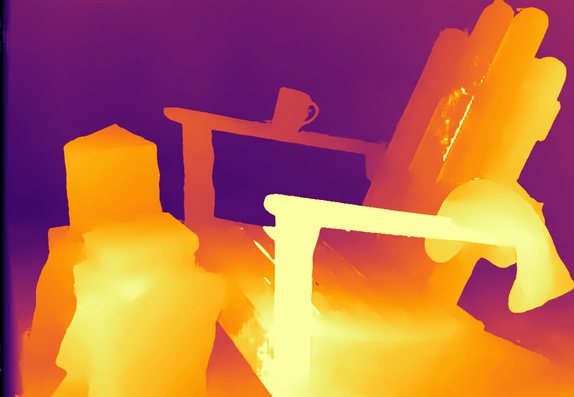} &
        \includegraphics[width=\itemwidth]{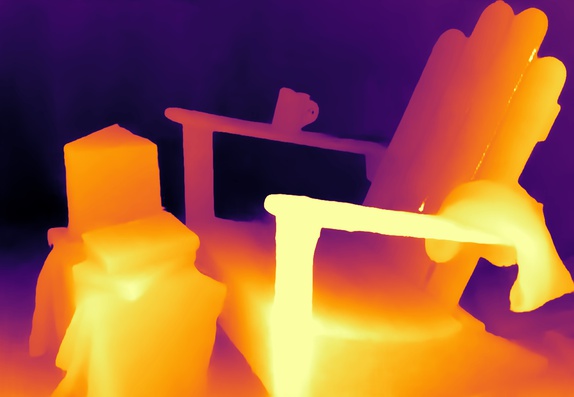} \\
        \scriptsize Sceneflow  & \scriptsize  Random pasted shapes & \scriptsize  Ours  \\
        \hline \\

        \includegraphics[width=\itemwidth]{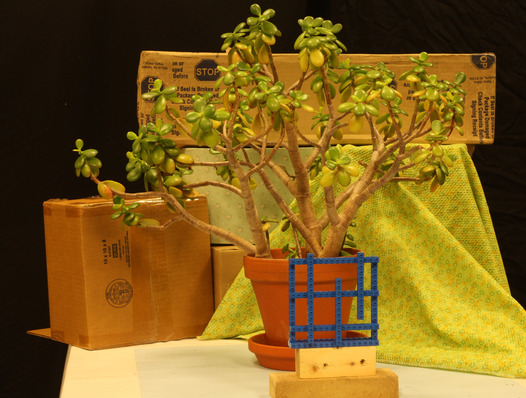} &
        \includegraphics[width=\itemwidth]{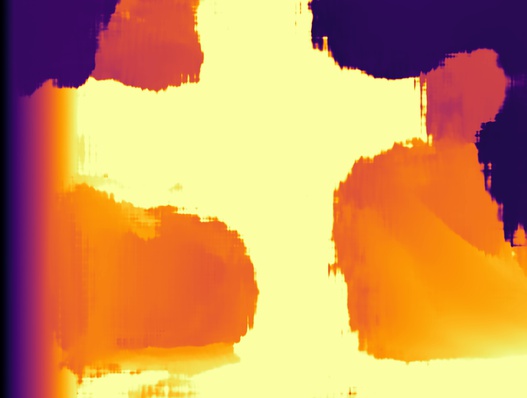} &
        \includegraphics[width=\itemwidth]{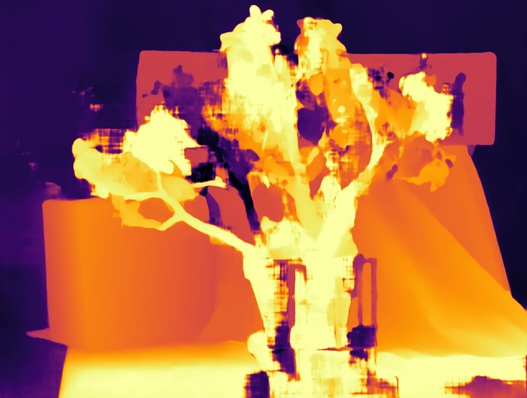} \\
        \scriptsize Input & \scriptsize  Affine warps &  \scriptsize  Superpixels \\
        \includegraphics[width=\itemwidth]{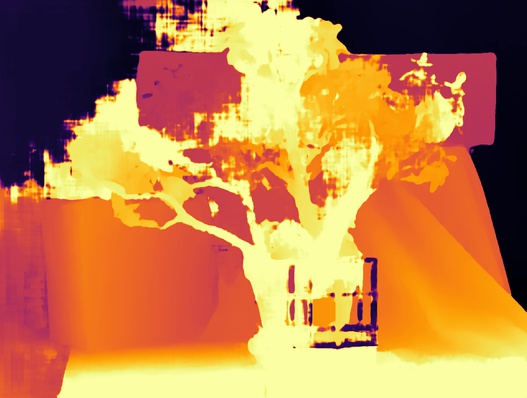} &
        \includegraphics[width=\itemwidth]{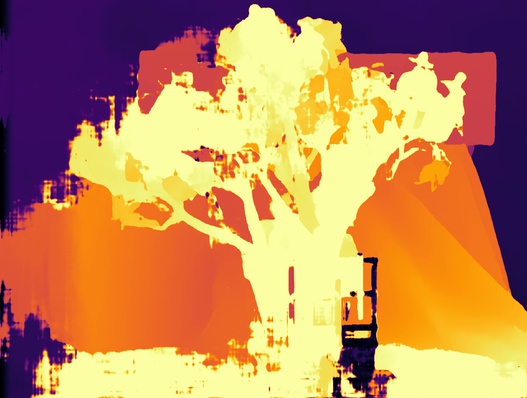} &
        \includegraphics[width=\itemwidth]{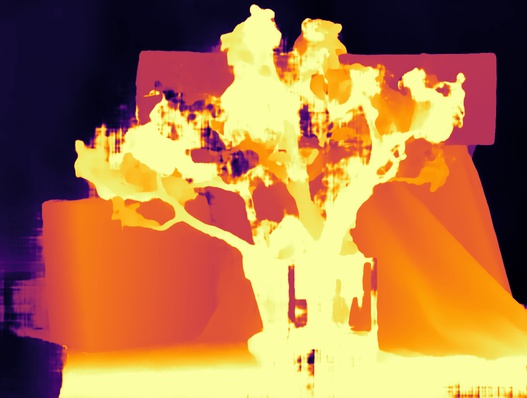} \\
        \scriptsize Sceneflow & \scriptsize  Random pasted shapes & \scriptsize  Ours  \\
        \hline \\

        \includegraphics[width=\itemwidth]{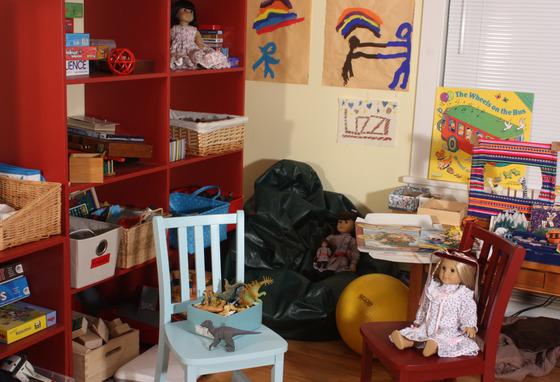} &
        \includegraphics[width=\itemwidth]{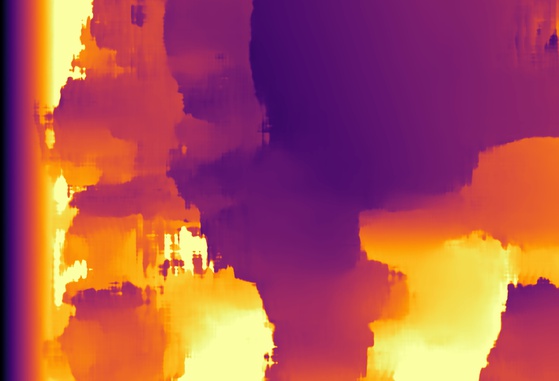} &
        \includegraphics[width=\itemwidth]{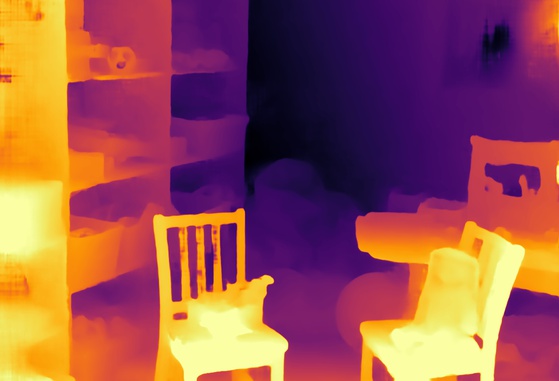} \\
        \scriptsize Input & \scriptsize  Affine warps &  \scriptsize  Superpixels \\
        \includegraphics[width=\itemwidth]{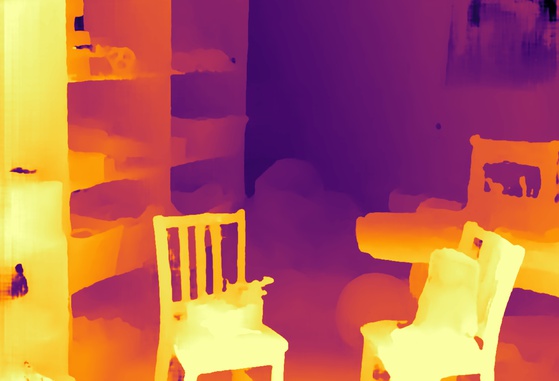} &
        \includegraphics[width=\itemwidth]{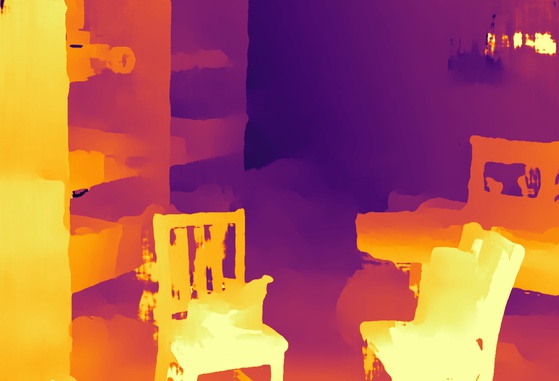} &
        \includegraphics[width=\itemwidth]{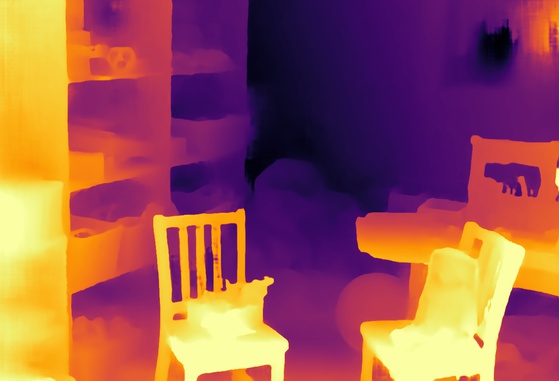} \\
        \scriptsize Sceneflow & \scriptsize  Random pasted shapes & \scriptsize  Ours  \\
    \end{tabular}
    \vspace{-10pt}
    \caption{{\bf Additional results comparing our method to the baseline algorithms.}}
    \vspace{-15pt}
    \label{fig:additional_baseline_results}
\end{figure}

\paragraph{Random Superpixels}
For this baseline, we segment $I_l$ into superpixels using \cite{Felzenszwalb2004,van2014scikit}, with parameters $\text{scale} \sim \text{Unif}[50, 200]$, $\sigma \sim \text{Unif}[0, 1]$, and $\text{min\_size} \sim \text{Unif}[75, 275]$.
We initialise $D$ as a plane, where pixel $i$ has disparity value $d_i$ defined by $d_i = ax + by +c$ with $a \sim \text{Unif}[-0.025, 0.025]$, $b \sim \text{Unif}[0.3, 0.4]$, and $c \sim \text{Unif}[15, 20]$.
Superpixels $s_j$ in $I_l$ are chosen with probability 0.6 to act as foreground objects.
We set the disparity values of these foreground superpixels as $d_j = \Sigma_{i \in j}\frac{d_i}{n} + x_j$, where $x_j  \sim \text{Unif}[0, 64]$ and $n$ is the number of pixels in superpixel $s_j$.
Finally we clip the values of $D$ to lie between 0 and $d_{\max}$.
Using $D$, we then forward warp $I_l$ to generate our right image $\tilde{I}_r$, handing occlusions and collisions in the way described in Section 3.2 of the main paper.
Examples of this training data are shown in Fig.~\ref{fig:random_super_pixels}.

\paragraph{SVSM selection module} For this baseline, we use the selection module from~\cite{luo2018single}. Specifically, we still make use of monocular depth estimates from~\cite{lasinger2019towards}, converting the single channel monocular estimate to a one-hot encoding over possible disparities.  We then forward warp the disparity and input image, subsequently taking the weighted sum. Interestingly, we did also try to train a network for image synthesis as per the paper (using KITTI), and use this to generate our synthetic right images. Whist the resulting images were reasonable, the output of the network is a distribution over disparities, and using the argmax of the distrbution as a training target for a stereo network resulted in very poor performance.

%%%%%%%%%%%%%%%%%%%%%%%%%%%%%%%%%%%%
\section{Evaluation Details}
%%%%%%%%%%%%%%%%%%%%%%%%%%%%%%%%%%%%
%Here we provide additional details for the experiments in main paper.
%\vspace{-10pt}

\subsection{Evaluation Image Resolution}
Due to architecture constraints, we make predictions at slightly different resolutions for each network; see Table \ref{tab:evaluation_resolutions} for details.
Note that all predictions are resized to the native resolution of the ground truth for evaluation.
When training a given architecture with different datasets we use the same resolution for all datasets for fair comparison.

\begin{table}
\footnotesize
\centering
\caption{{\bf Evaluation image resolution.} Here we show the prediction resolutions for each architecture. Note that all predictions are resized to the native resolution of the ground truth for evaluation.}
{\begin{tabular}{|l||c|c|c|c|}
\hline
\small
%   & & KITTI '12 & KITTI '15  & Midd.  & ETH3D  \\
Architecture & KITTI `12& KITTI `15 & Middlebury & ETH3D \\
\hline
iResnet~\cite{liang2018learning} &  $1280 \times 384$ & $1280 \times 384$ & $1280 \times 768$ & $768 \times 448$ \\
PSM~\cite{chang2018pyramid} & $1280 \times 384$ & $1280 \times 384$ & $1280 \times 768$ & $768 \times 448$ \\
GANet~\cite{Zhang2019GANet} & $1248 \times 384$ & $1248 \times 384$ & $1248 \times 768$ & $768 \times 432$ \\
\hline
\end{tabular}}
\label{tab:evaluation_resolutions}
\end{table}

\subsection{KITTI Evaluation Splits}
KITTI 2012 results in the paper are reported on the 40 images from the validation split of \cite{chang2018pyramid}.
The image indices used by both us and \cite{chang2018pyramid} are \texttt{[3, 6, 20, 26, 38, 41, 43, 44, 49, 60, 67, 70, 81, 84, 89, 97, 109, 119, 122, 123, 129, 130, 132, 134, 141, 144, 152, 158, 165, 171, 174, 179, 184, 186]}.
The KITTI 2015 indices are
\texttt{[1, 3, 6, 20, 26, 35, 38, 41, 43, 44, 49, 60, 67, 70, 81, 84, 89, 97, 109, 119, 122, 123, 129, 130, 132, 134, 141, 144, 152, 158, 159, 165, 171, 174, 179, 182, 184, 186, 187, 196]}.

\section{Additional KITTI Results}
In Table~\ref{tab:kitti_2012_2015_finetuning_full_results} we present  additional metrics for the KITTI 2012 experiments from Table~1 in the main paper where we compare different methods for generating training data.
We also show additional metrics for KITTI 2015.
As in the main paper, our method brings consistent improvement over the baselines.

\begin{table}
\footnotesize
\centering
\vspace{-5pt}
\caption{{\bf KITTI 2012 and 2015 Results.} Additional metrics for PSMNet~\cite{chang2018pyramid} trained with different stereo data.}
{\begin{tabular}{|p{3.6cm}|p{2.2cm}||c|c|c|c|}
\hline
\small
%   & & KITTI '12 & KITTI '15  & Midd.  & ETH3D  \\
Synthesis approach & Training data & EPE Noc & $>$3px Noc & EPE All & $>$3px All  \\%$>$3px & $>$3px & $>$2px  & $>$1px \\
\hline
\hline
\multicolumn{6}{|c|}{\bf KITTI 2012}\\
\hline
Affine warps & MfS & 3.04 & 12.72 & 3.74 & 14.78 \\
Random pasted shapes & MfS & 1.90 & 9.97 & 2.45 & 11.89 \\
Random superpixels & MfS & 1.09 & 5.16 & 1.28 & 6.08 \\
Synthetic & Sceneflow  & 0.97 & 5.09 & 1.04 & 5.76 \\
\hline
Ours & MfS & \textbf{0.79} & \textbf{3.60} & \textbf{0.95} & \textbf{4.43} \\
\hline
\hline
\multicolumn{6}{|c|}{\bf KITTI 2015}\\
\hline
Affine warps & MfS & 2.05 & 13.67 & 2.33 & 14.94 \\
Random pasted shapes & MfS & 1.23 & 6.24 & 1.49 & 7.57 \\
Random superpixels & MfS & 1.22 & 5.62 & 1.23 & 5.90 \\
Synthetic & Sceneflow  & 1.18 & 5.56 & 1.19 & 5.80  \\
\hline
Ours & MfS & \textbf{1.02} & \textbf{4.57} & \textbf{1.04} & \textbf{4.75} \\
\hline
\end{tabular}}
\vspace{-10pt}
\label{tab:kitti_2012_2015_finetuning_full_results}
\end{table}

%%%%%%%%%%%%%%%%%%%%%%%%%%%%%%%%%%%%%%%%%%%%%%%%%%%%
\section{Recovering from Monocular Depth Errors}
%%%%%%%%%%%%%%%%%%%%%%%%%%%%%%%%%%%%%%%%%%%%%%%%%%%%

\begin{figure}
    \centering
    \newcommand{\itemwidth}{0.30\columnwidth}
    \centering
    \vspace{10pt}
    \begin{tabular}{ccc}
            \includegraphics[width=\itemwidth]{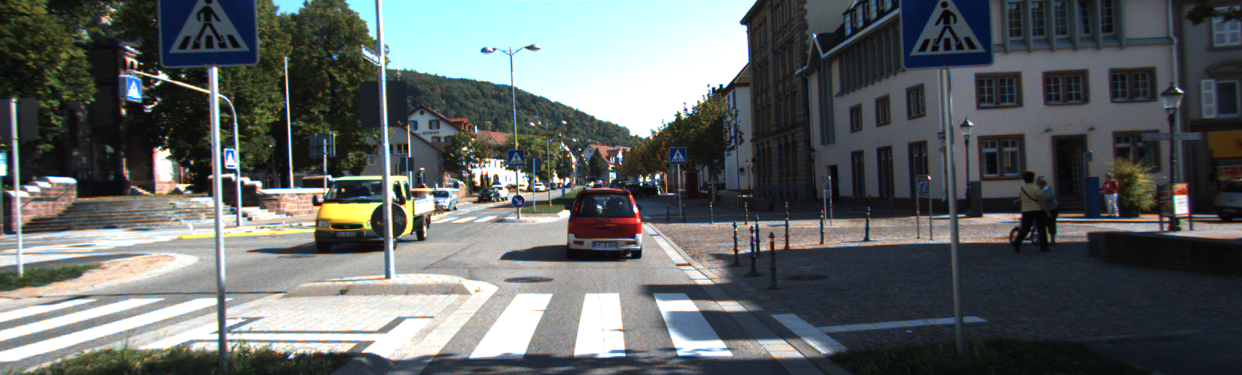}\vspace{-4pt} \\
            \scriptsize Input \\
        \includegraphics[width=\itemwidth]{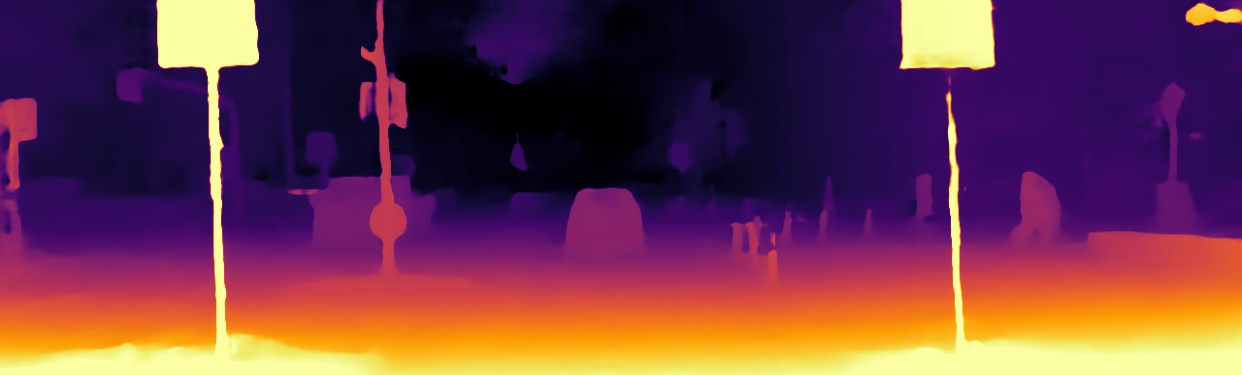} &
        \includegraphics[width=\itemwidth]{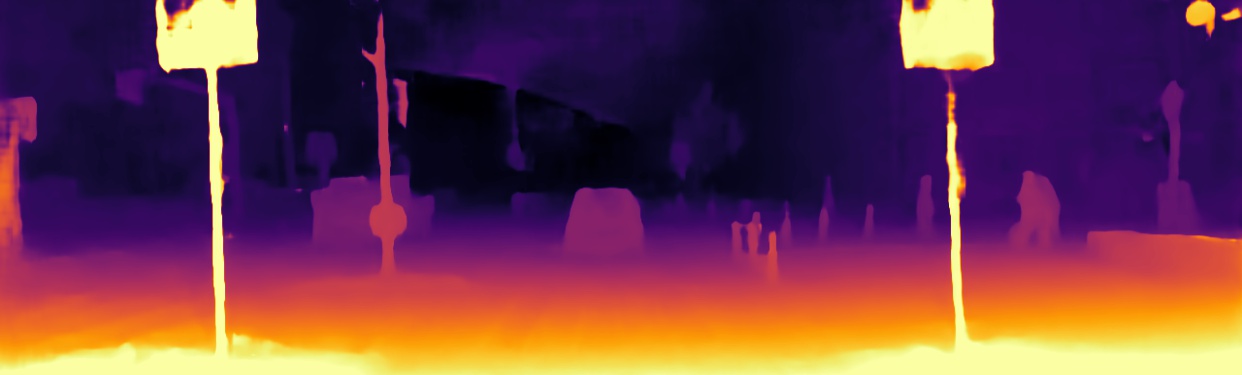} &
        \includegraphics[width=\itemwidth]{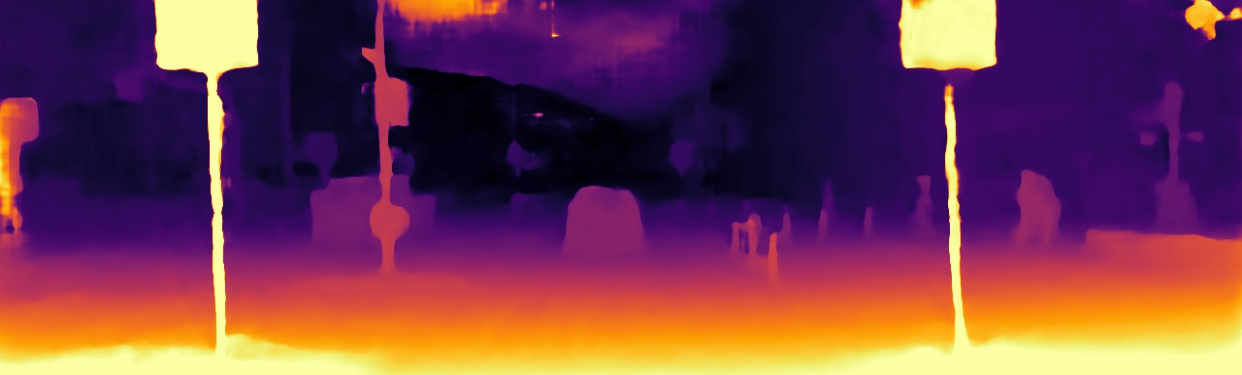} \vspace{-4pt} \\
        \scriptsize  Stereo prediction &  \scriptsize  Stereo prediction &  \scriptsize  Stereo prediction \vspace{-3pt}  \\
        \scriptsize  (MfS+MiDaS\cite{lasinger2019towards}) &  \scriptsize  (MfS+Megadepth\cite{li2018megadepth}) &  \scriptsize  (MfS+\cite{godard2019digging}) \vspace{6pt}  \\
         \includegraphics[width=\itemwidth]{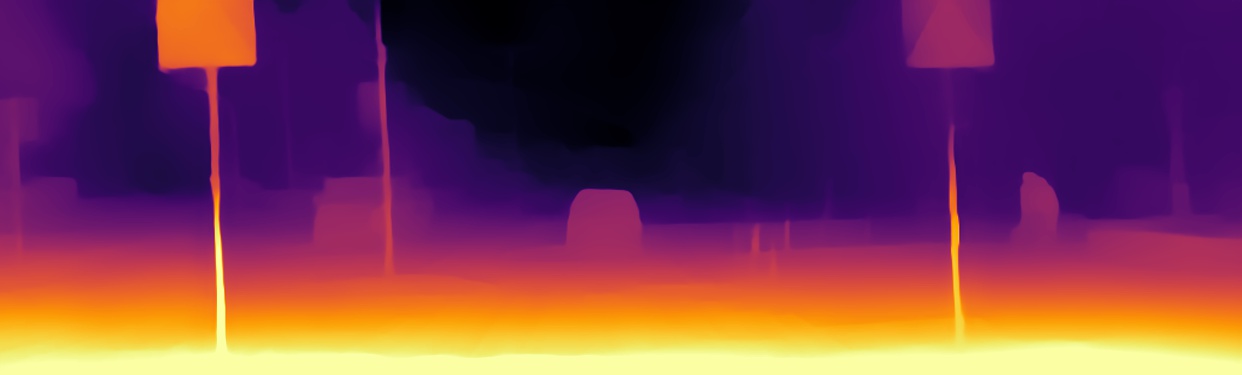} &
        \includegraphics[width=\itemwidth]{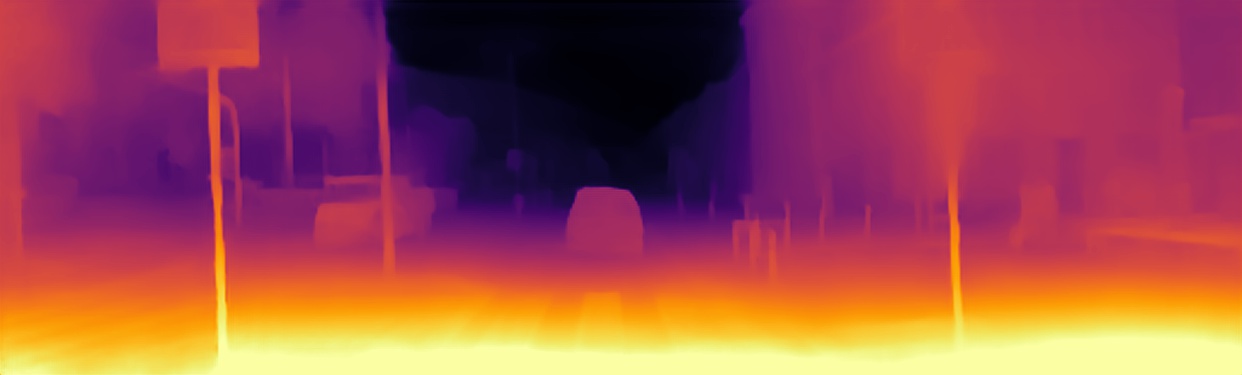}  &
        \includegraphics[width=\itemwidth]{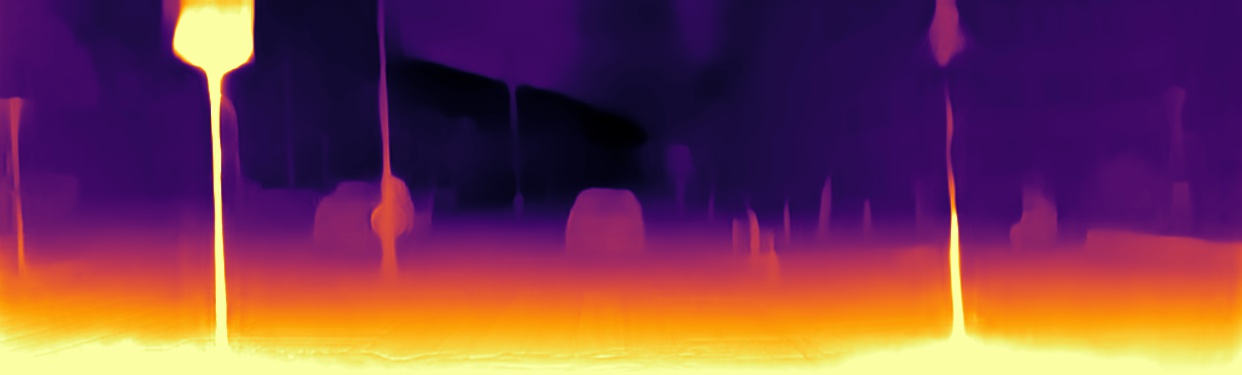} \vspace{-4pt} \\
         \scriptsize  Mono prediction  & \scriptsize  Mono prediction & \scriptsize Mono prediction \vspace{-3pt}  \\
         \scriptsize (MiDaS \cite{lasinger2019towards}) & \scriptsize (Megadepth \cite{li2018megadepth}) & \scriptsize  Monodepth2 \cite{godard2019digging} \vspace{4pt} \\
    \end{tabular}
    \vspace{-10pt}
    \caption{{\bf We can recover from errors in monocular depth estimation.} }
    \vspace{-15pt}
    \label{fig:monodepth_errors}
\end{figure}

Fig.~4 in the main paper shows that our trained stereo networks can overcome some of the errors of monocular depth estimation.
In Fig.~\ref{fig:monodepth_errors} here, we observe the same result across three different monocular depth networks: MiDaS~\cite{lasinger2019towards}, Megadepth~\cite{li2018megadepth}, and Monodepth2~\cite{godard2019digging}. In each case, problems present in monocular depths, such as missing objects and uneven ground planes do not transfer to our eventual stereo predictions.
We note here that although Monodepth2~\cite{godard2019digging} was trained using monocular videos from KITTI, our resulting stereo network has never seen images from this dataset.

\section{Additional Qualitative Results}
\begin{figure}[!t]
\centering
\newcommand{\itemwidth}{0.35\columnwidth}
\centering
\begin{tabular}{@{\hskip 2mm}c@{\hskip 2mm}c@{\hskip 2mm}c@{\hskip 2mm}c@{}}
{\scriptsize Input (left) image} & {\scriptsize Trained with Sceneflow} & {\scriptsize Trained with our MfS} \\
\includegraphics[width=\itemwidth]{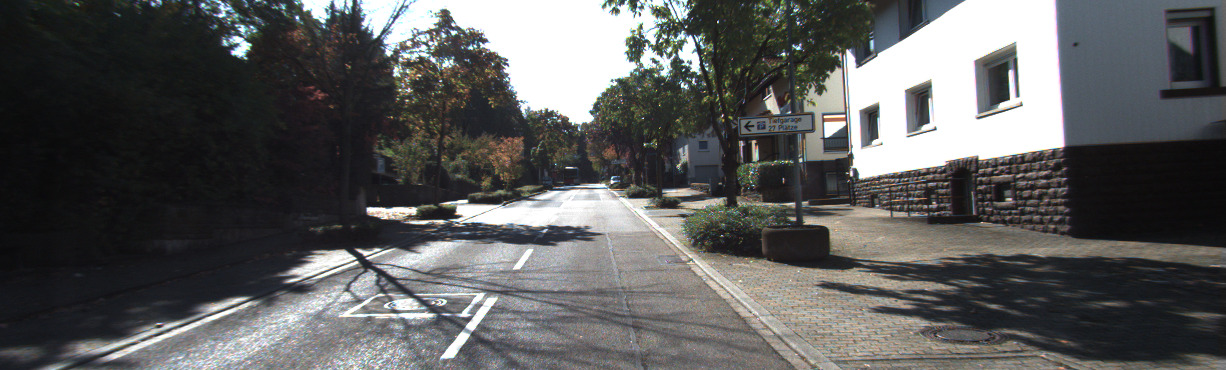} &
\includegraphics[width=\itemwidth]{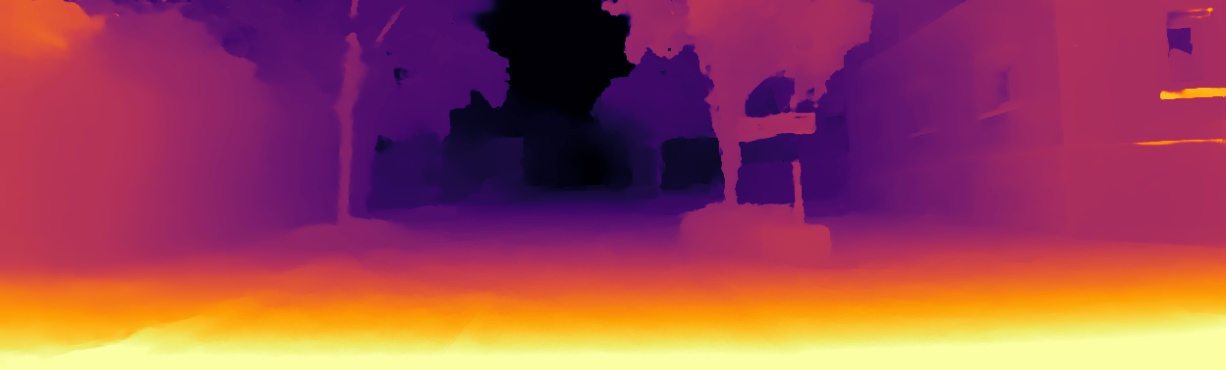} &
\includegraphics[width=\itemwidth]{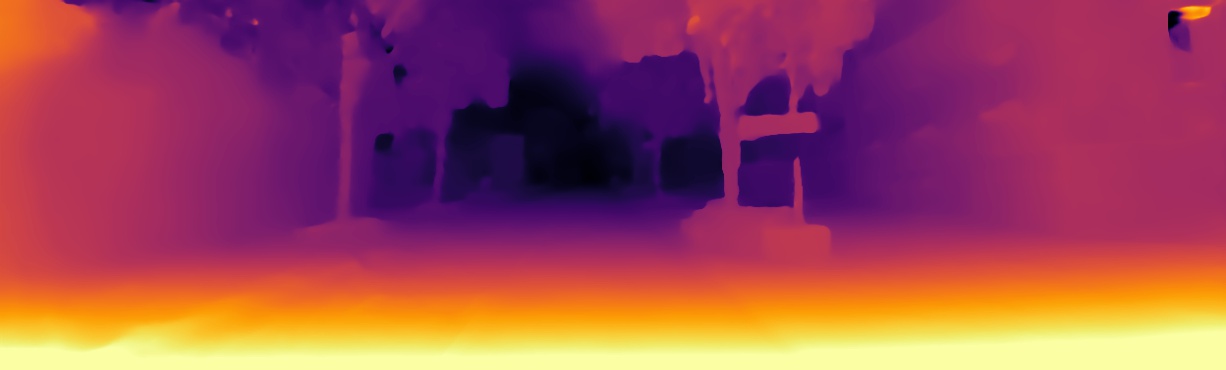} \\
\includegraphics[width=\itemwidth]{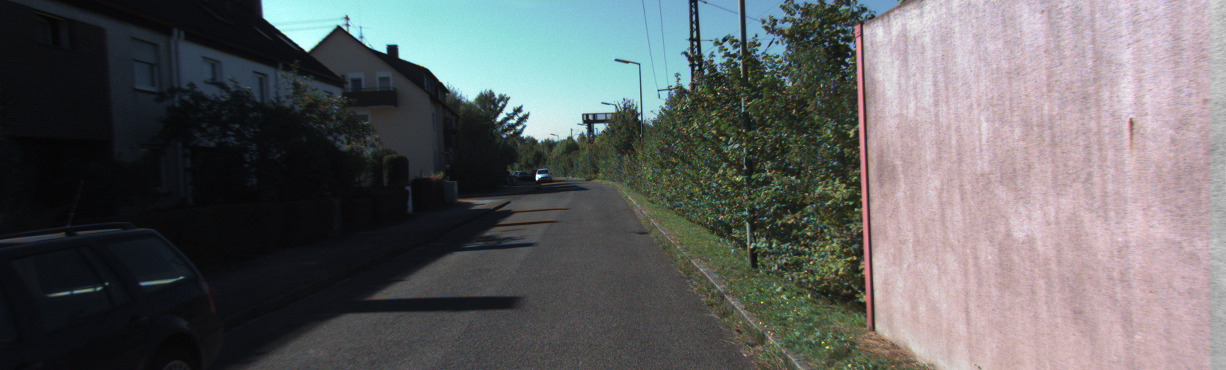} &
\includegraphics[width=\itemwidth]{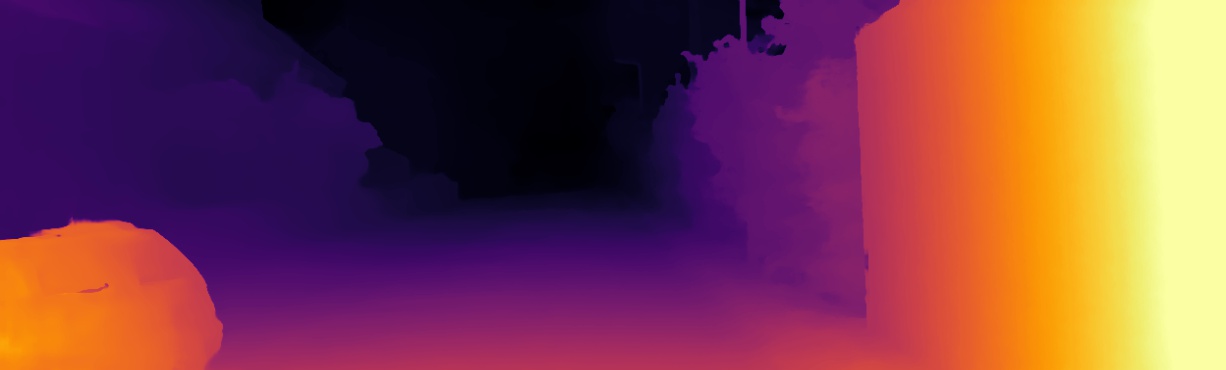} &
\includegraphics[width=\itemwidth]{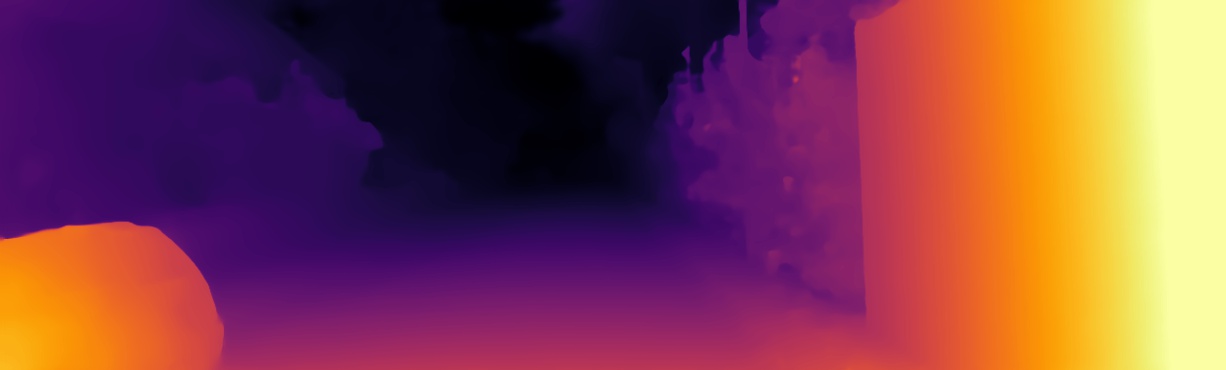} \\
\includegraphics[width=\itemwidth]{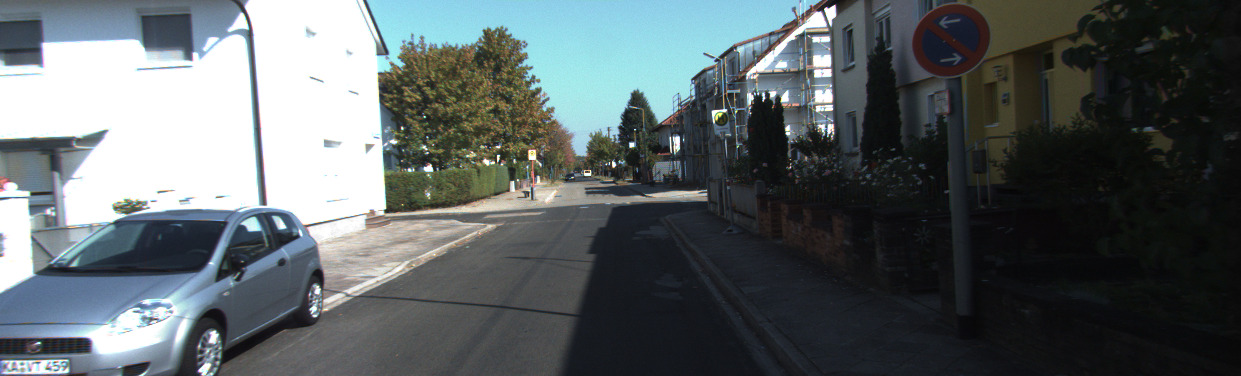} &
\includegraphics[width=\itemwidth]{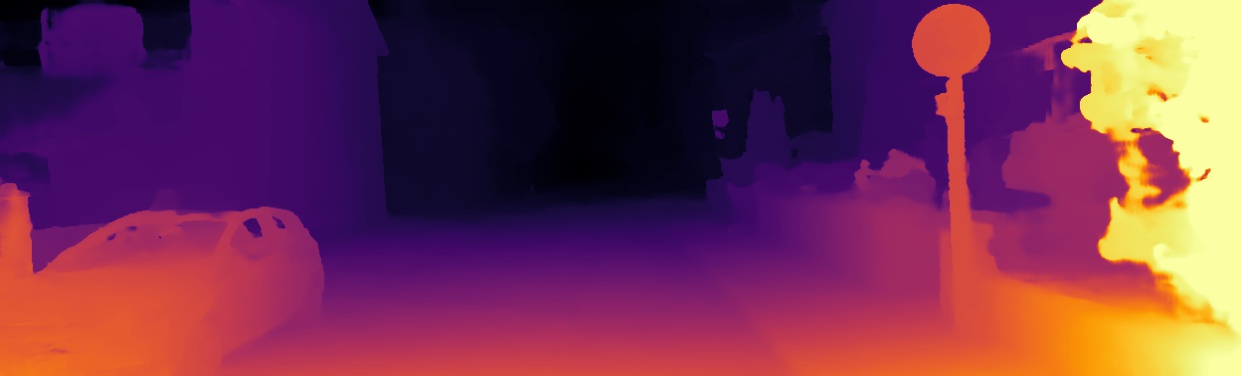} &
\includegraphics[width=\itemwidth]{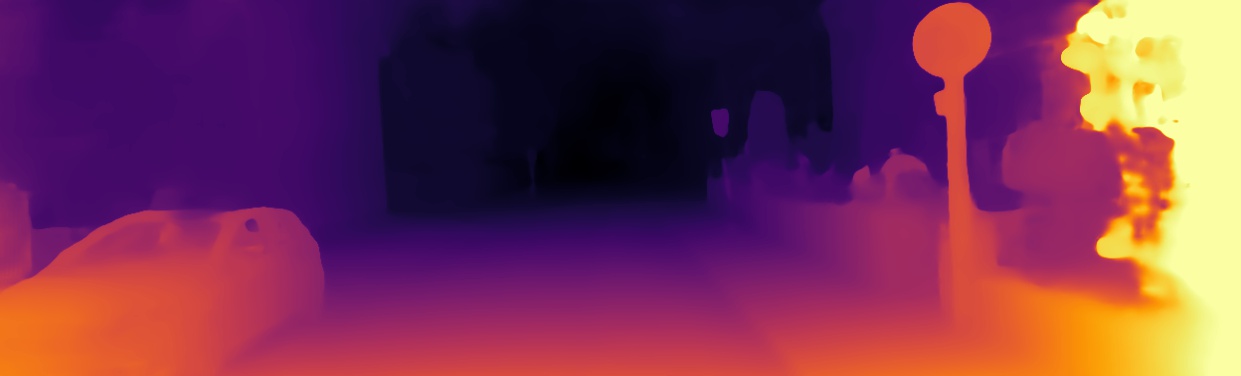} \\
\includegraphics[width=\itemwidth]{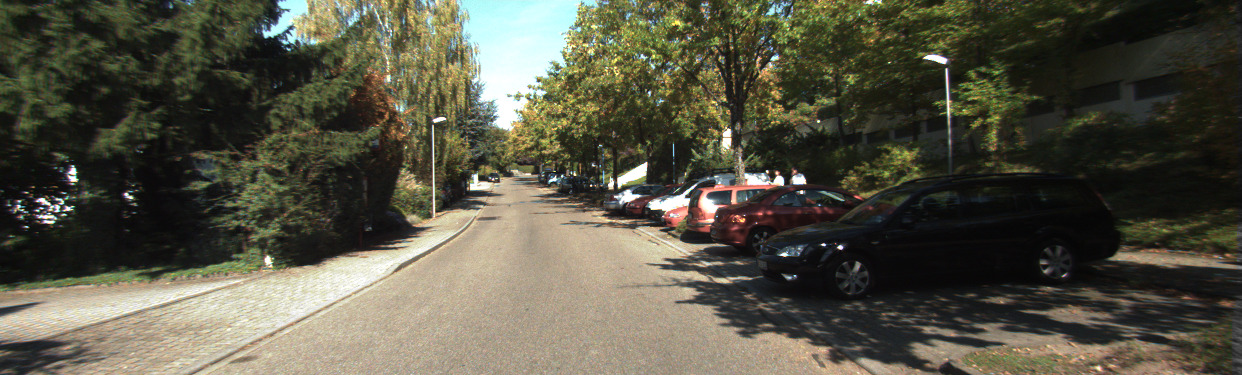} &
\includegraphics[width=\itemwidth]{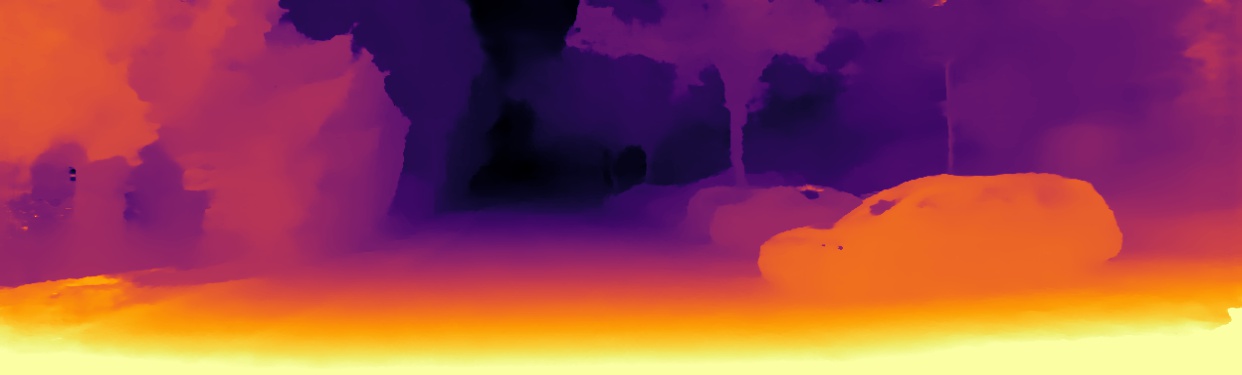} &
\includegraphics[width=\itemwidth]{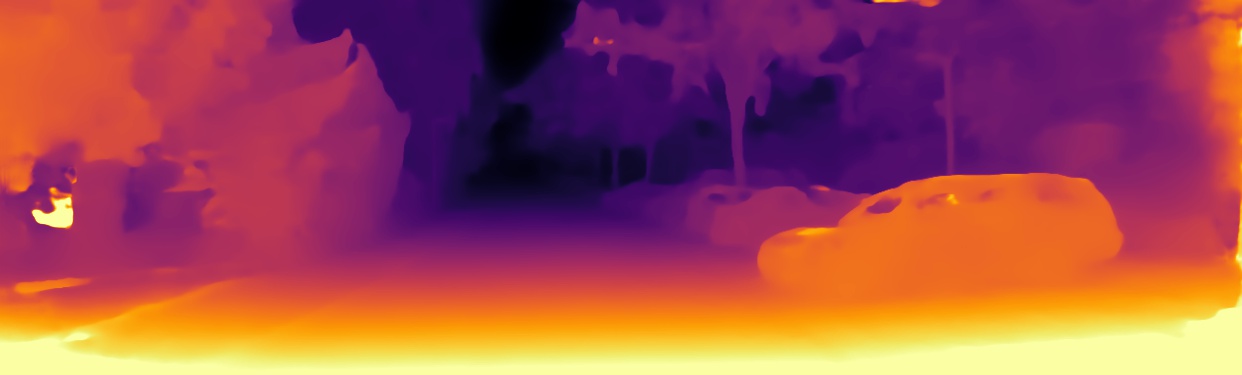} \\
\includegraphics[width=\itemwidth]{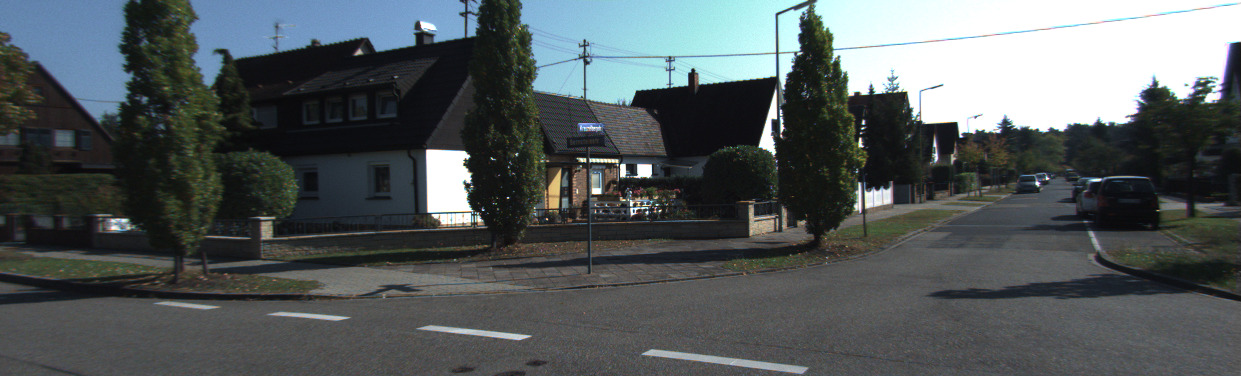} &
\includegraphics[width=\itemwidth]{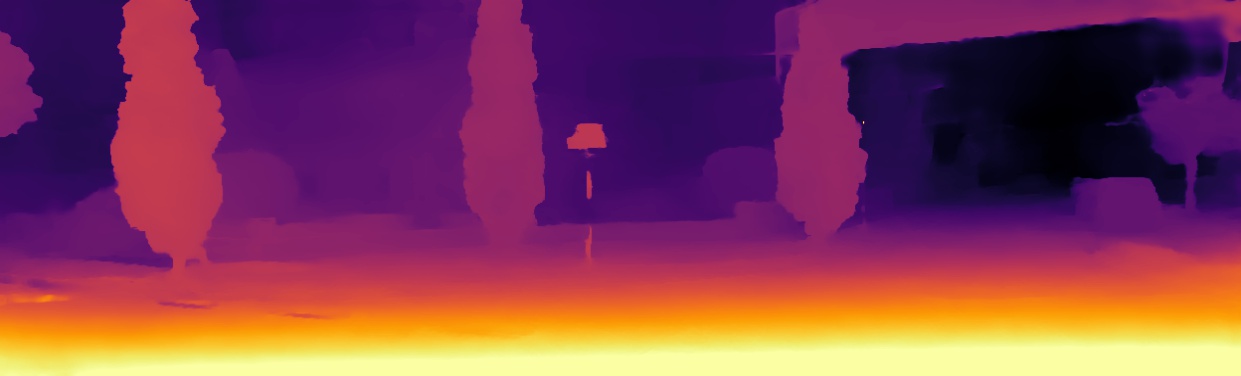} &
\includegraphics[width=\itemwidth]{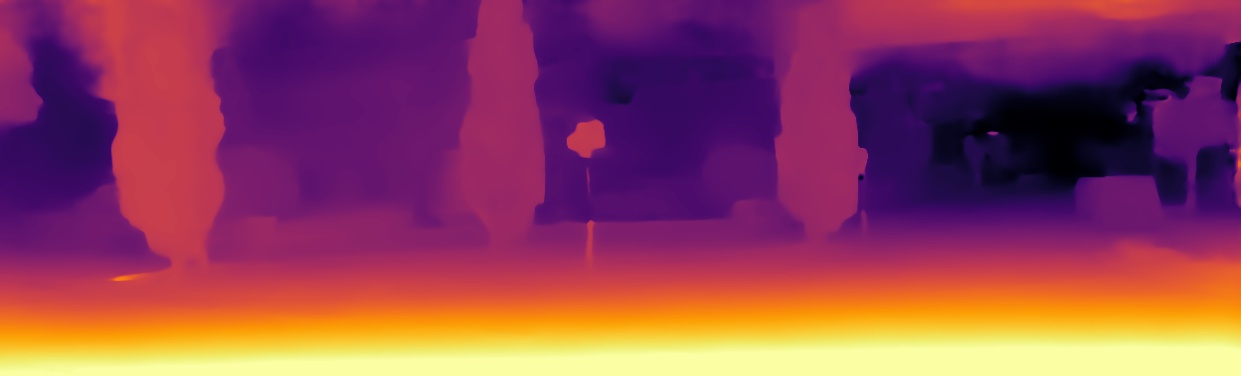} \\
\includegraphics[width=\itemwidth]{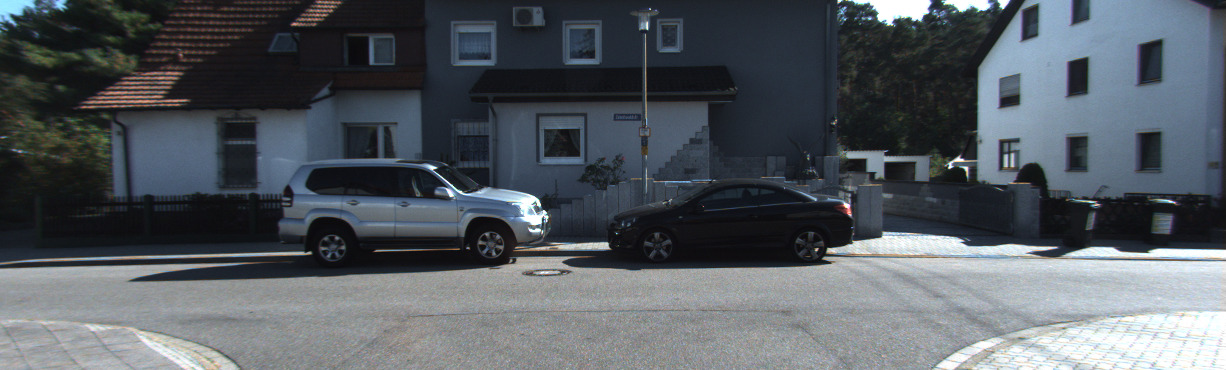} &
\includegraphics[width=\itemwidth]{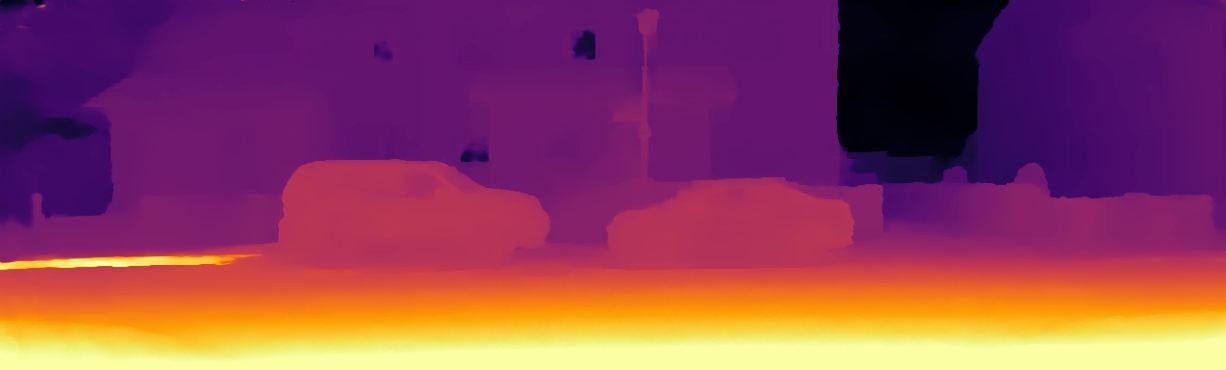} &
\includegraphics[width=\itemwidth]{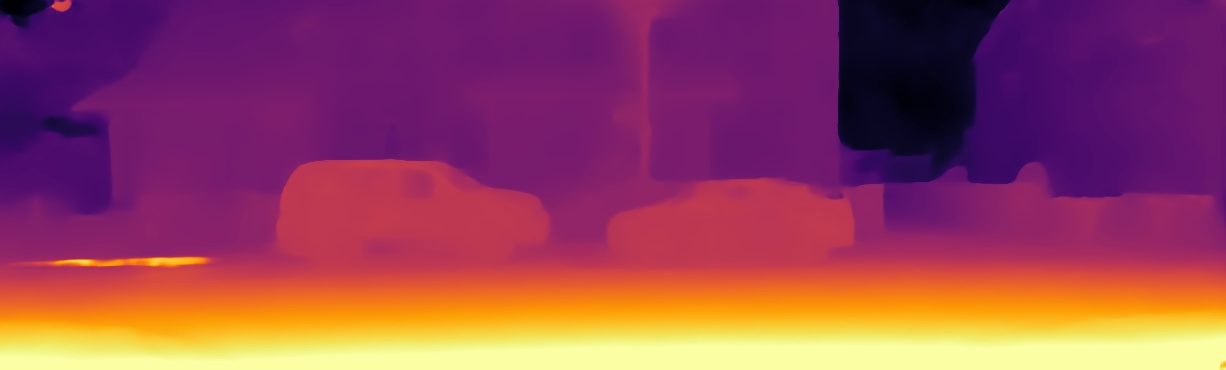} \\
\includegraphics[width=\itemwidth]{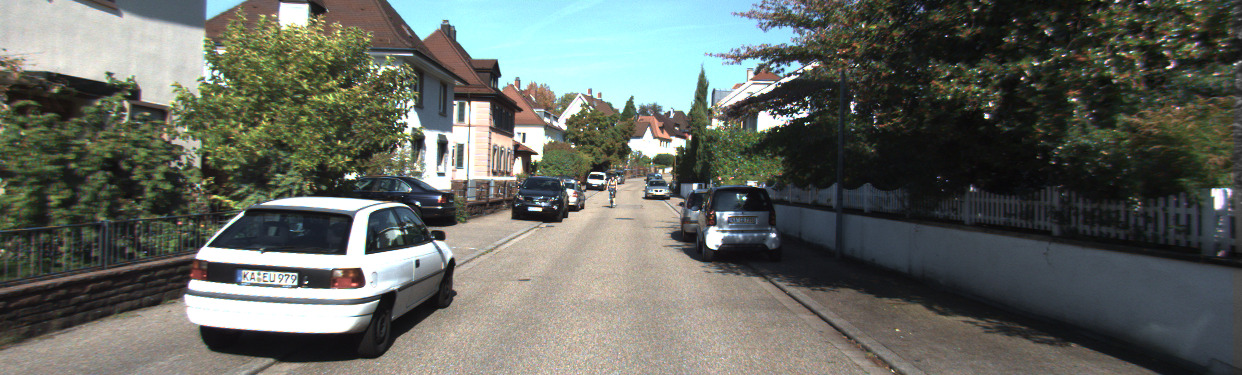} &
\includegraphics[width=\itemwidth]{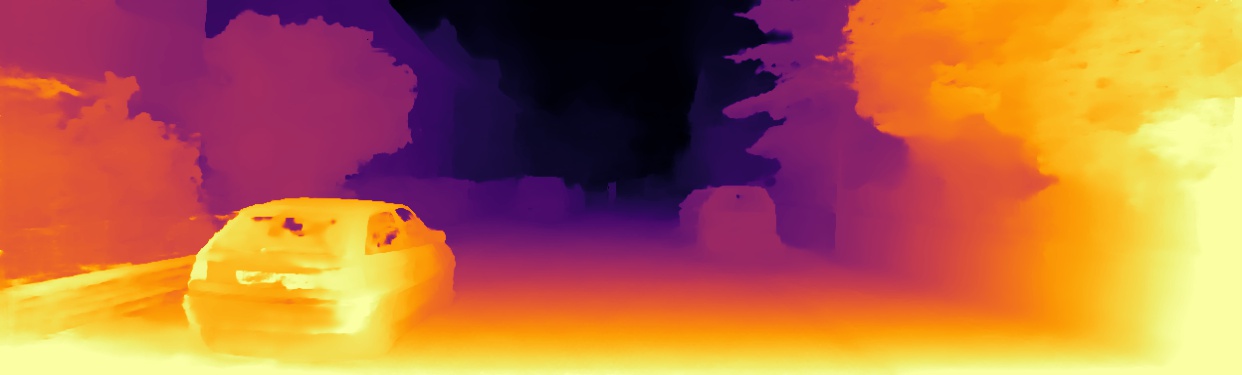} &
\includegraphics[width=\itemwidth]{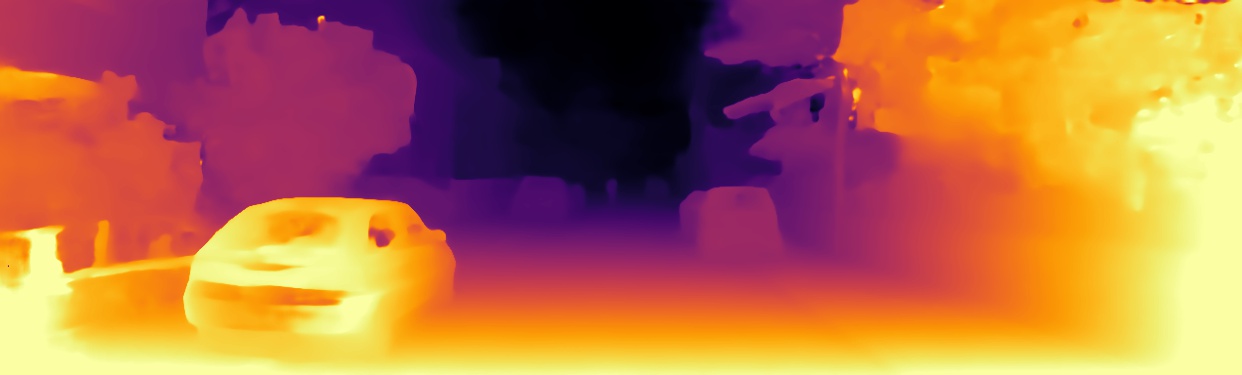} \\
\includegraphics[width=\itemwidth]{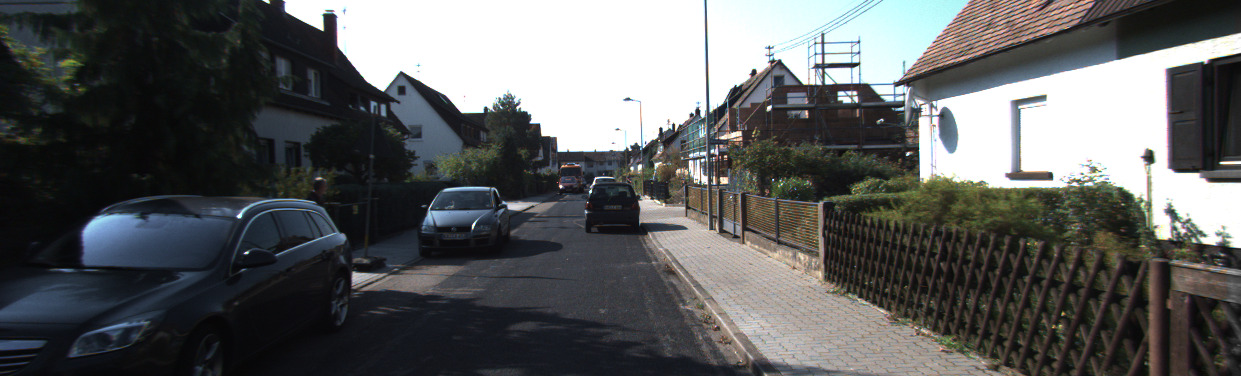} &
\includegraphics[width=\itemwidth]{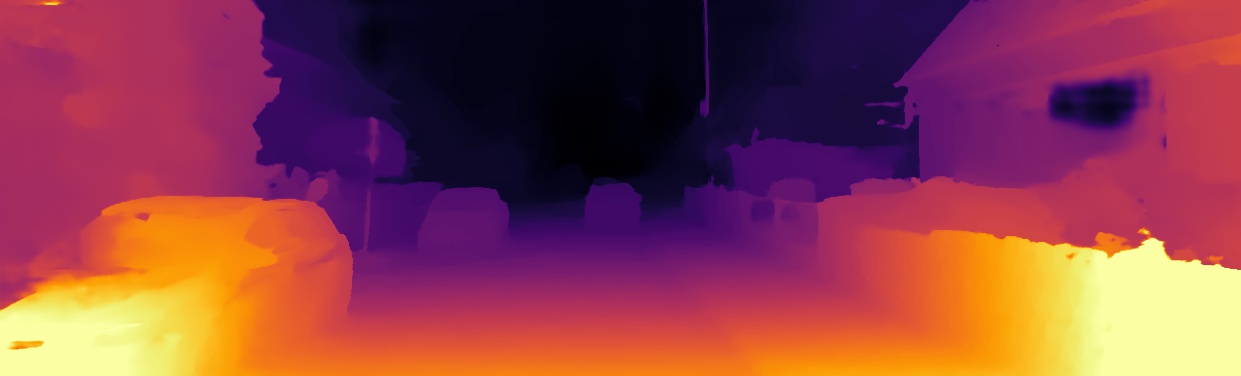} &
\includegraphics[width=\itemwidth]{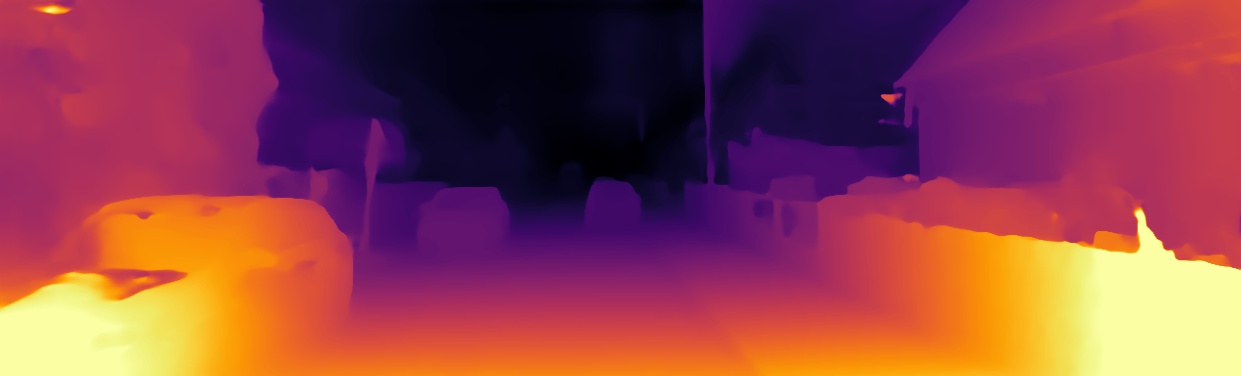} \\
\includegraphics[width=\itemwidth]{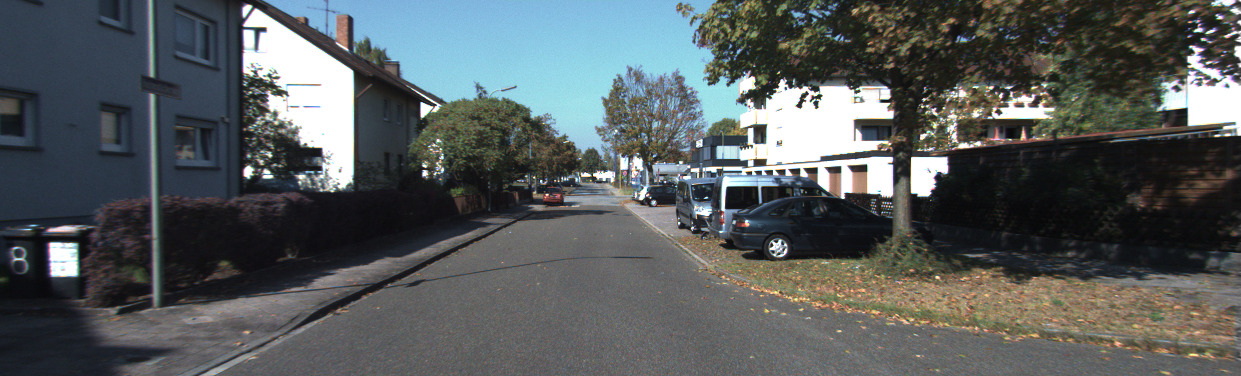} &
\includegraphics[width=\itemwidth]{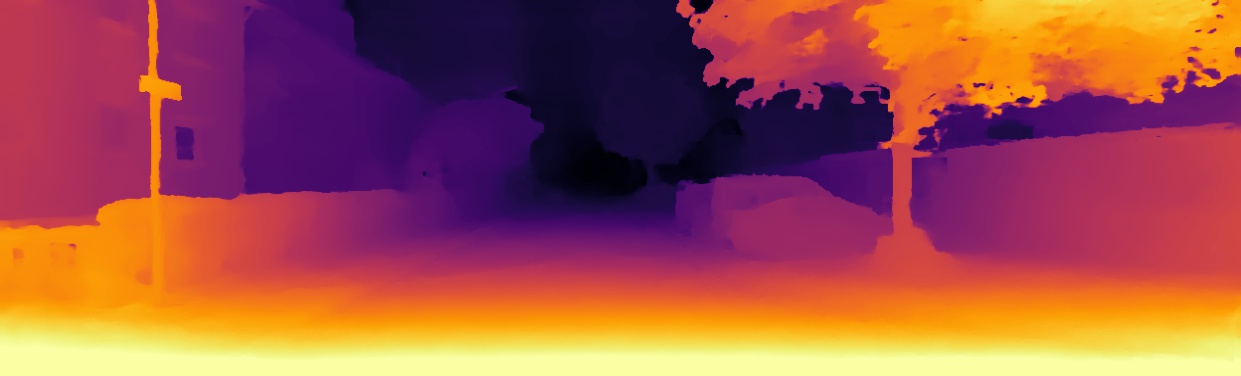} &
\includegraphics[width=\itemwidth]{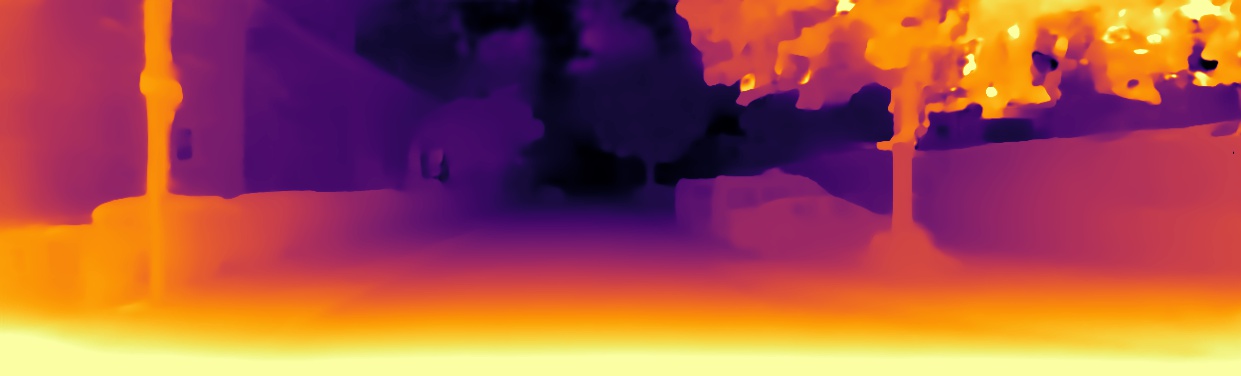} \\
\includegraphics[width=\itemwidth]{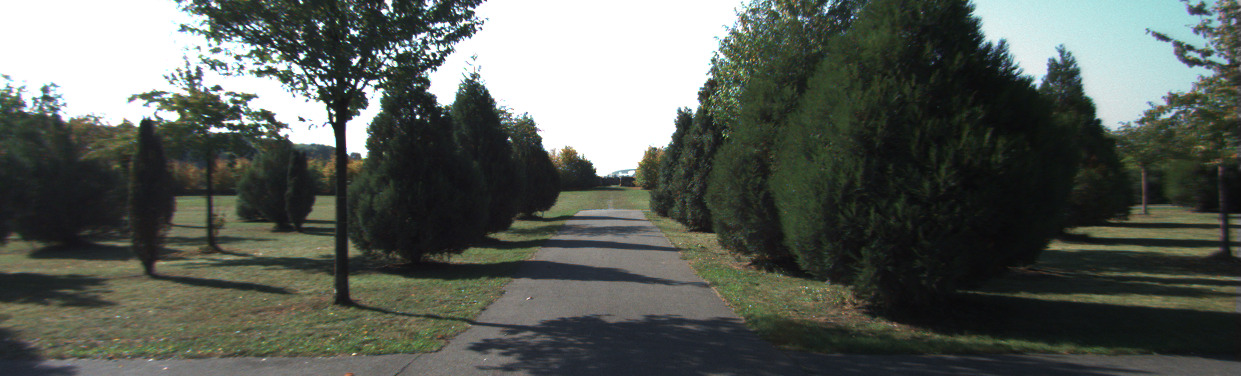} &
\includegraphics[width=\itemwidth]{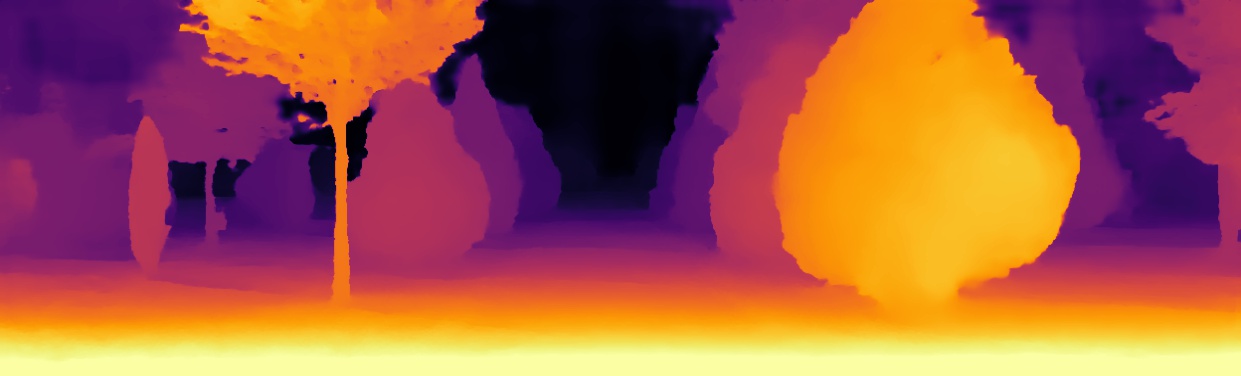} &
\includegraphics[width=\itemwidth]{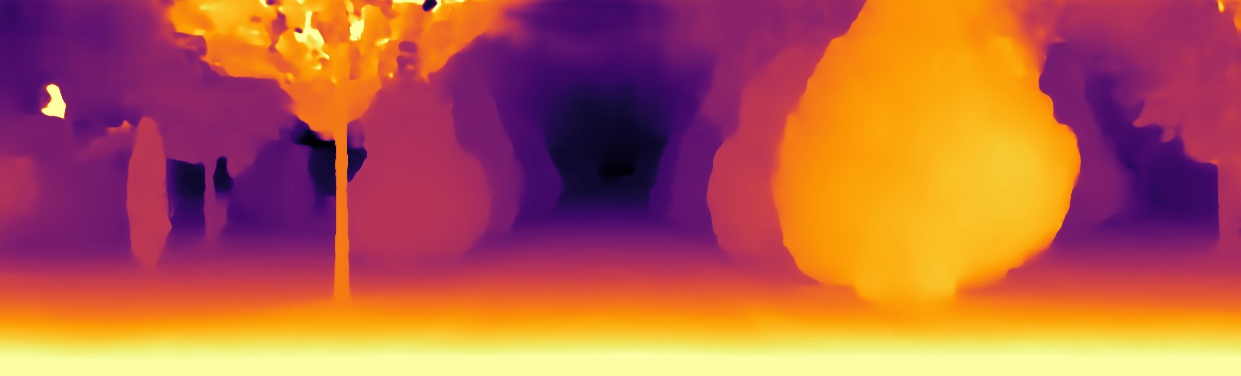} \\
\end{tabular}
\caption{\bf Additional GANet~\cite{Zhang2019GANet} KITTI 2012 qualitative results.}
\label{fig:kitti12-more-qual}
\end{figure}

\begin{figure}[!t]
\centering
\newcommand{\itemwidth}{0.35\columnwidth}
\begin{tabular}{@{\hskip 2mm}c@{\hskip 2mm}c@{\hskip 2mm}c@{\hskip 2mm}c@{}}
{\scriptsize Input (left) image} & {\scriptsize Trained with Sceneflow} & {\scriptsize Trained with our MfS} \\
\includegraphics[width=\itemwidth]{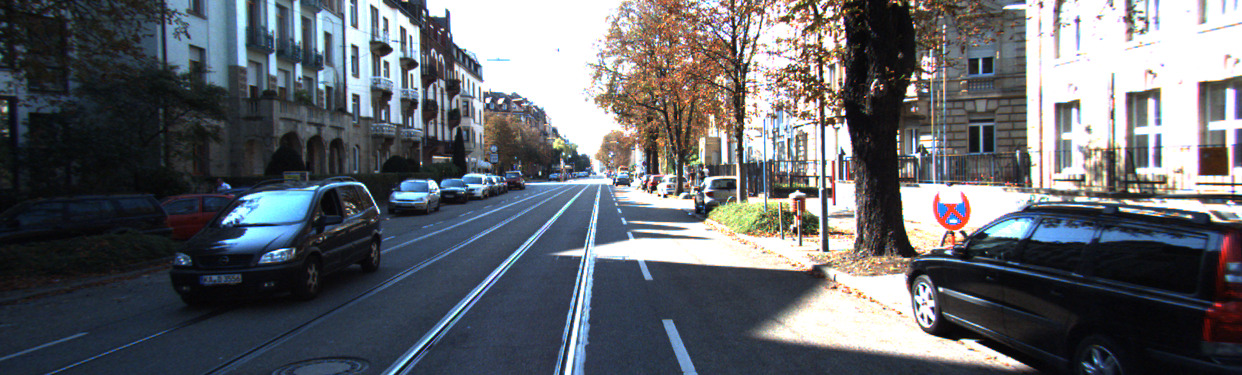} &
\includegraphics[width=\itemwidth]{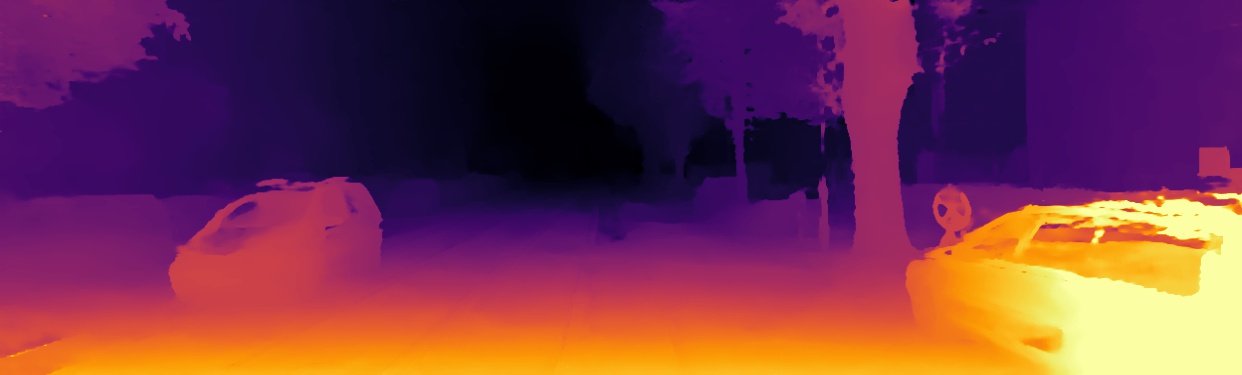} &
\includegraphics[width=\itemwidth]{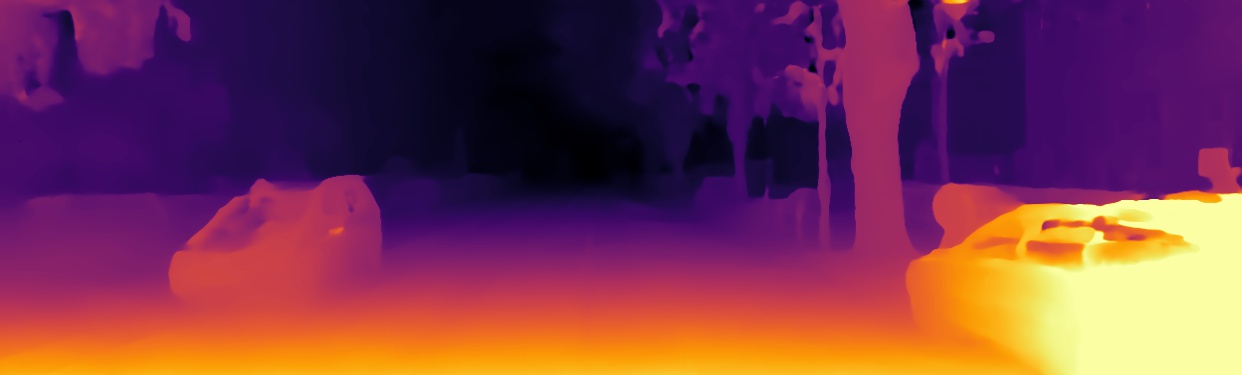} \\
\includegraphics[width=\itemwidth]{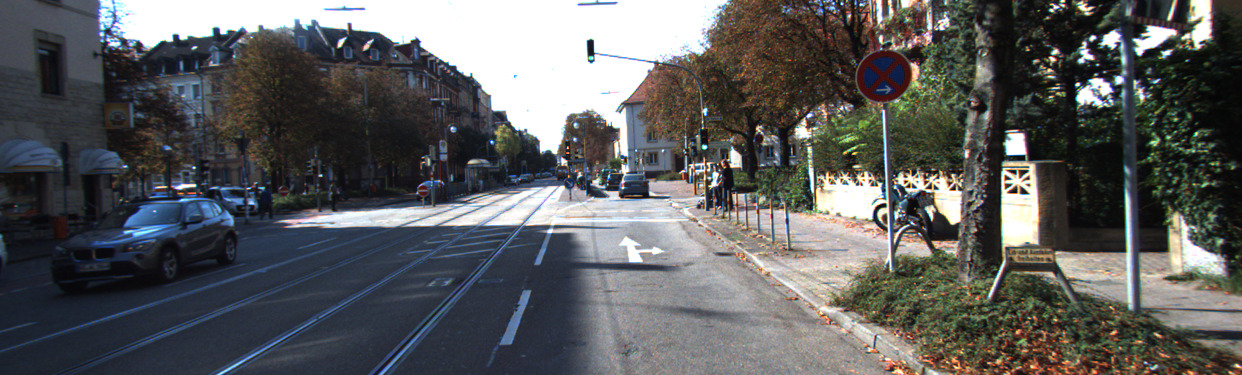} &
\includegraphics[width=\itemwidth]{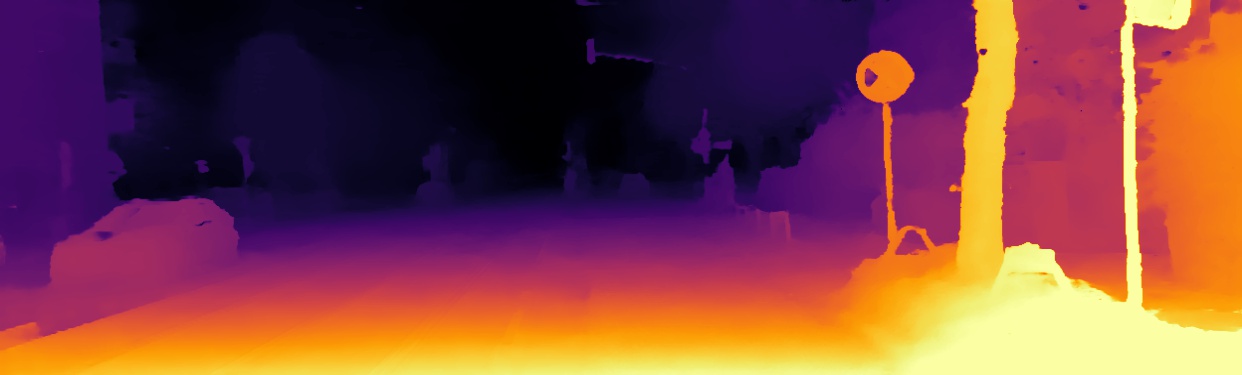} &
\includegraphics[width=\itemwidth]{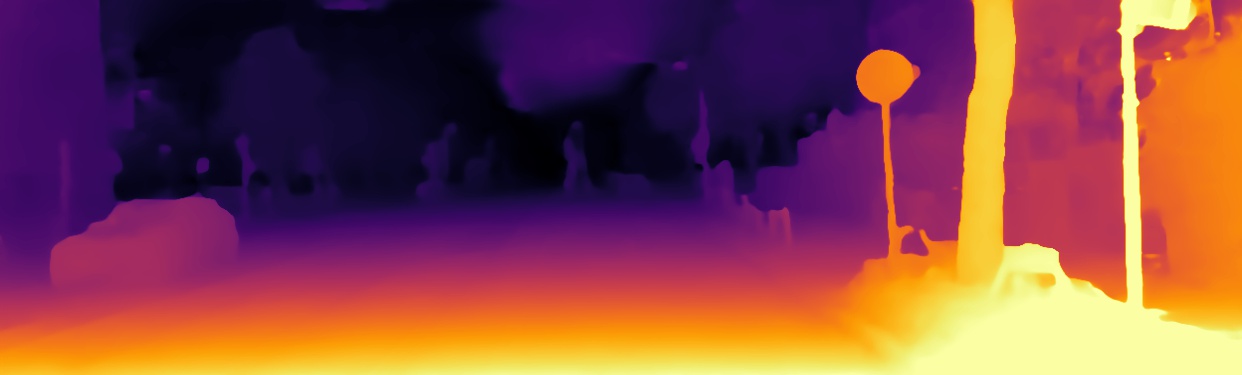} \\
\includegraphics[width=\itemwidth]{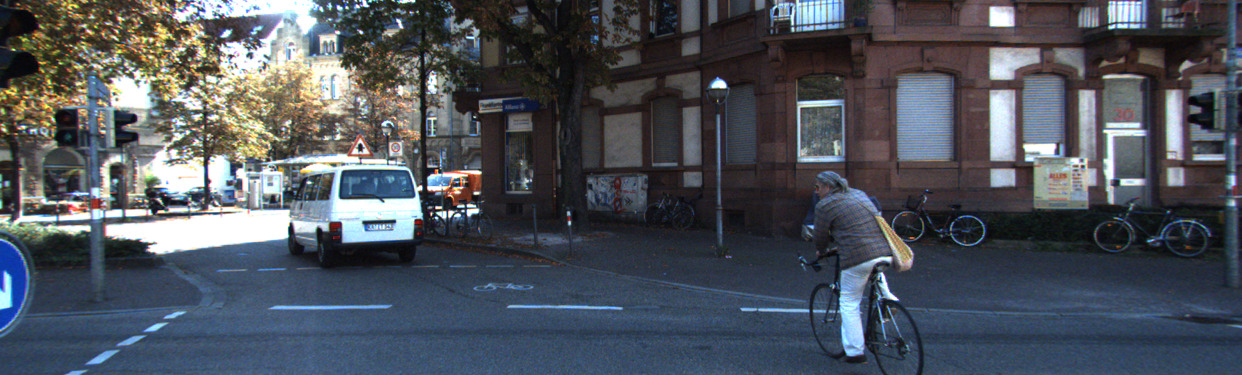} &
\includegraphics[width=\itemwidth]{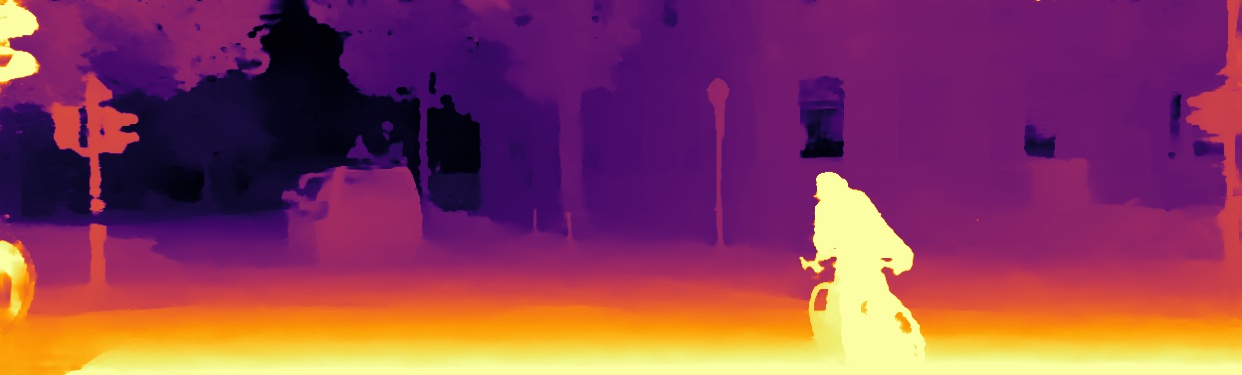} &
\includegraphics[width=\itemwidth]{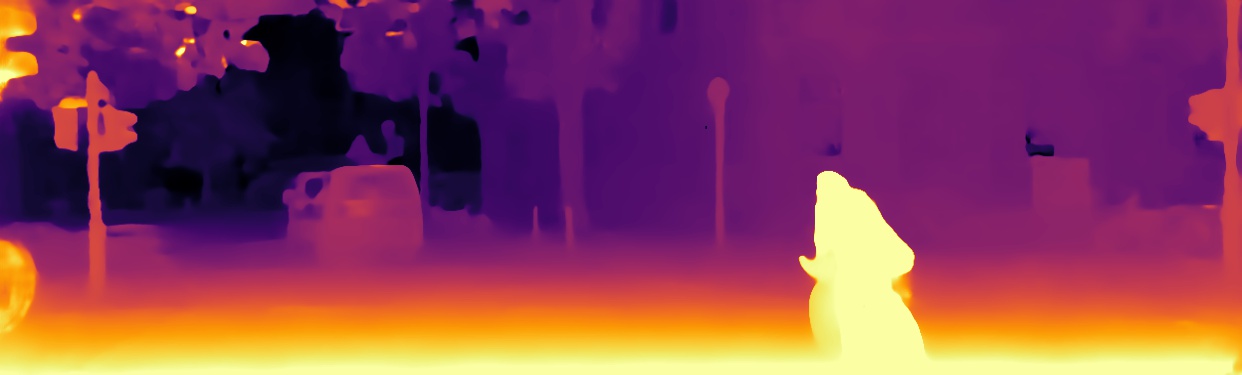} \\
\includegraphics[width=\itemwidth]{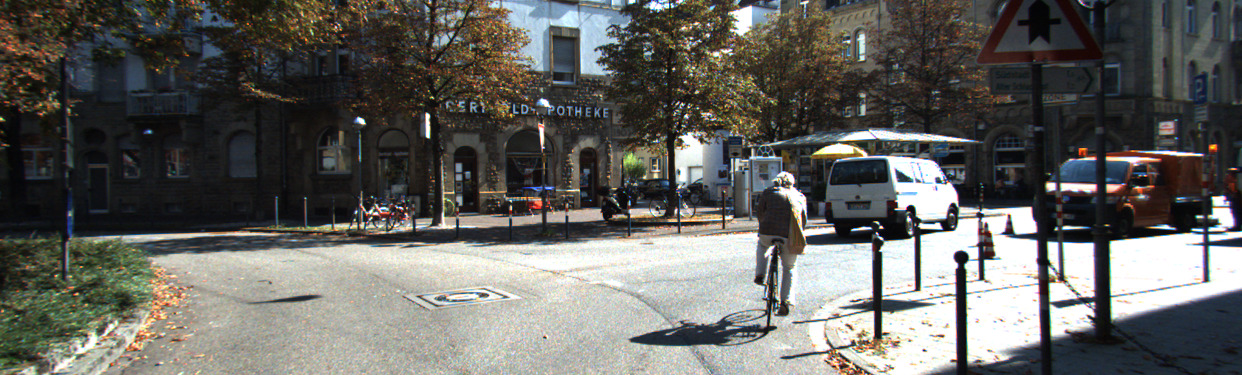} &
\includegraphics[width=\itemwidth]{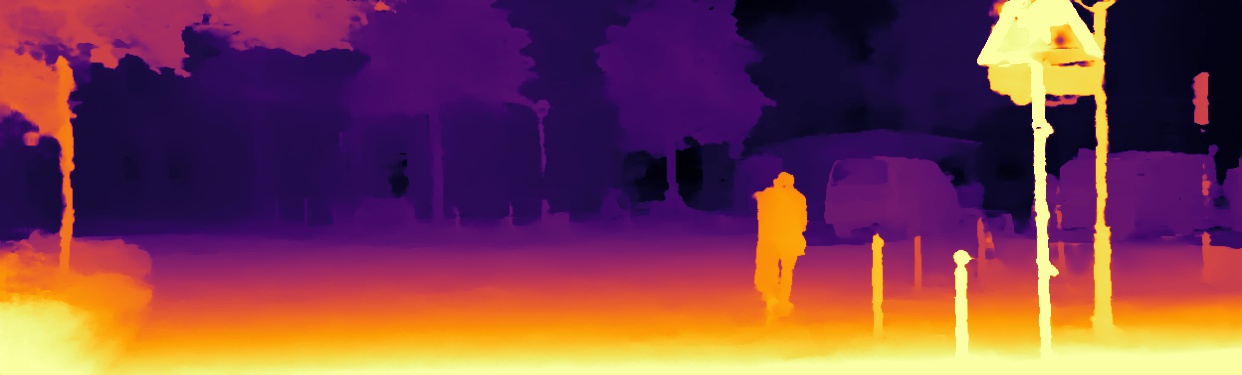} &
\includegraphics[width=\itemwidth]{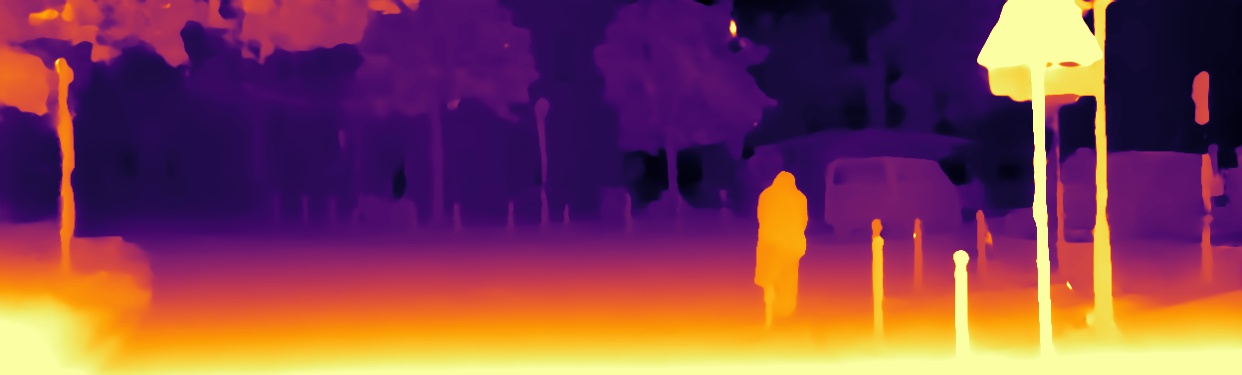} \\
\includegraphics[width=\itemwidth]{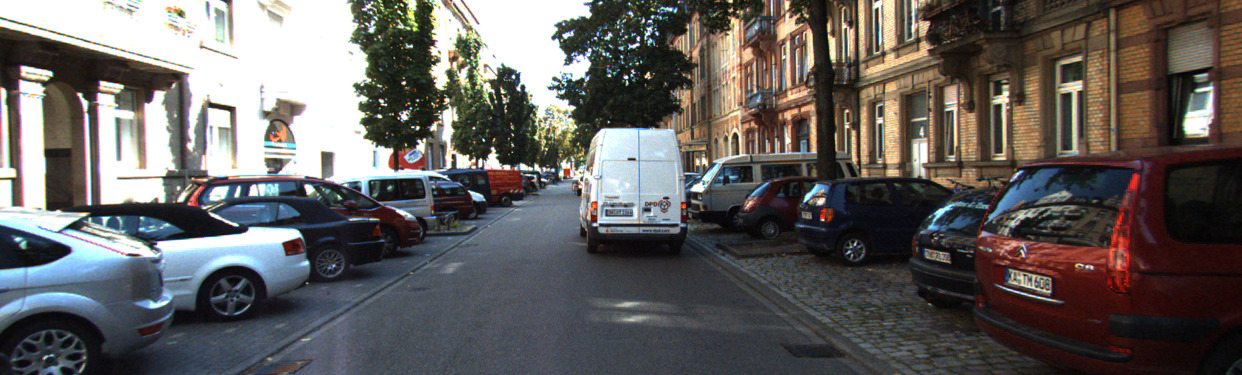} &
\includegraphics[width=\itemwidth]{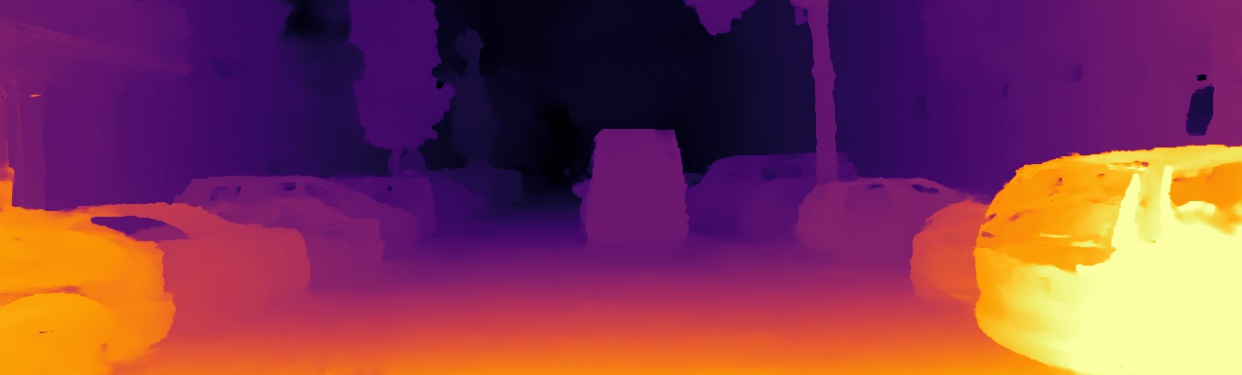} &
\includegraphics[width=\itemwidth]{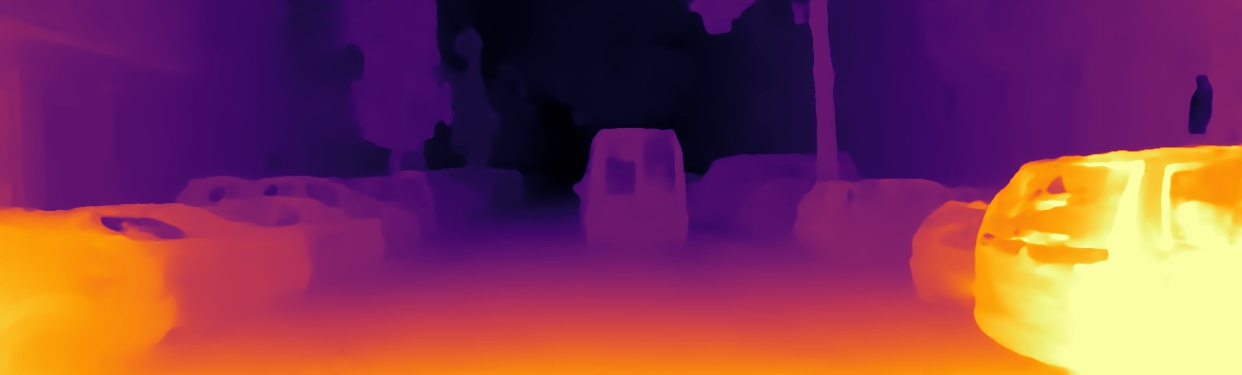} \\
\includegraphics[width=\itemwidth]{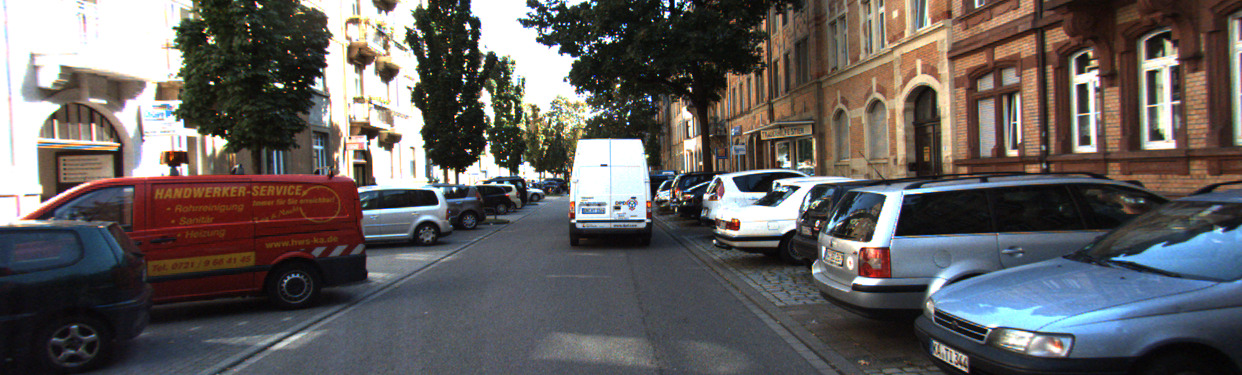} &
\includegraphics[width=\itemwidth]{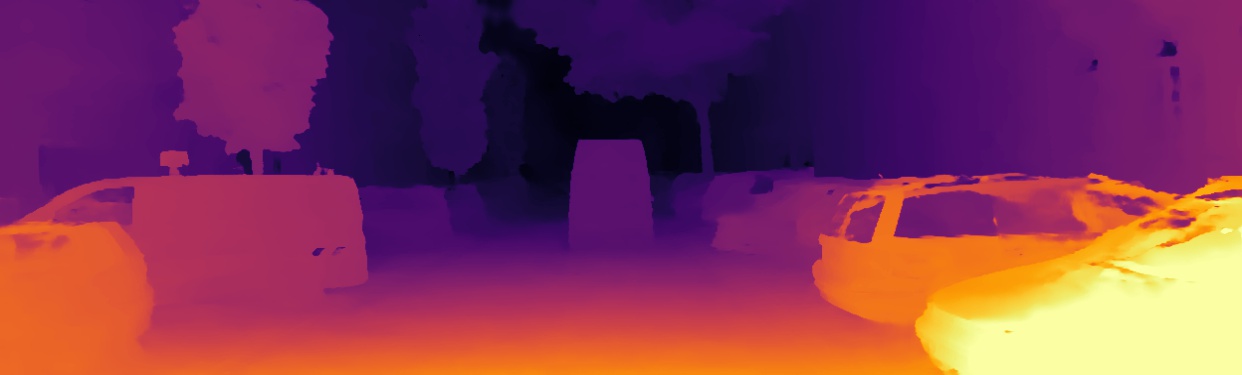} &
\includegraphics[width=\itemwidth]{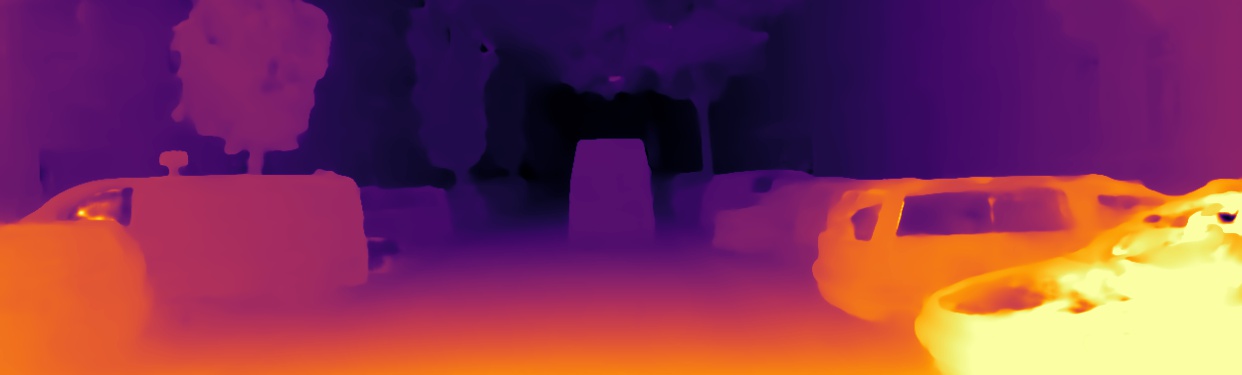} \\
\includegraphics[width=\itemwidth]{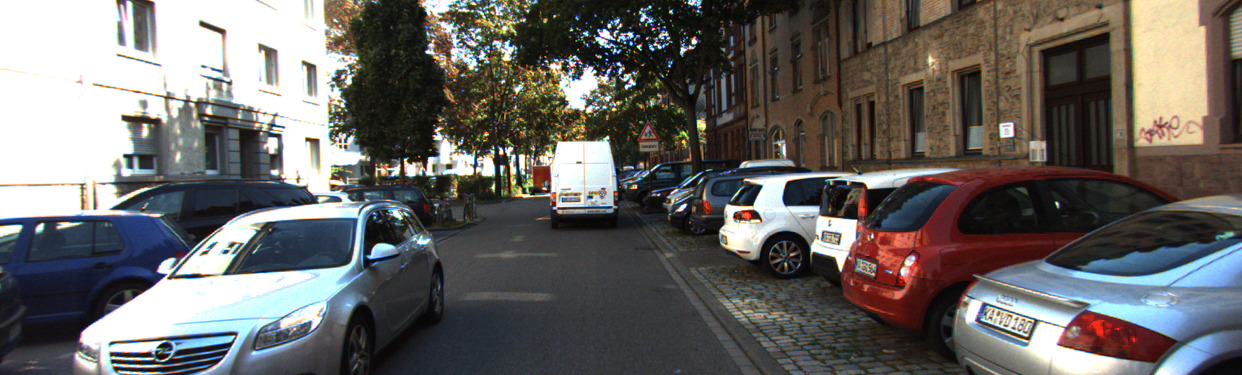} &
\includegraphics[width=\itemwidth]{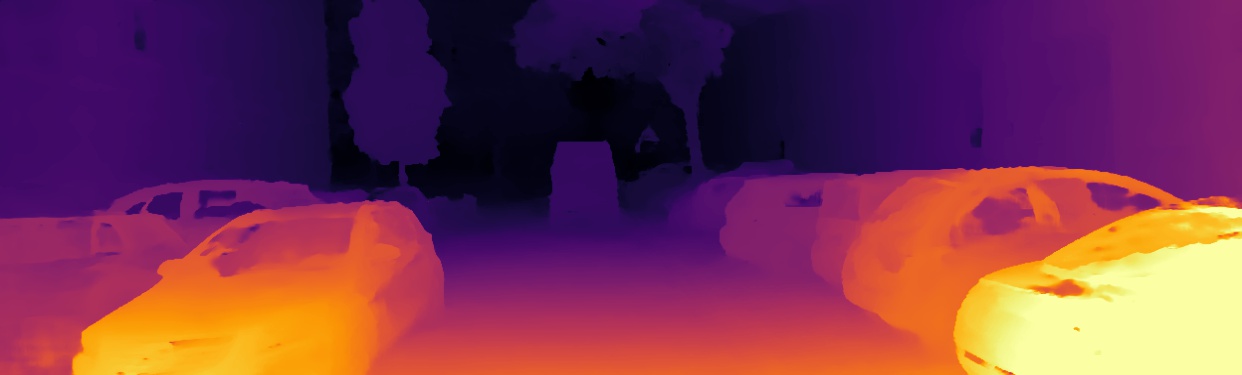} &
\includegraphics[width=\itemwidth]{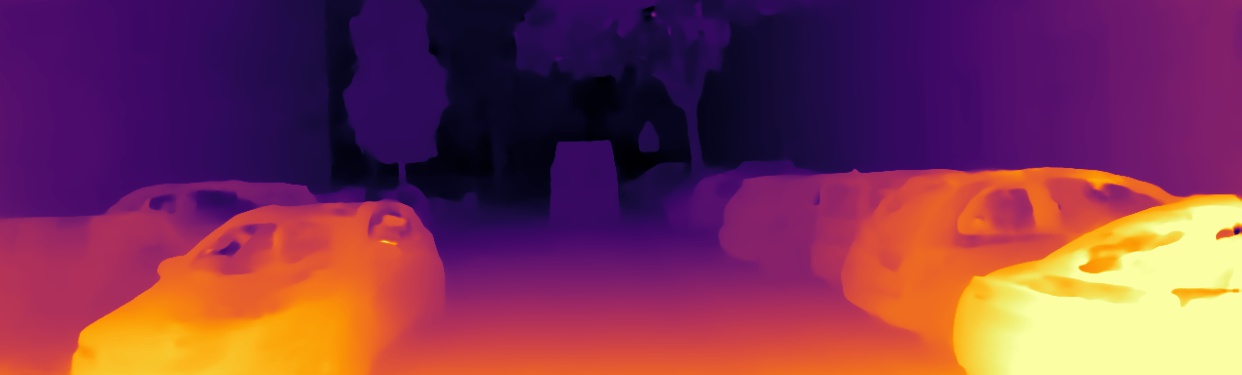} \\
\includegraphics[width=\itemwidth]{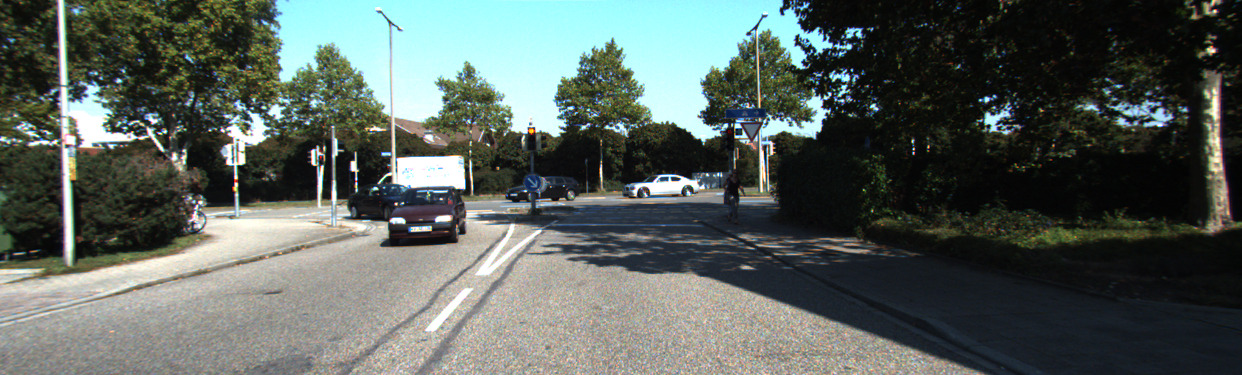} &
\includegraphics[width=\itemwidth]{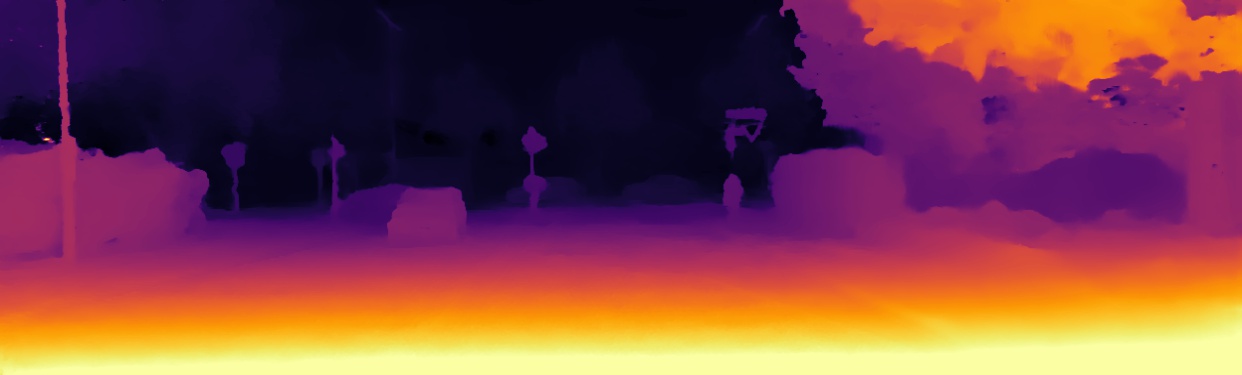} &
\includegraphics[width=\itemwidth]{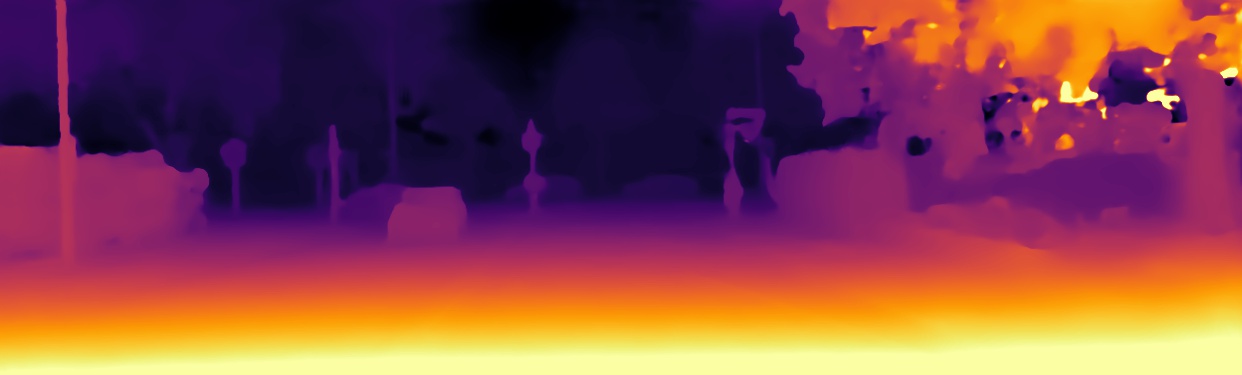} \\
\includegraphics[width=\itemwidth]{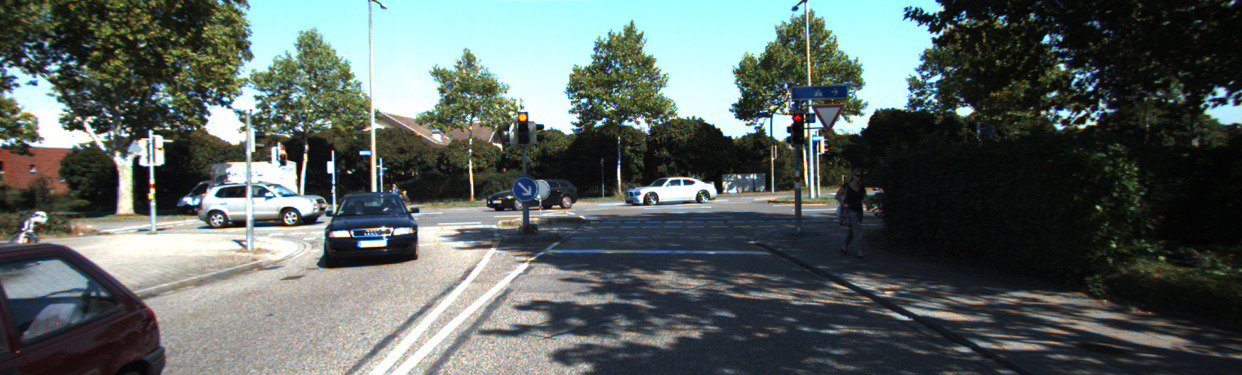} &
\includegraphics[width=\itemwidth]{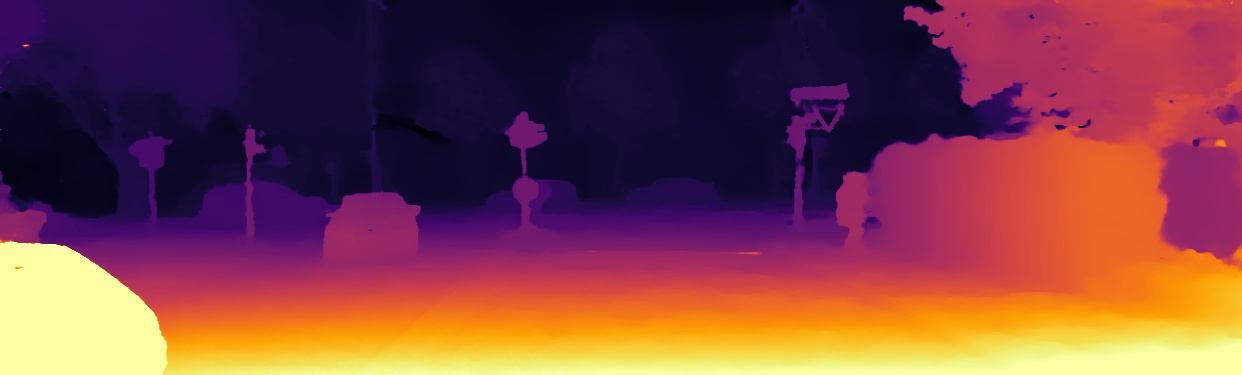} &
\includegraphics[width=\itemwidth]{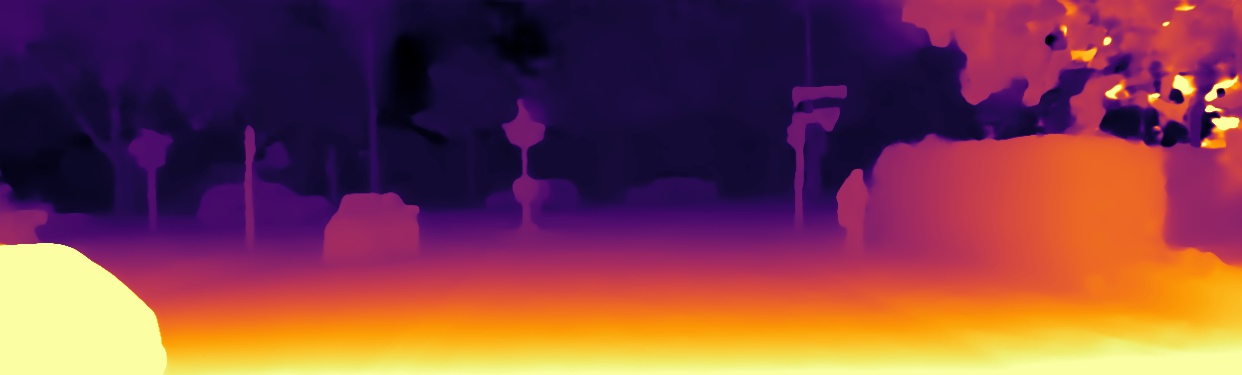} \\
\includegraphics[width=\itemwidth]{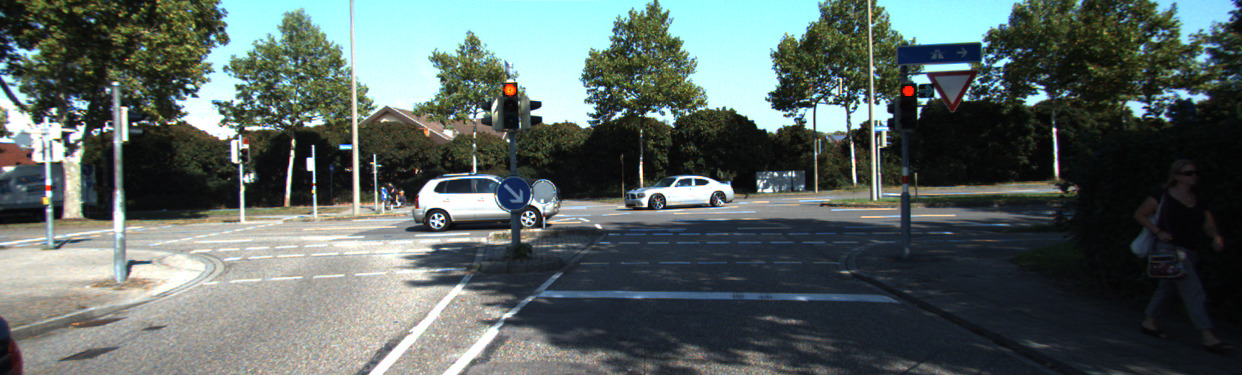} &
\includegraphics[width=\itemwidth]{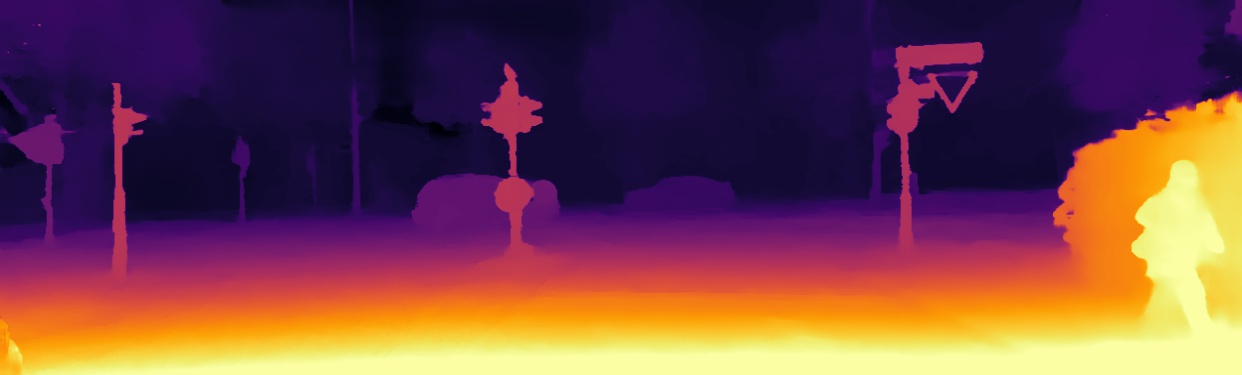} &
\includegraphics[width=\itemwidth]{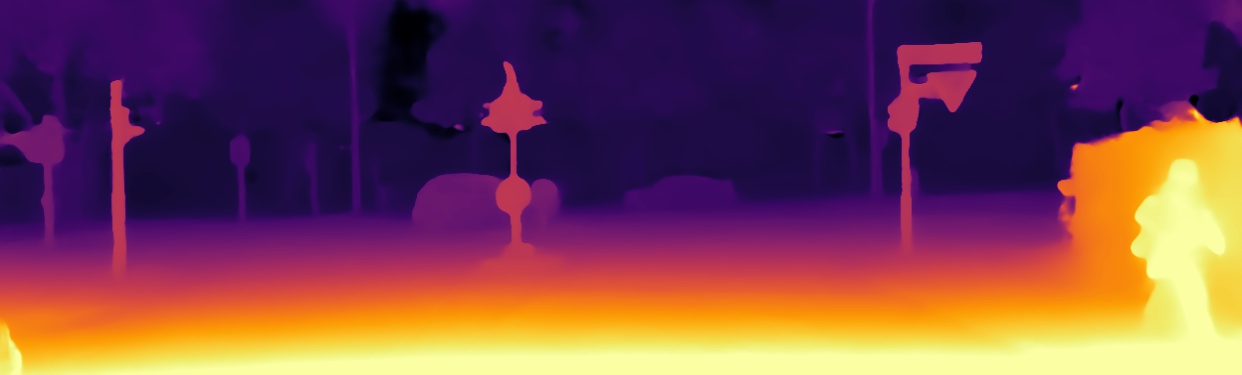} \\
\end{tabular}
\caption{\bf Additional GANet~\cite{Zhang2019GANet} KITTI 2015 qualitative results.}
\label{fig:kitti15-more-qual}
\end{figure}

\begin{figure}[!t]
\centering
\newcommand{\itemwidth}{0.25\columnwidth}
\begin{tabular}{@{\hskip 2mm}c@{\hskip 2mm}c@{\hskip 2mm}c@{\hskip 2mm}c@{}}
{\scriptsize Input (left) image} & {\scriptsize Trained with Sceneflow} & {\scriptsize Trained ours} & {\scriptsize Trained ours } \\
& & {\scriptsize  (MfS + MiDaS~\cite{lasinger2019towards})} & {\scriptsize (MfS + MegaDepth~\cite{li2018megadepth})} \\
\includegraphics[width=\itemwidth]{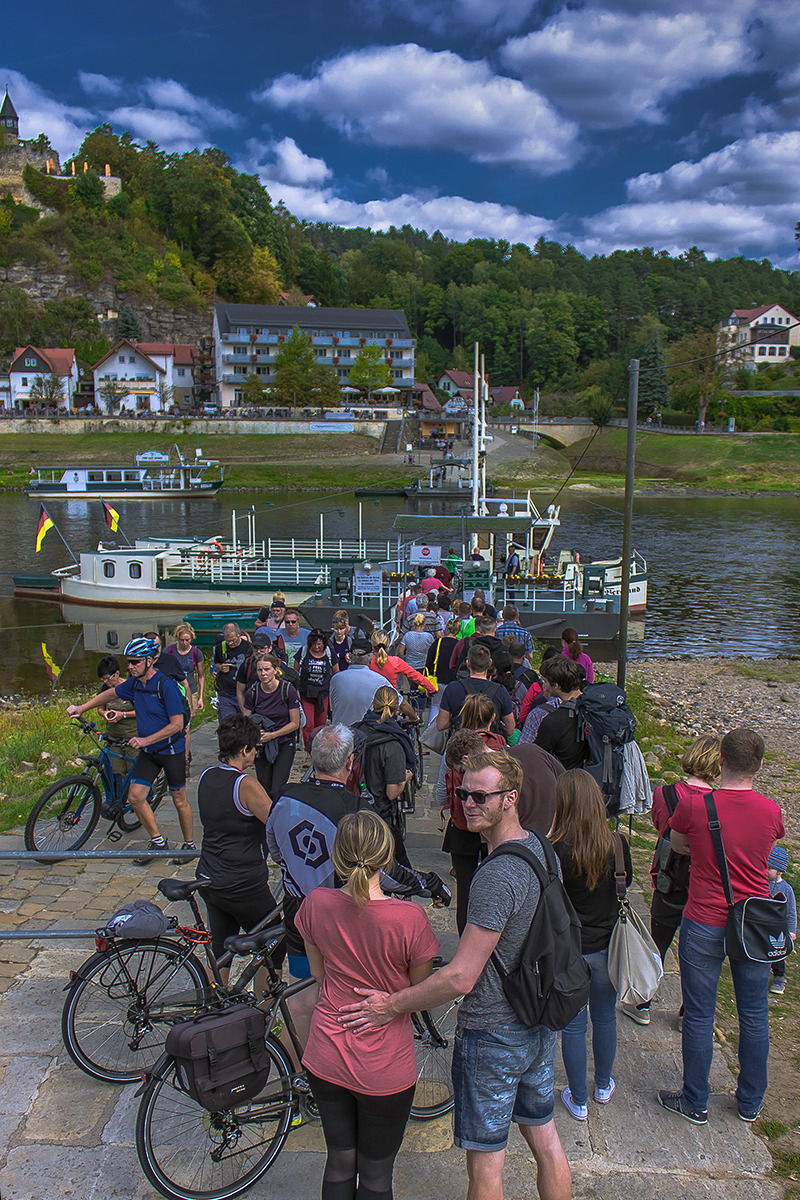} &
\includegraphics[width=\itemwidth]{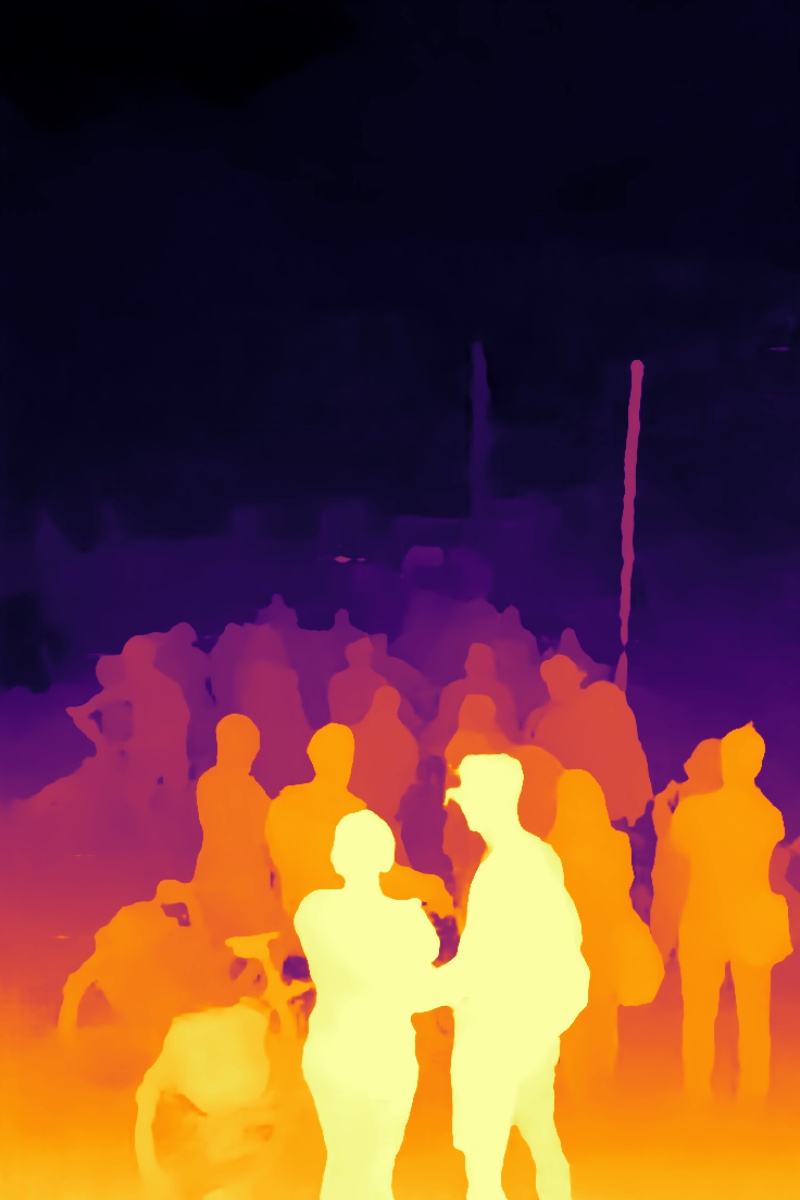} &
\includegraphics[width=\itemwidth]{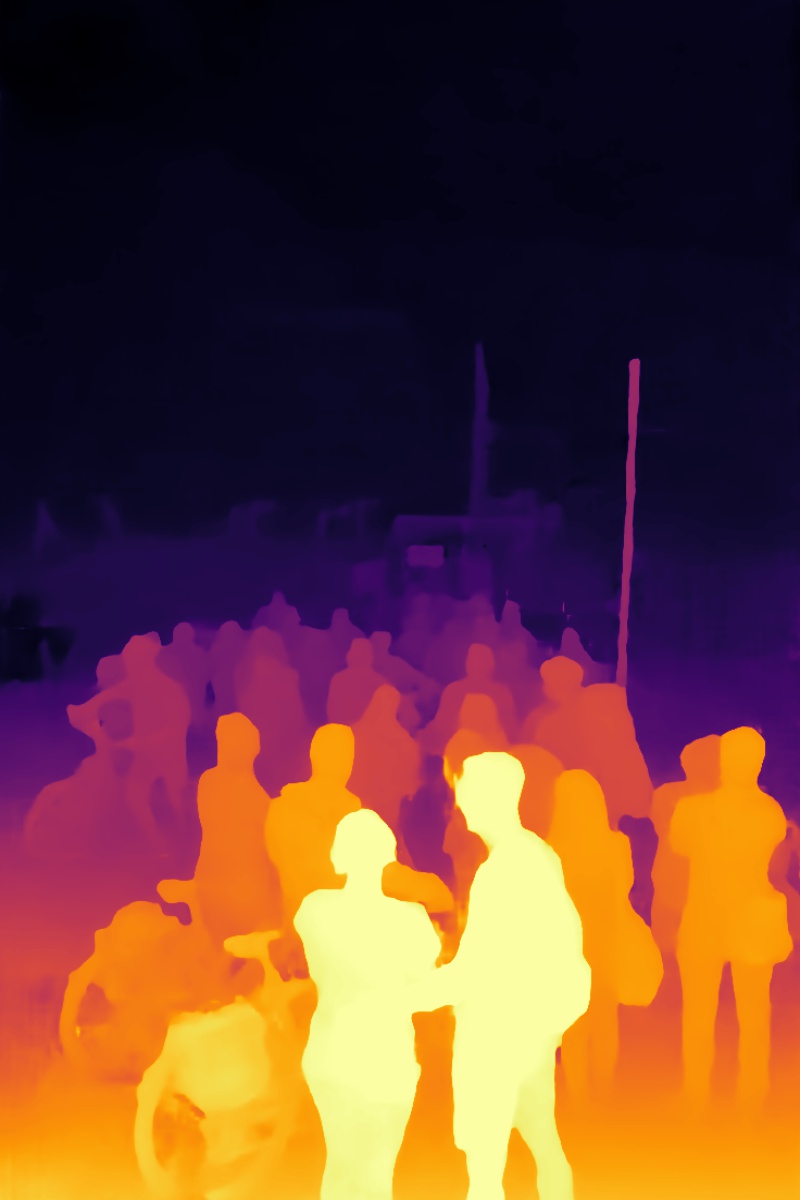} &
\includegraphics[width=\itemwidth]{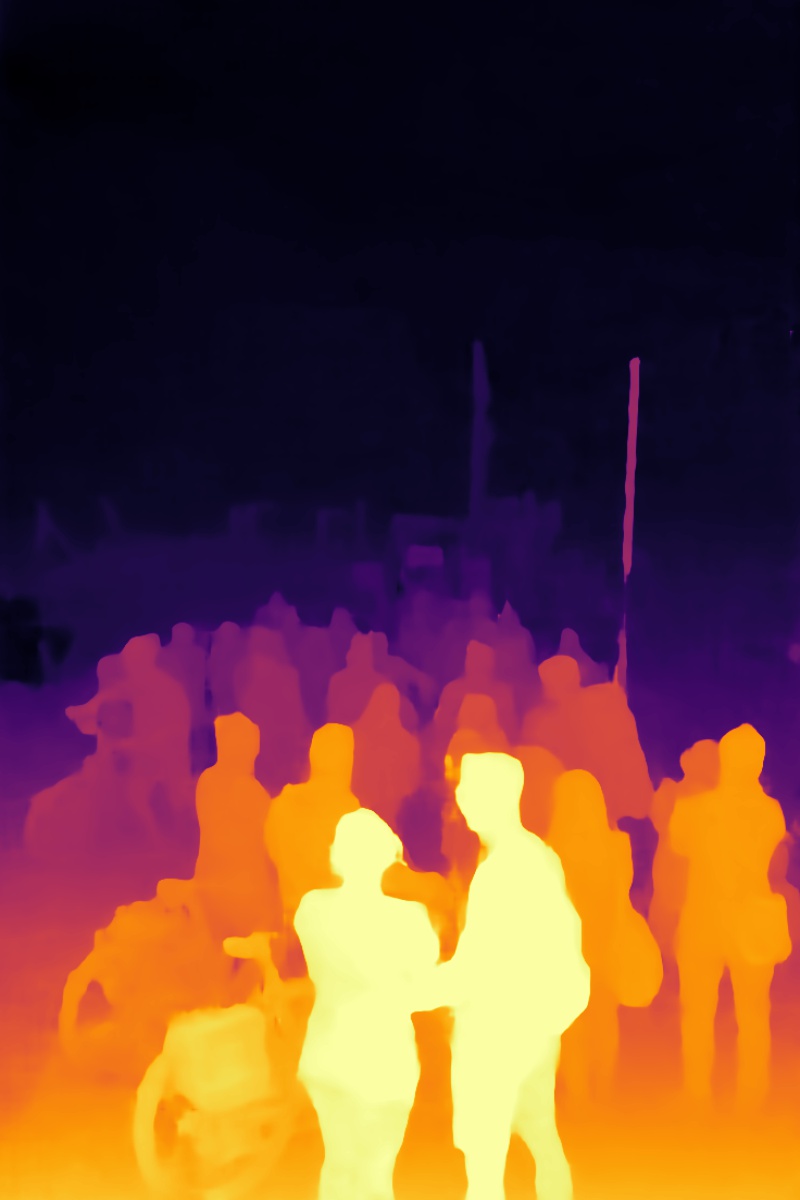} \\
\includegraphics[width=\itemwidth]{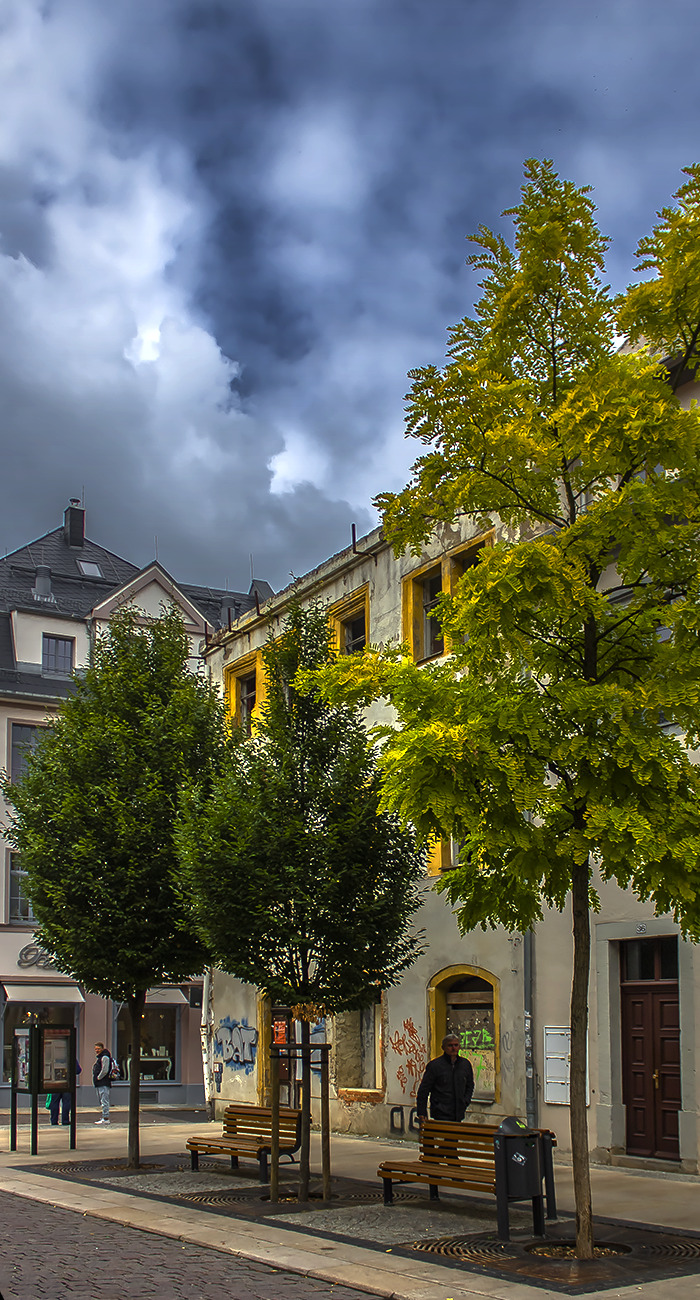} &
\includegraphics[width=\itemwidth]{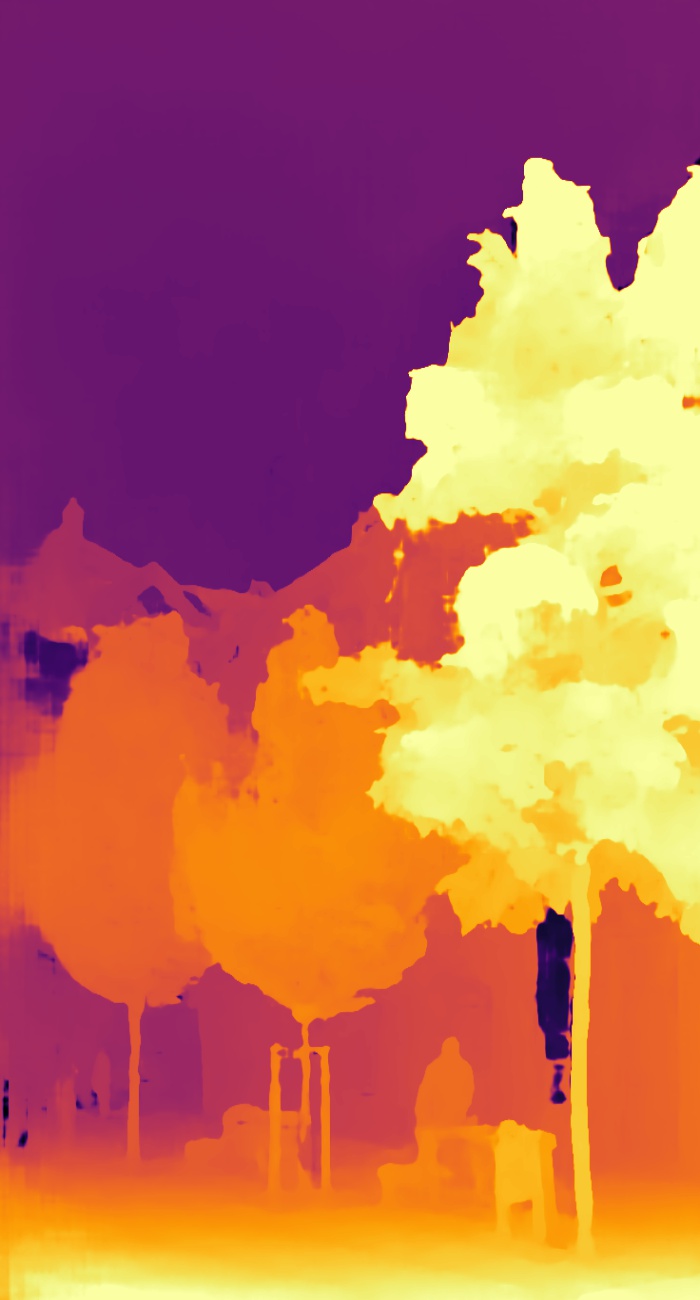} &
\includegraphics[width=\itemwidth]{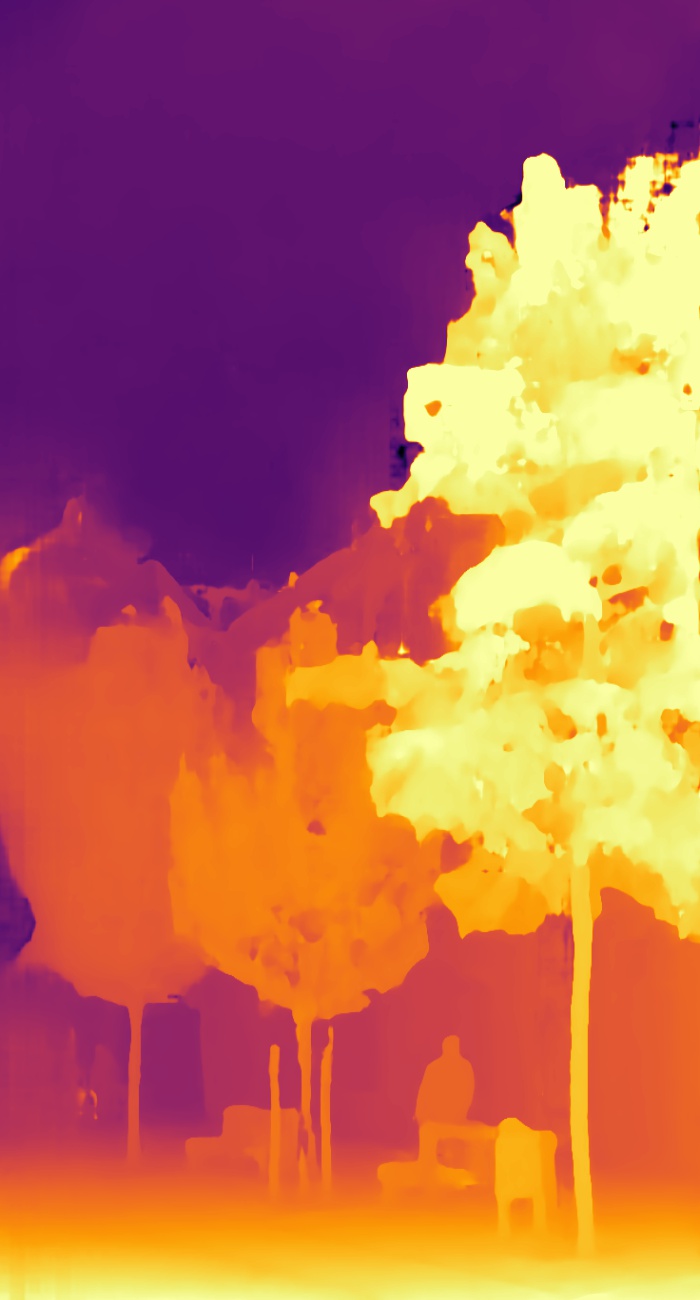} &
\includegraphics[width=\itemwidth]{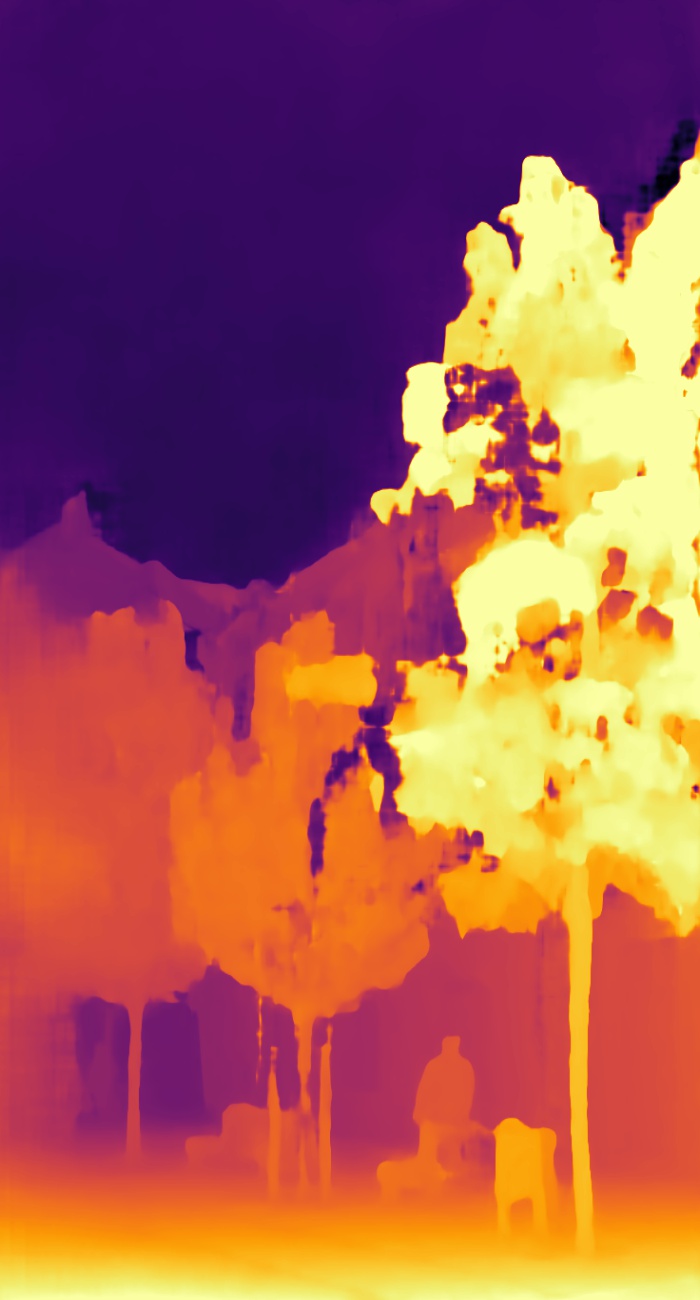} \\
\includegraphics[width=\itemwidth]{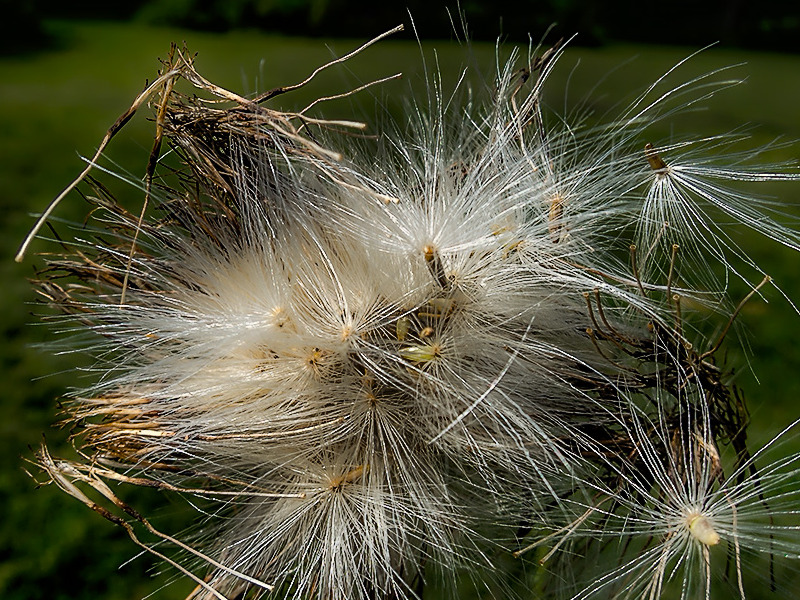} &
\includegraphics[width=\itemwidth]{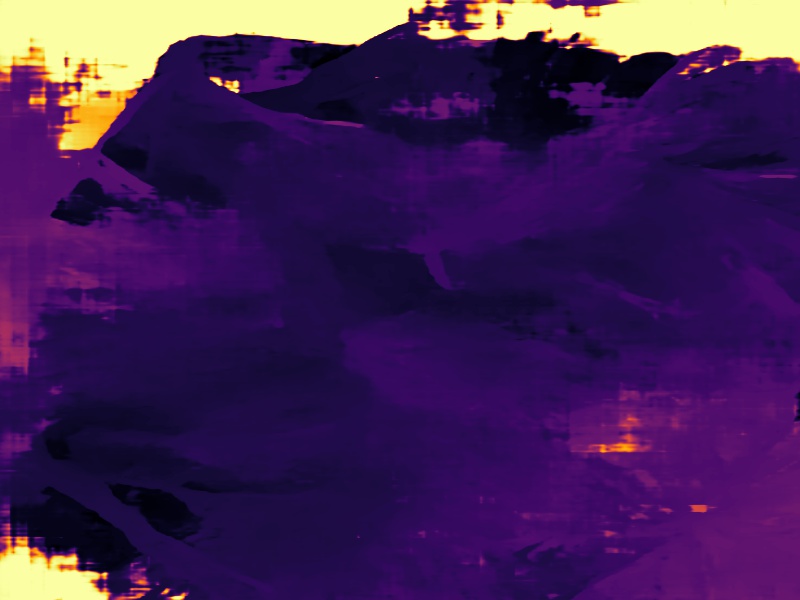} &
\includegraphics[width=\itemwidth]{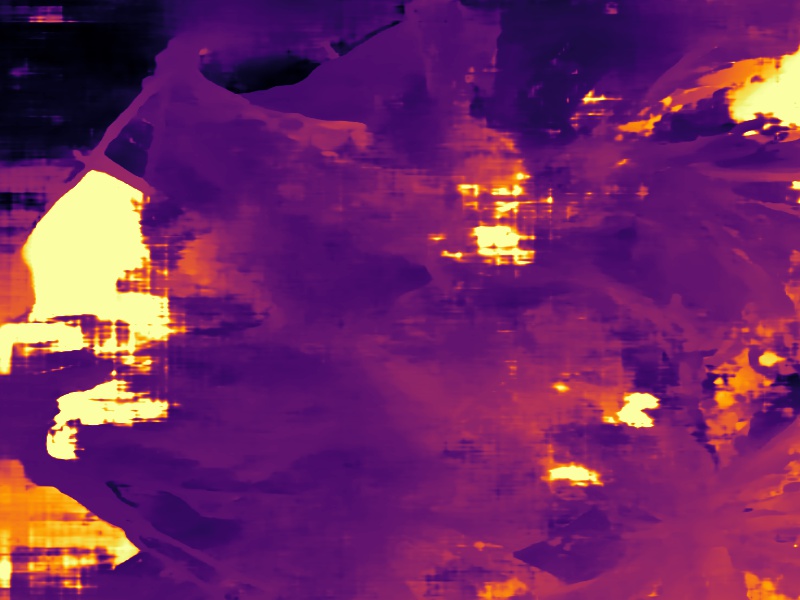} &
\includegraphics[width=\itemwidth]{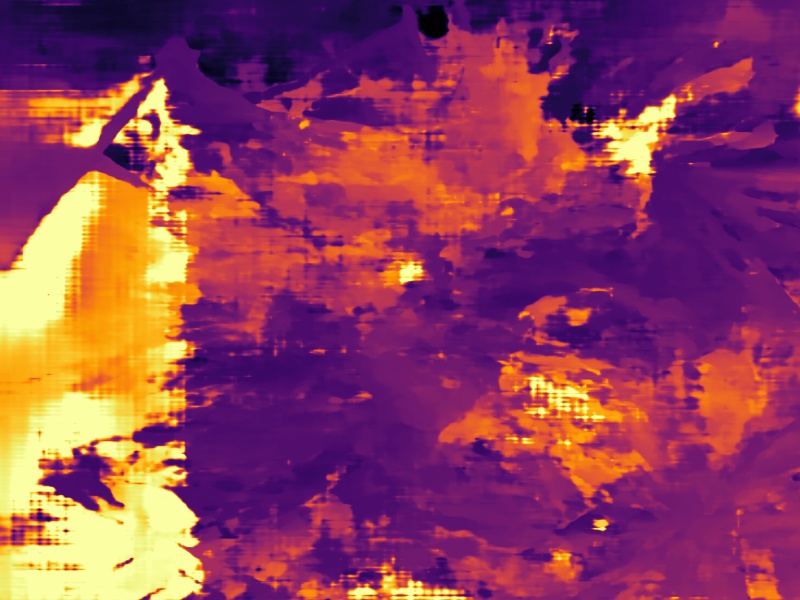} \\
\end{tabular}
\caption{\bf Additional Flickr1024~\cite{flickr1024} qualitative results.}
\label{fig:flickr1024-1-more-qual}
\end{figure}

\begin{figure}[!t]
\centering
\newcommand{\itemwidth}{0.25\columnwidth}
\centering
\begin{tabular}{@{\hskip 2mm}c@{\hskip 2mm}c@{\hskip 2mm}c@{\hskip 2mm}c@{}}
{\scriptsize Input (left) image} & {\scriptsize Trained with Sceneflow} & {\scriptsize Trained ours} & {\scriptsize Trained ours } \\
& & {\scriptsize  (MfS + MiDaS~\cite{lasinger2019towards})} & {\scriptsize (MfS + MegaDepth~\cite{li2018megadepth})} \\
\includegraphics[width=\itemwidth]{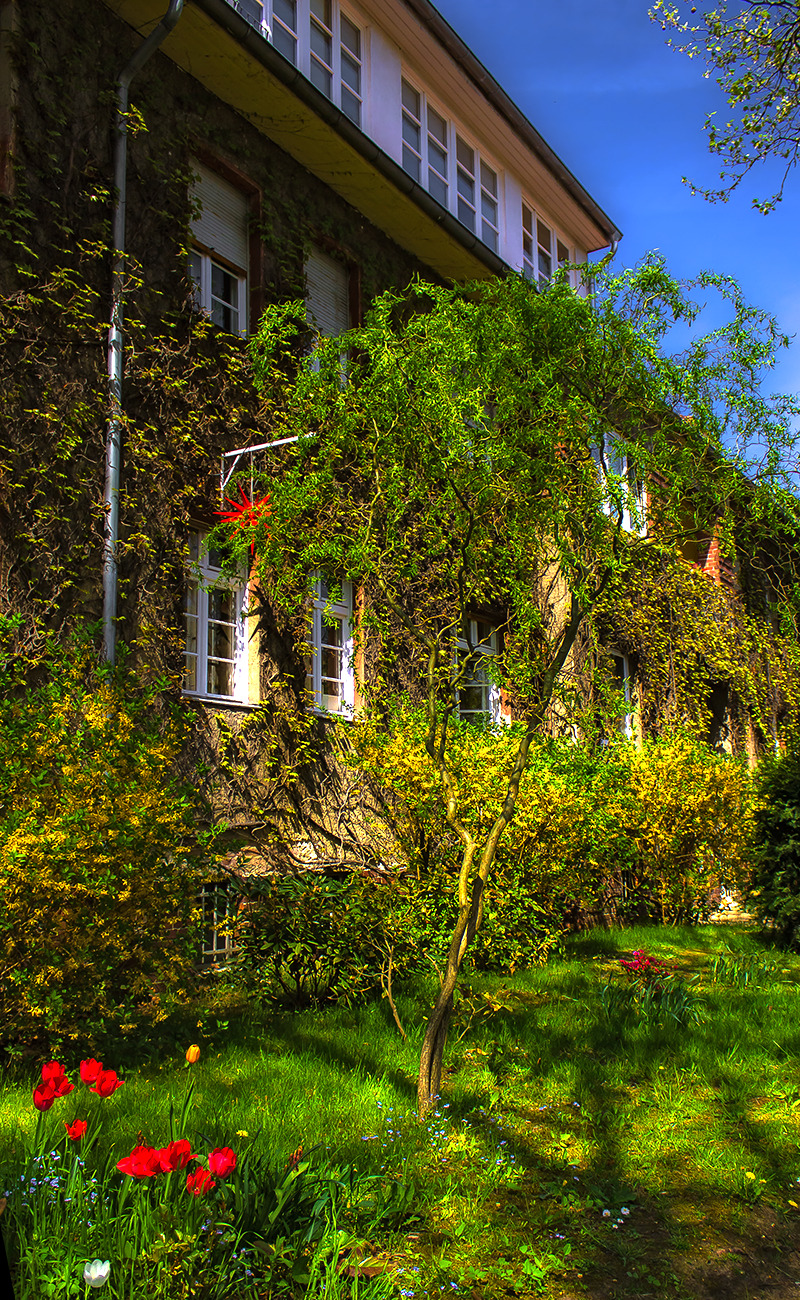} &
\includegraphics[width=\itemwidth]{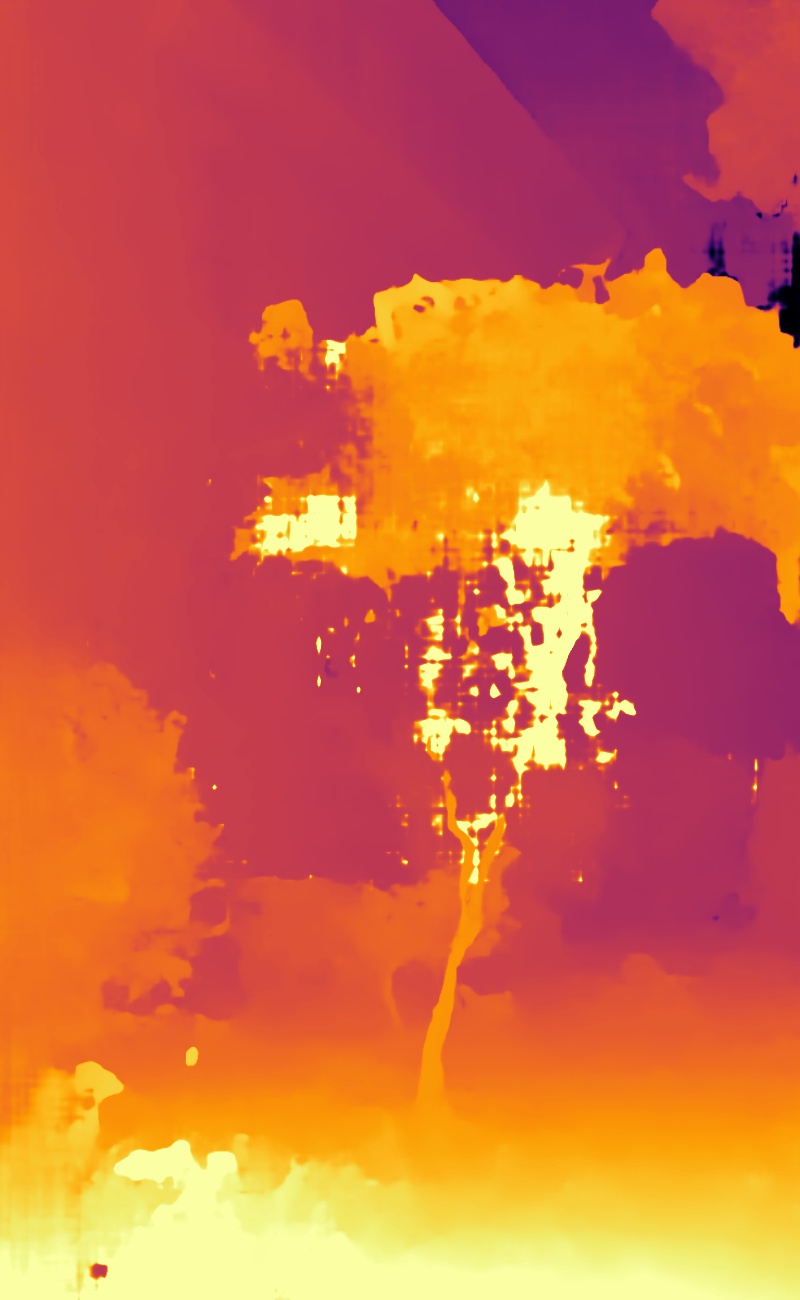} &
\includegraphics[width=\itemwidth]{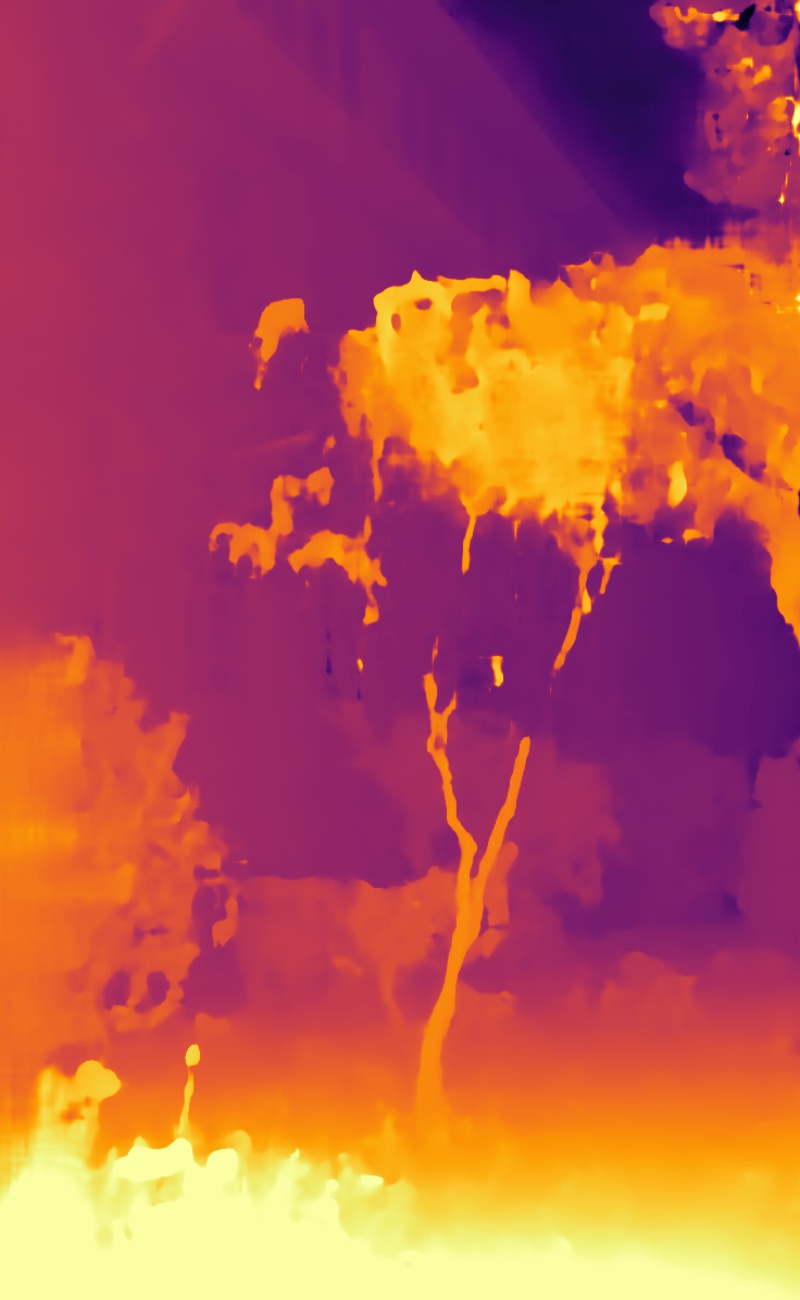} &
\includegraphics[width=\itemwidth]{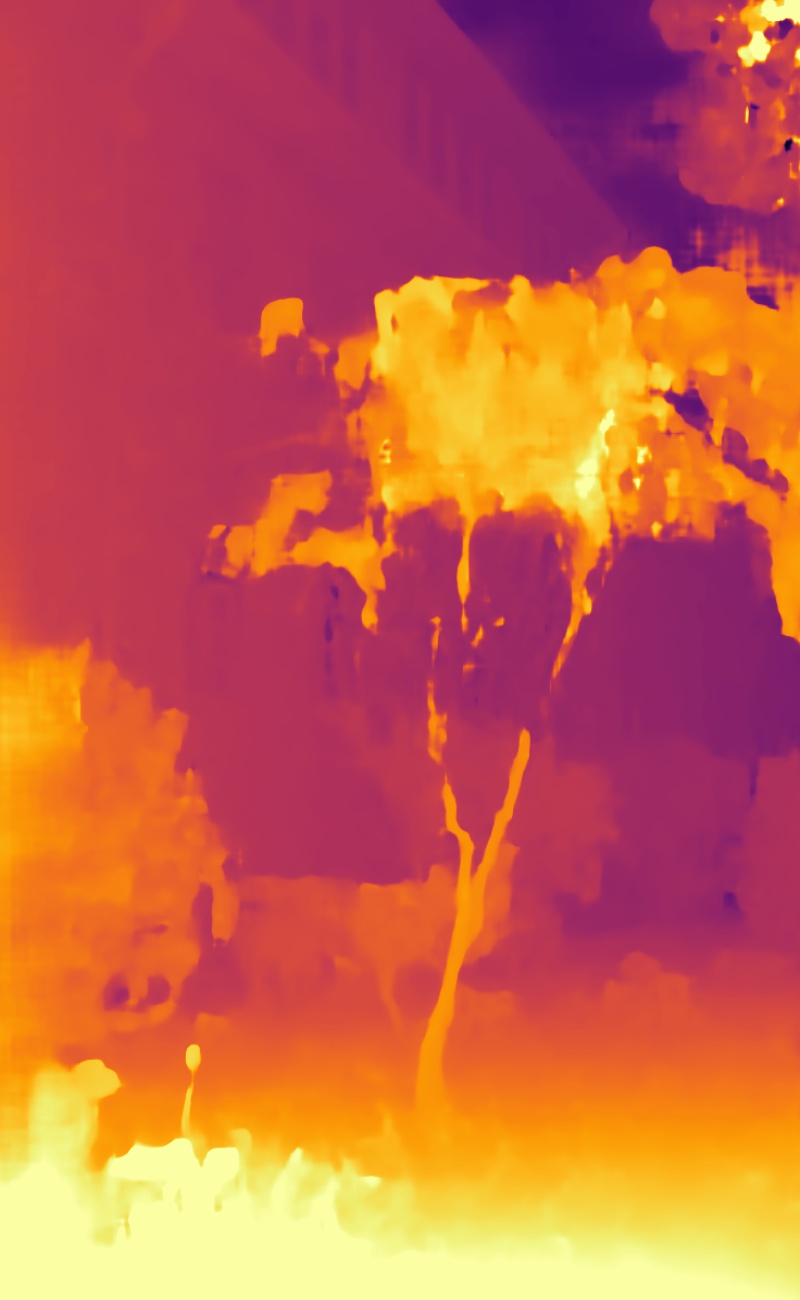} \\
\includegraphics[width=\itemwidth]{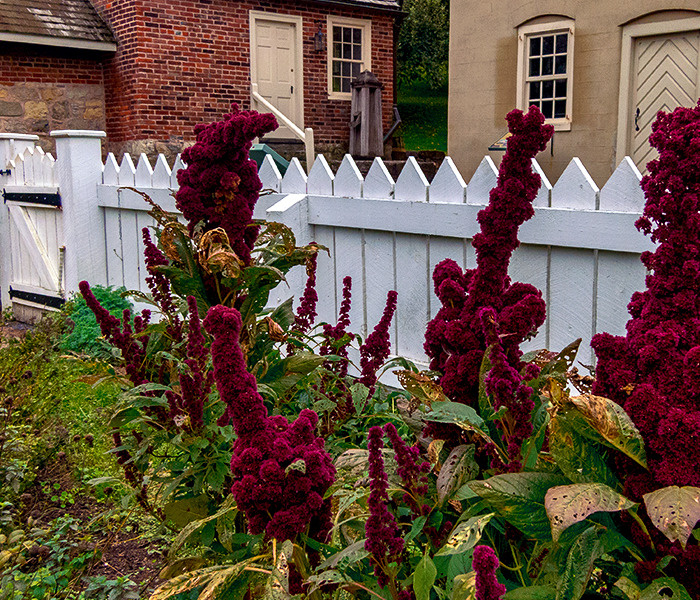} &
\includegraphics[width=\itemwidth]{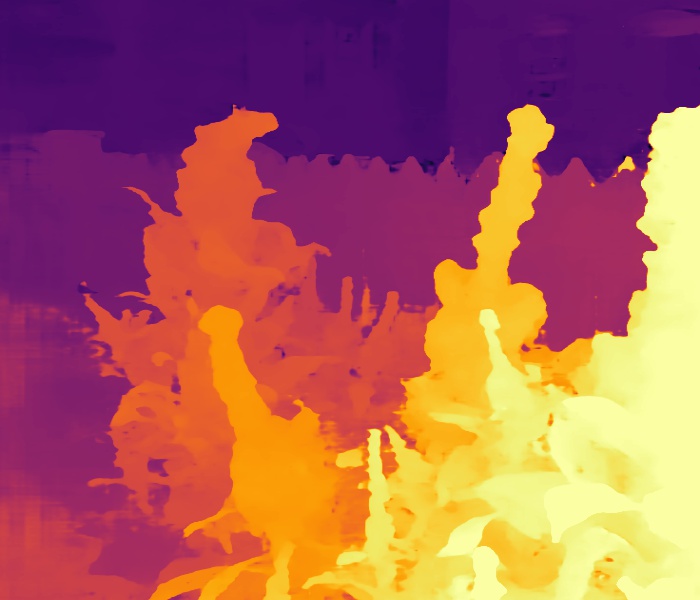} &
\includegraphics[width=\itemwidth]{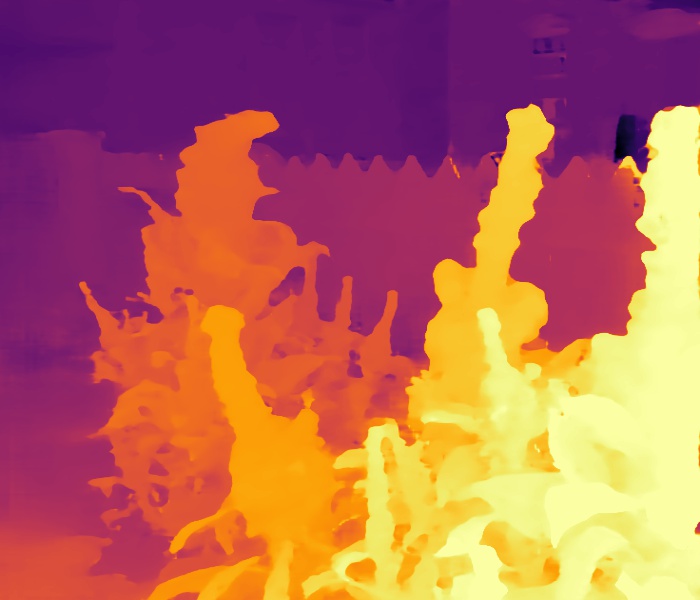} &
\includegraphics[width=\itemwidth]{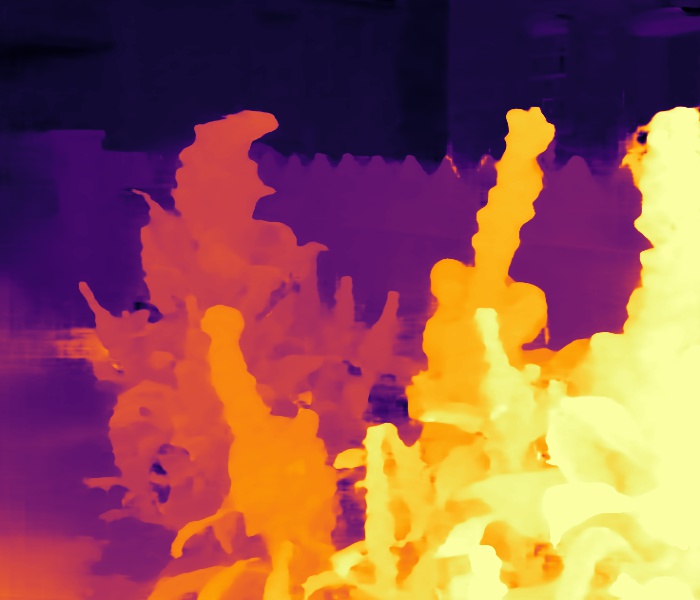} \\
\includegraphics[width=\itemwidth]{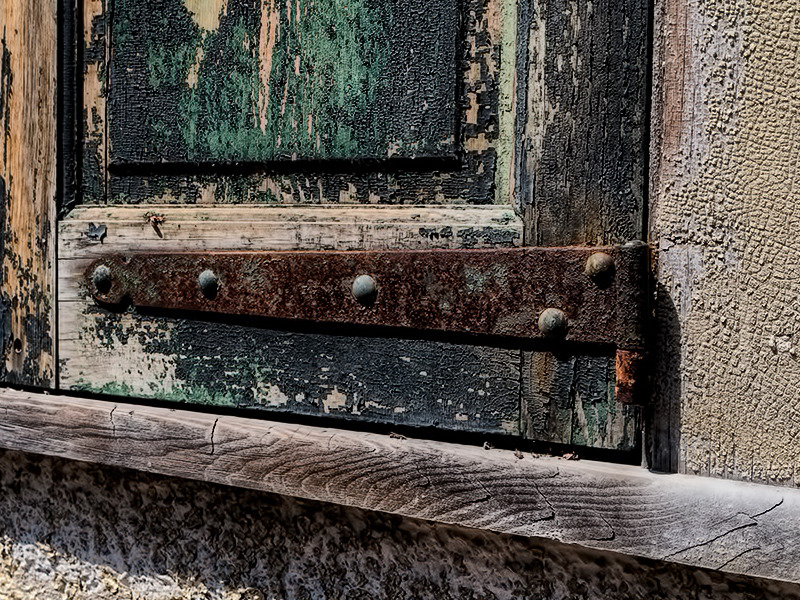} &
\includegraphics[width=\itemwidth]{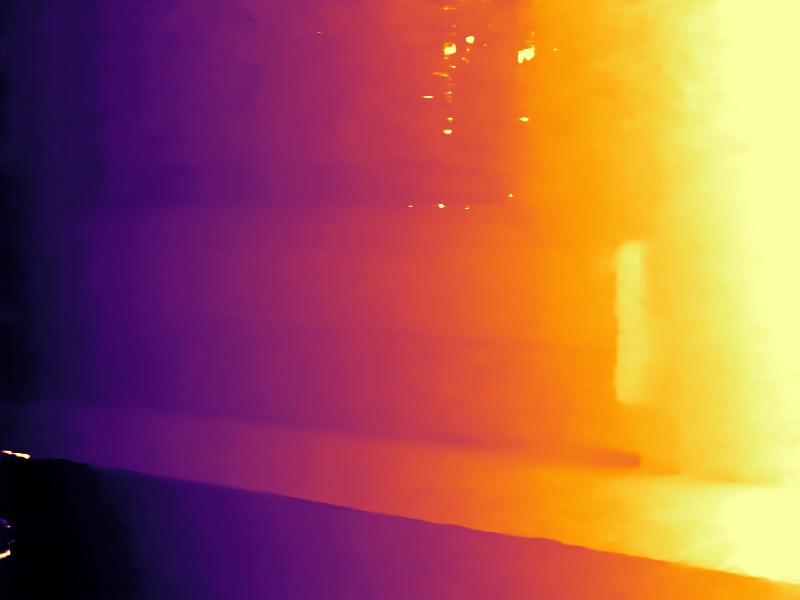} &
\includegraphics[width=\itemwidth]{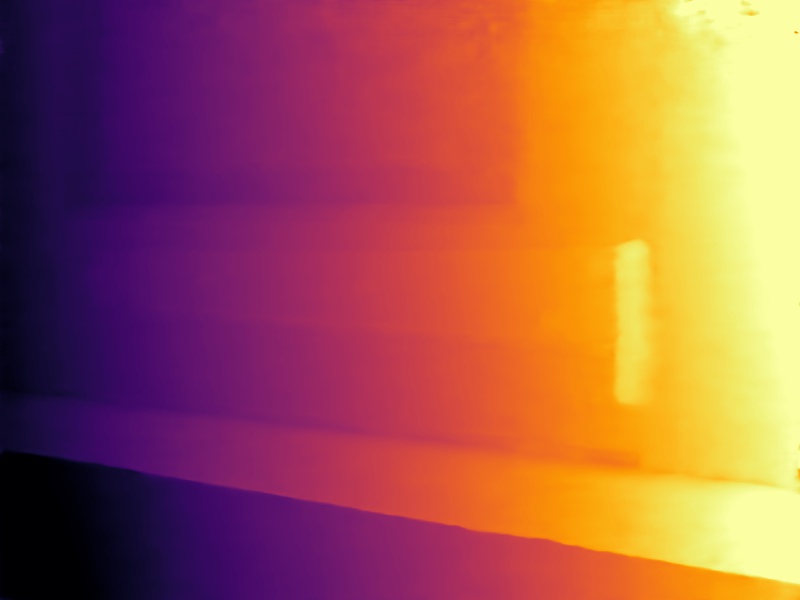} &
\includegraphics[width=\itemwidth]{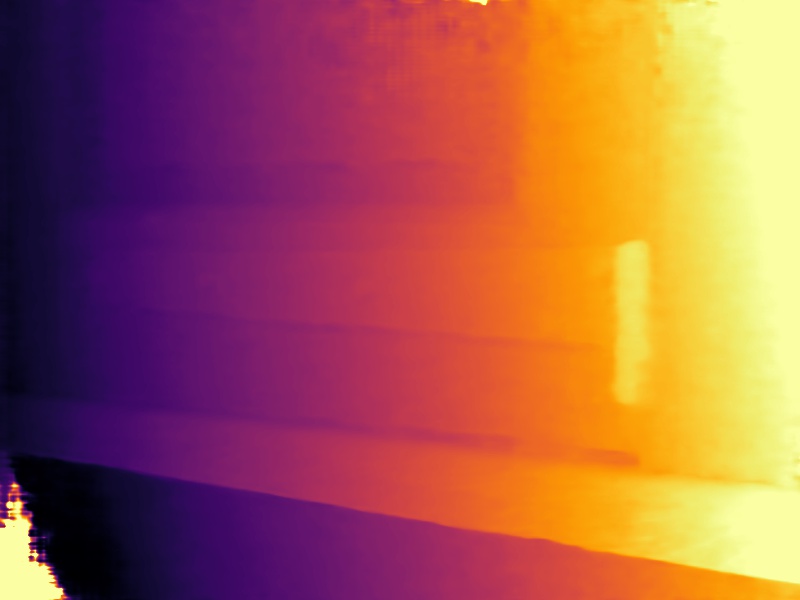} \\
\end{tabular}
\caption{\bf Additional Flickr1024~\cite{flickr1024} qualitative results.}
\label{fig:flickr1024-2-more-qual}
\end{figure}

In Figs.~\ref{fig:kitti12-more-qual} and~\ref{fig:kitti15-more-qual} we present comparisons using GANet~\cite{Zhang2019GANet} to Sceneflow training versus our MfS dataset for both KITTI 2012 and KITTI 2015.
In Figs.~\ref{fig:flickr1024-1-more-qual} and~\ref{fig:flickr1024-2-more-qual} we compare PSM~\cite{chang2018pyramid} networks using Flickr1024~\cite{flickr1024}, a dataset stereo images collected online.
For Flickr1024, we should results using our method using depth generated by MiDaS~\cite{lasinger2019towards} or MegaDepth~\cite{li2018megadepth}.
It is worth remembering that MegaDepth~\cite{li2018megadepth} does not use any synthetic or ground truth depth at training time but it still can be used to train a stereo model that produces high quality predictions.
In all cases we see that our fully automatic data generation pipeline results in high quality stereo predictions without directly requiring any synthetic training data which is time consuming to create.

% \clearpage
% \bibliographystyle{splncs04}
% \bibliography{main}

\end{document}